\documentclass[11pt]{article}

\usepackage[margin=1in]{geometry}
\usepackage[T1]{fontenc}
\usepackage[utf8]{inputenc}
\usepackage{lmodern}
\usepackage{microtype}

\usepackage{amsmath,amssymb,amsfonts}
\usepackage{booktabs}
\usepackage{multirow}
\usepackage{graphicx}
\usepackage{xcolor}
\usepackage{algorithm}
\usepackage{algorithmic}
\usepackage{enumitem}
\usepackage{subcaption}
\usepackage{nicefrac}
\usepackage{pdflscape,adjustbox}
\usepackage{float}
\usepackage[round,authoryear]{natbib}
\usepackage[colorlinks=true,linkcolor=blue,citecolor=blue,urlcolor=blue]{hyperref}

\newcommand{\E}{\mathbb{E}}
\newcommand{\Prob}{\mathbb{P}}
\newcommand{\R}{\mathbb{R}}
\newcommand{\cX}{\mathcal{X}}
\newcommand{\cD}{\mathcal{D}}
\newcommand{\cY}{\mathcal{Y}}

\newcommand{\method}[1]{\texttt{#1}}

\title{Benchmarking Tabular Foundation Models for Conditional Density Estimation in Regression}

\author{  Rafael Izbicki\thanks{Department of Statistics, Federal University of S\~ao Carlos (UFSCar), Brazil. \texttt{rafaelizbicki@gmail.com}}
  \and
  Pedro L. C. Rodrigues\thanks{Univ. Grenoble Alpes, Inria, CNRS, Grenoble INP, LJK, France. \texttt{pedro.rodrigues@inria.fr}}}

\date{}

\begin{document}

\maketitle

% This source uses a standard article + natbib setup for arXiv compatibility.

% ===================================================================
%  ABSTRACT
% ===================================================================
\begin{abstract}

Conditional density estimation (CDE) -- recovering the full conditional distribution of a response given tabular covariates -- is essential in settings with heteroscedasticity, multimodality, or asymmetric uncertainty. Recent tabular foundation models, such as \method{TabPFN} and \method{TabICL}, naturally produce predictive distributions, but their effectiveness as general-purpose CDE methods has not been systematically evaluated -- unlike their performance for point prediction, which is well studied. We benchmark three tabular foundation model variants against a diverse set of parametric, tree-based, and neural CDE baselines on 39 real-world datasets, across training sizes from 50 to 20,000, using six metrics covering density accuracy, calibration, and computation time.  Across all sample sizes, foundation models achieve the best CDE loss, log-likelihood, and CRPS on the large majority of datasets tested. Calibration is competitive at small sample sizes but, for some metrics and datasets, lags behind task-specific neural baselines at larger sample sizes, suggesting that post-hoc recalibration may be a valuable complement. In a photometric redshift case study using SDSS~DR18, \method{TabPFN} exposed to  50,000 training galaxies outperforms all baselines trained on the full 500,000-galaxy dataset. Taken together, these results establish tabular foundation models as strong off-the-shelf conditional density estimators.

\end{abstract}

\section{Introduction}
\label{sec:intro}

Conditional density estimation (CDE) seeks to estimate the full distribution of
a response variable given covariates, rather than only its conditional mean.
In this paper, we focus on the standard univariate-response setting, where the goal is to estimate the full conditional density of a scalar response given tabular covariates.
This is useful when uncertainty, asymmetry, or multimodality matter, with
applications in areas such as photometric redshift estimation, risk analysis, simulation-based inference,
and treatment-response modeling
\citep{schmidt2020evaluation,izbicki2017photo,koenker2005quantile,hothorn2014conditional,izbicki2019abc,cranmer2020frontier,dalmasso2020confidence,dalmasso2024likelihood,cabezas2025cp4sbi}.

Over the years, CDE has been studied through a wide range of approaches,
including classical nonparametric estimators \citep{rosenblatt1969conditional,hyndman1996estimating,izbicki2016nonparametric},
mixture density networks \citep{bishop1994mixture}, and flow-based models
\citep{papamakarios2021normalizing}.
Recent tabular foundation models, such as \method{TabPFN} and \method{TabICL},
have shown strong performance for point-predictions in tabular supervised learning and naturally
produce predictive distributions
\citep{hollmann2025tabpfn,qu2025tabicl,mcelfresh2024neural}.
This makes them plausible candidates for CDE. However, it remains unclear
whether these distributional outputs are competitive with purpose-built CDE
methods in terms of both density accuracy and calibration. Recent work has begun
to evaluate tabular foundation models from a distributional perspective, but
existing benchmark efforts still focus mostly on point prediction
\citep{erickson2025tabarena,landsgesell2026distributional}.

In this paper, we present a broad empirical benchmark of tabular foundation
models for CDE. We compare \method{TabPFN} and \method{TabICL} with a diverse
set of classical and modern CDE baselines on  thirty-nine real-world datasets, across multiple sample sizes and
using six evaluation metrics. We find that tabular foundation models are
consistently competitive across a broad range of settings, most of the time surpassing purpose-built density estimators, although they sometimes do not excel at calibration.

\begin{figure}[H]
  \centering
  \includegraphics[scale=0.54]{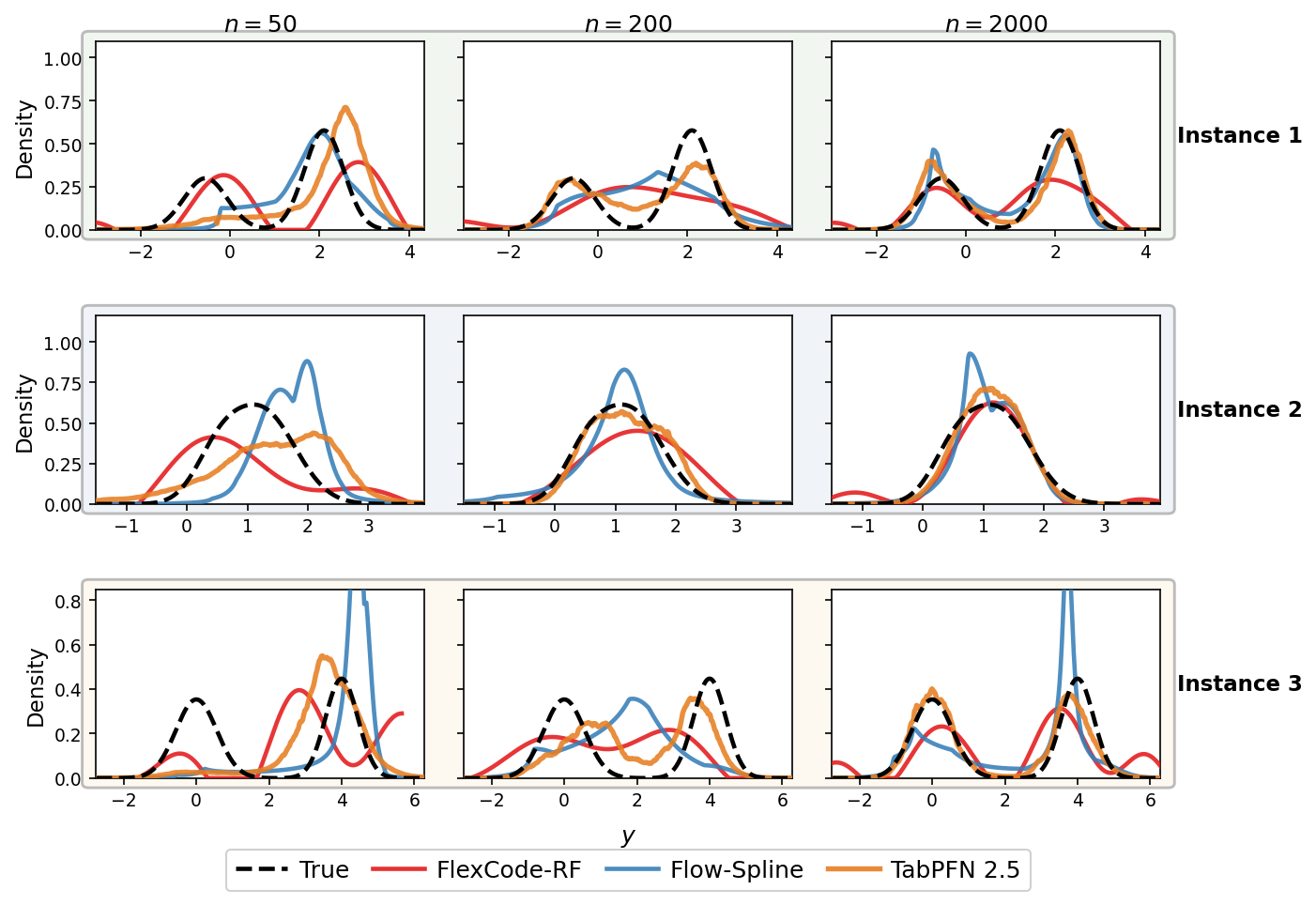}
  \caption{\textbf{Illustration of CDE on a bimodal synthetic DGP.}
    Each row corresponds to a different test instance whose true conditional
    density (dashed black) is a two-component Gaussian mixture with
    covariate-dependent means, variances, and weights.
    Columns show three training sizes ($n\in\{50,200,2{,}000\}$).
    \method{TabPFN-2.5} (orange) already captures the bimodal structure at
    $n=200$, while \method{Flow-Spline} (blue) and \method{FlexCode-RF} (red)
    require considerably more data and still show spurious peaks or roughness. Note that this is a controlled synthetic example designed to illustrate differences among methods; see Section~\ref{sec:results} for real data analyses.}
  \label{fig:bimodal_illustration}
\end{figure}

To illustrate the CDE task and the differences among methods,
Figure~\ref{fig:bimodal_illustration} shows estimated conditional densities for
a synthetic bimodal   process in which two Gaussian components
have covariate-dependent means, variances, and mixture weights.  Three test
instances are shown, each exhibiting a different bimodal shape, and columns
correspond to increasing training-set sizes.
Among the methods compared, \method{TabPFN-2.5} recovers the bimodal structure
at $n=200$, whereas both \method{Flow-Spline} and \method{FlexCode-RF}
require  more data and still exhibit spurious peaks or roughness at
moderate sample sizes.  This example motivates our systematic evaluation:   can tabular foundation models offer a strong out-of-the-box alternative to standard approaches?

\subsection{Relation to Other Work}
\label{sec:related}

\textbf{Conditional density estimation.}
Conditional density estimation has a long history in statistics, going back at least to the nonparametric formulation of \citet{rosenblatt1969conditional}. Classical estimators include  kernel-based, local-polynomial, and local-likelihood approaches \citep{hyndman1996estimating,fan1996estimation,bashtannyk2001bandwidth,hyndman2002nonparametric,efromovich2007conditional}. These methods established many of the core ideas in CDE, including kernel smoothing, local estimation, and bandwidth selection \citep{Izbicki2025}. 
More recent work has developed flexible machine-learning approaches to CDE. Neural methods include mixture density networks \citep{bishop1994mixture}, kernel mixture networks \citep{ambrogioni2017kmn}, and flow-based models built on normalizing flows \citep{dinh2014nice,dinh2017realnvp,papamakarios2017maf,huang2018naf,durkan2019nsf,papamakarios2021normalizing}. Another  line is FlexCode \citep{izbicki2017converting}, which casts CDE as a regression problem. Gaussian-process extensions have also been developed for heteroscedastic and non-Gaussian  distributions \citep{le2005heteroscedastic,platanios2013mgpch,dutordoir2018gpcde}.
 A closely related literature studies probabilistic prediction through parametric families whose distributional parameters depend on covariates. This includes regression quantiles \citep{koenker1978regression}, GAMLSS \citep{rigby2005generalized}, and   distributional regression \citep{kneib2023rage,klein2024distributional}. Related ideas also appear in tree-based and ensemble methods such as quantile regression forests \citep{meinshausen2006quantile}, transformation forests \citep{hothorn2021transformation}, and NGBoost \citep{duan2020ngboost}, as well as in Bayesian approaches including BART-based density regression \citep{chipman2010bart,orlandi2021drbart,li2023adaptive}. Overviews of these and related methods are given by \citet{dalmasso2020cdetools}, \citet{kneib2023rage}, and \citet{klein2024distributional}. 
Taken together, this motivates the families of baselines considered in our benchmark.
\vspace{2mm}

\textbf{Tabular foundation models.}
Recent tabular foundation models build on the prior-data fitted network (PFN) paradigm, in which a transformer is meta-trained on many synthetic supervised tasks and then applied in context to a new dataset without gradient updates at test time \citep{muller2022transformers,nagler2023statistical}. Early work in this area focused mainly on point prediction for tabular classification and regression, typically evaluated by accuracy, AUC, RMSE, or $R^2$. In this line, \citet{hollmann2025tabpfn} introduced \method{TabPFN} for small- to medium-sized tables, while \citet{qu2025tabicl} scaled in-context learning to much larger tables with \method{TabICL}. Benchmark efforts such as \citet{erickson2025tabarena} likewise assess these models mainly through point-prediction metrics.
At the same time, at least some original tabular foundation model papers already
go beyond point prediction. In particular, \citet{hollmann2025tabpfn}
explicitly presents density estimation as one of \method{TabPFN}'s foundation
model abilities and shows proof-of-concept density-estimation experiments.
The broader official \method{TabPFN} ecosystem also exposes related extension-based
capabilities, although the corresponding extensions are explicitly described as
experimental and ``less rigorously tested than the core \texttt{tabpfn}
library'' \citep{priorlabs2025tabpfnextensions}.
More recently, \citet{qu2026tabiclv2} extend the \method{TabICL} line to both
regression and classification, making distributional prediction a more direct
use case. Thus, the main gap is not that tabular foundation models lack
distributional outputs, but rather that their quality as \emph{conditional
density estimators} has only recently begun to be evaluated systematically.
Indeed, \citet{landsgesell2026distributional} note that leading tabular
foundation models already produce full predictive distributions, whereas major
benchmarks still rely mostly on point-estimate metrics. Our work complements
this emerging line by benchmarking tabular foundation models explicitly as CDE
methods.
\vspace{2mm}

\textbf{Benchmarks for tabular learning.}
Most benchmark papers in tabular machine learning focus on point prediction.   For example, \citet{grinsztajn2022tree} showed that tree-based methods remain highly competitive on standard tabular benchmarks. Subsequent efforts broadened the empirical picture in several directions: \citet{mcelfresh2023tabzilla} compared 19 algorithms; \citet{gardner2023tableshift} studied robustness under distribution shift; \citet{rubachev2025tabred} introduced industry-style benchmarks with time-based splits and feature-engineering effects; \citet{ye2024closerlook} evaluated tabular methods on more than 300 datasets in TALENT; \citet{holzmueller2024better} studied strong default baselines on large collections of regression and classification tasks; and \citet{erickson2025tabarena} proposed a continuously maintained living benchmark.
As a result, that literature has generally said more about point-prediction performance than about the quality of the full conditional distribution. 
The closest benchmark-style study to ours is \citet{landsgesell2026distributional}, who evaluate the distributional outputs of RealTabPFN-v2.5 and TabICL-v2 on 20 OpenML datasets using proper scoring rules, demonstrating that leading tabular foundation models produce meaningful predictive distributions. Our work extends this line in three directions: (i)~we include purpose-built CDE baselines from five methodological families---parametric, quantile-based, regression-based, tree-based, and neural---addressing the practical question of whether foundation models can replace established density estimators; (ii)~we substantially expand the experimental scope (39 datasets, sample sizes up to $20{,}000$ in the main benchmark and $500{,}000$ in the SDSS case study, and six evaluation metrics including calibration diagnostics); and (iii)~we provide diagnostic analyses of failure modes and a scaling case study that isolates the sample-efficiency advantage of foundation models.

\subsection{Novelty}
\label{sec:novelty}

Our main contributions are:

\begin{itemize}[leftmargin=*]
  \item \textbf{A broad benchmark of tabular foundation models for CDE:}
    We evaluate the conditional density outputs of \method{TabPFN} and
    \method{TabICL} against a diverse set of parametric 
    and nonparametric baselines.

  \item \textbf{Large-scale experimental design:}
    The benchmark covers  
    thirty-nine real-world datasets from OpenML and SDSS~DR18, with sample
    sizes ranging from 50 to 20{,}000 and covariate dimensions from
    5 to 563.

  \item \textbf{Multifaceted evaluation:}
    We evaluate all methods using six complementary metrics---CDE loss,
    log-likelihood, CRPS, PIT-based calibration, 90\% coverage,   and computation time---to assess accuracy, calibration, and
    efficiency jointly.
\end{itemize}

\section{Background}
\label{sec:background}

\subsection{Conditional Density Estimation}
\label{sec:cde_background}

Let $(X,Y)\sim P_{XY}$, where $X\in\cX\subseteq\R^d$ denotes the covariates and
$Y\in\cY\subseteq\R$ a univariate response. Conditional density estimation
 seeks to estimate the full conditional distribution of $Y$ given $X=x$.
When a conditional density exists, it is a function $f(y\mid x)$ such that
\begin{equation}
  \Prob(Y\in A\mid X=x)=\int_A f(y\mid x)\,dy
  \label{eq:cond_density}
\end{equation}
for every measurable set $A\subseteq\cY$.
Unlike mean or quantile regression, CDE aims to recover the entire shape of the
predictive distribution, including heteroscedasticity, asymmetry, heavy tails,
and multimodality. This is useful whenever downstream decisions depend on
uncertainty summaries such as predictive intervals, tail probabilities, or
quantiles. Throughout the paper, each method receives a training sample
$\cD=\{(x_i,y_i)\}_{i=1}^n$ and returns either a conditional density
$\hat f(y\mid x)$, a conditional distribution function $\hat F(y\mid x)$, or a
set of predictive quantiles from which a density can be derived.

\section{Methods Compared}
\label{sec:methods}

\subsection{Tabular Foundation Models}
\label{sec:methods_tfm}

We evaluate three tabular foundation model variants:
       \method{TabPFN-2.5}, \method{RealTabPFN-2.5}, and
       \method{TabICL-Quantiles}
\citep{hollmann2025tabpfn,qu2025tabicl,qu2026tabiclv2}. All three models are
used as pretrained in-context learners, without task-specific gradient updates. That is, no fine tuning is performed.

The two \method{TabPFN} variants output bar distributions over the response
range.  
We convert these bar distributions into conditional densities as follows: for each bin, we divide the predicted probability mass by the bin width to obtain a density value at the bin center; we then interpolate these bin-center density values onto a fine evaluation grid of 200 equally spaced points using linear interpolation; finally, the interpolated density is renormalized to integrate to one. 
By contrast,
\method{TabICL-Quantiles} outputs predictive quantiles. We convert these
quantiles into a conditional distribution function by interpolation and then
differentiate numerically to obtain a density. Full implementation details for
these conversions are given in Appendix~\ref{app:tfm_details}.

\subsection{Baselines}
\label{sec:methods_baselines}

We compare tabular foundation models with a diverse set of classical and modern CDE baselines spanning five families:
\vspace{2mm}

\textbf{Parametric distributional regression.} We include six models that assume a specific parametric family for $Y\mid X=x$ and allow one or more distributional parameters to depend linearly on the covariates: a homoscedastic and a heteroscedastic Gaussian model (\method{LinearGauss-Homo} and \method{LinearGauss-Hetero}), a linear model with Student-$t$ residuals (\method{Student-t}), log-normal models with constant and covariate-dependent scale (\method{LogNormal-Homo} and \method{LogNormal-Hetero}), and a gamma GLM with log link (\method{Gamma-GLM}) \citep{rigby2005generalized,kneib2023rage}. Each is also included in a ridge-regularized variant (\method{-Ridge} suffix) \citep{hoerl1970ridge}.
Parametric methods via ML estimation are not run when the number of parameters is larger than the number of sample points.
\vspace{2mm}

\textbf{Quantile-based methods.} \method{Quantile-Tree} fits a gradient-boosted tree independently at 49 quantile levels and converts the resulting quantile function to a density by interpolation and numerical differentiation \citep{koenker1978regression,chen2016xgboost}.
\vspace{2mm}

\textbf{Regression-based density estimation.} \method{FlexCode-RF} expresses the conditional density as a cosine basis expansion whose coefficients are estimated by random-forest regression \citep{izbicki2017converting}. 
\vspace{2mm}

\textbf{Tree-based probabilistic models.} \method{BART-Homo} and \method{BART-Hetero} use XBART to estimate the conditional mean (and, in the heteroscedastic variant, an input-dependent variance), inducing a Gaussian predictive density \citep{chipman2010bart,he2019xbart}.
\vspace{2mm}

\textbf{Neural density estimators.} \method{MDN} is a mixture density network with a tuned number of Gaussian components \citep{bishop1994mixture}. \method{Flow-Spline} is a conditional normalizing flow based on rational-quadratic splines \citep{durkan2019nsf,papamakarios2021normalizing}. \method{CatMLP} discretizes the response into bins (in the same spirit as \method{TabPFN}) and fits a multilayer perceptron with a softmax output using cross-entropy loss.
\vspace{2mm}

Detailed
implementation choices for each baseline are reported in
Appendix~\ref{app:baseline_details}.

\section{Experimental Setup}
\label{sec:setup}

We evaluate three tabular foundation model variants and the diverse set of
 baselines on a variety of real-world
regression tasks. Experiments vary the dataset and training-set size, and all
methods are assessed using proper scoring rules, calibration diagnostics,  and computation time. Unless otherwise noted, results
are averaged over 5 independent repetitions.

\subsection{Evaluation Metrics}
\label{sec:metrics}

We evaluate all methods using six complementary metrics covering density
accuracy, calibration, sharpness, and computational cost. Let $\hat f(y\mid x)$
denote an estimated conditional density and let
\begin{equation}
  \hat F(y\mid x)=\int_{-\infty}^{y}\hat f(t\mid x)\,dt
  \label{eq:estimated_cdf}
\end{equation}
be the associated predictive distribution function. When a method outputs
$\hat F$ or predictive quantiles directly, we evaluate the corresponding implied
distribution.

Our main density-specific metric is the CDE loss
\citep{izbicki2016nonparametric,schmidt2020evaluation},
\begin{equation}
  L(\hat f)=\int\!\!\int \hat f(y\mid x)^2\,dy\,dP_X(x)
  -2\,\E_{(X,Y)}[\hat f(Y\mid X)].
  \label{eq:cde_loss}
\end{equation}
This is a proper scoring rule, minimized in expectation by the true conditional
density. Note that CDE loss values are typically negative on real data, with more negative values indicating better density estimates.
We estimate it on held-out test
data. We also report mean test log-likelihood (i.e., the cross-entropy) and the continuous ranked probability
score (CRPS) \citep{gneiting2007strictly}, which are also  proper scoring rules for CDE.

We also evaluate calibration of the CDEs. 
A CDE is calibrated if events predicted to occur with a given probability occur with approximately that same frequency under the data-generating distribution. In the conditional-distribution setting, a standard notion of probabilistic calibration is that, when $(X,Y)$ is drawn from the test distribution, the random variable $
U = \hat F(Y \mid X)$ 
should be approximately uniformly distributed on $[0,1]$ \citep{schmidt2020evaluation,zhao2021diagnostics}. We measure calibration by computing
  the
Kolmogorov--Smirnov statistic of the probability integral transform (PIT) values, as well as empirical coverage  of 90\% predictive intervals.
 Finally, we also evaluate  total fit-and-predict wall-clock
time. 

For each metric, we also test whether each foundation model significantly outperforms every parametric and nonparametric competitor on a given dataset. Specifically, for foundation method $F$ and competitor $C$, we compute a one-sided Welch $t$-test using the conservative variance estimate $\widehat{\mathrm{SE}}_{\mathrm{diff}}^2 = \widehat{\mathrm{SE}}_F^2 + \widehat{\mathrm{SE}}_C^2$, which ignores the positive correlation induced by shared test sets and therefore yields a conservative test (i.e., it is harder to reject the null). We apply Holm--Bonferroni correction across all comparisons for a given foundation method and dataset, and use $\alpha=0.1$. 

\subsection{Protocol and Hyperparameter Tuning}
\label{sec:protocol}

Each experiment is repeated over 5 independent train/test splits (except for $n=50$, in which case we used 50 random splits due to small testing size). In each split, 25\% of the available data is held out as the test set.
We report the
mean of each metric across repetitions together with standard errors. 
All
tabular foundation models are used with their default pretrained settings, and no fine-tuning is performed.

For the baselines, hyperparameters are selected on the training data only.
Ridge-regularized parametric models choose the regularization strength by
cross-validation; \method{FlexCode-RF} selects the number of basis functions by
5-fold cross-validation; and the
remaining baselines (\method{MDN}, \method{Flow-Spline}, \method{Quantile-Tree},
\method{BART-Homo}, \method{BART-Hetero}, and \method{CatMLP}) are tuned by
random search with 3-fold cross-validation drawing  using the
CDE loss as the selection criterion. After tuning,
the selected configuration is refit on the full training set. Complete search
spaces and implementation details are reported in
Appendix~\ref{app:tuning_details}.

\subsection{Datasets}
\label{sec:datasets}

We include  39 regression datasets from OpenML and SDSS~DR18
\citep{sdss_dr18}, spanning domains such as housing, robotics, computer
hardware, transportation, molecular prediction, retail,  music, bioinformatics, energy, and photometric
redshift estimation.  The covariate dimension $d$ (after one-hot encoding of categorical features) ranges from 5 to 563 across the real-world datasets.
For each dataset, we evaluate methods on nested subsamples
of size $n \in \{$50, 500, $1{,}000${, 5{,}000, 10{,}000,} $20{,}000\}$ whenever
available. Complete dataset details are reported in
Appendix~\ref{app:datasets}.

\section{Results}
\label{sec:results}

\begin{figure}[H]
  \centering
  \includegraphics[scale=0.52]{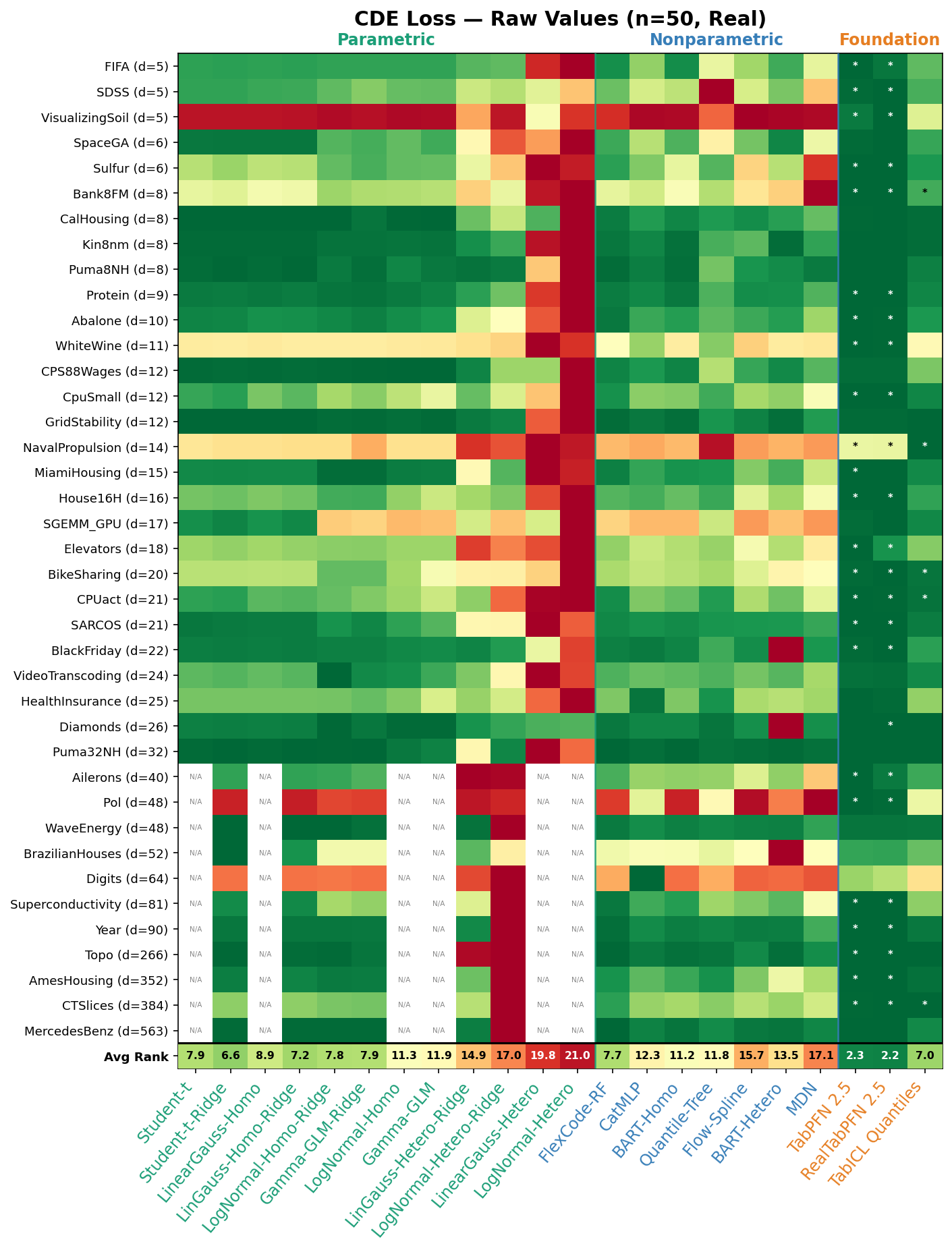}
  \caption{\textbf{CDE loss across real-world datasets at
    $n = 50$.}
   Per-dataset raw CDE loss values (lower/greener is better).
    Datasets are sorted by covariate dimension $d$.
    A $*$ marks foundation models that significantly outperform all
    parametric and nonparametric competitors on that dataset.
    The two \method{TabPFN} variants (orange) achieve the top two average ranks (bottom row), while \method{TabICL-Quantiles} (7.0) is narrowly outranked by \method{Student-t-Ridge} (6.6).
    Even at this extremely small sample size, a foundation model achieves the best CDE loss on 82\% of datasets.
  }
  \label{fig:rankings_cde_n50}
\end{figure}

\begin{figure}[H]
  \centering
  \includegraphics[scale=0.49]{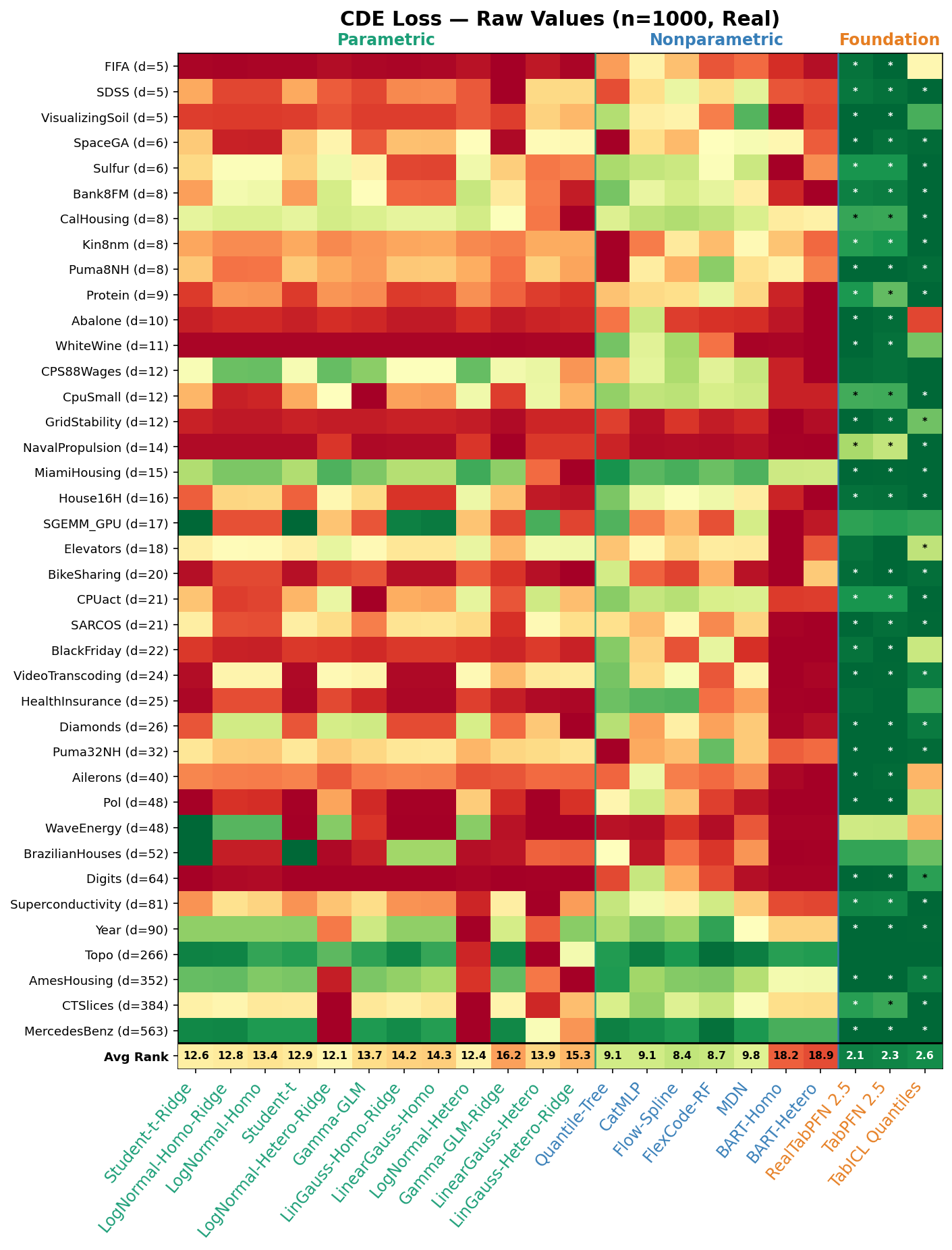}
  \caption{\textbf{CDE loss across real-world datasets at
    $n = 1{,}000$.}
   Per-dataset raw CDE loss values (lower/greener is better).
    Datasets are sorted by $d$.
    A $*$ marks foundation models that significantly outperform all
     competitors on that dataset.
    All  foundation models occupy the top three average ranks.  Foundation models achieve the best CDE loss on 92\% of datasets. }
  \label{fig:rankings_cde_n1000}
\end{figure}

\begin{figure}[H]
  \centering
  \includegraphics[scale=0.43]{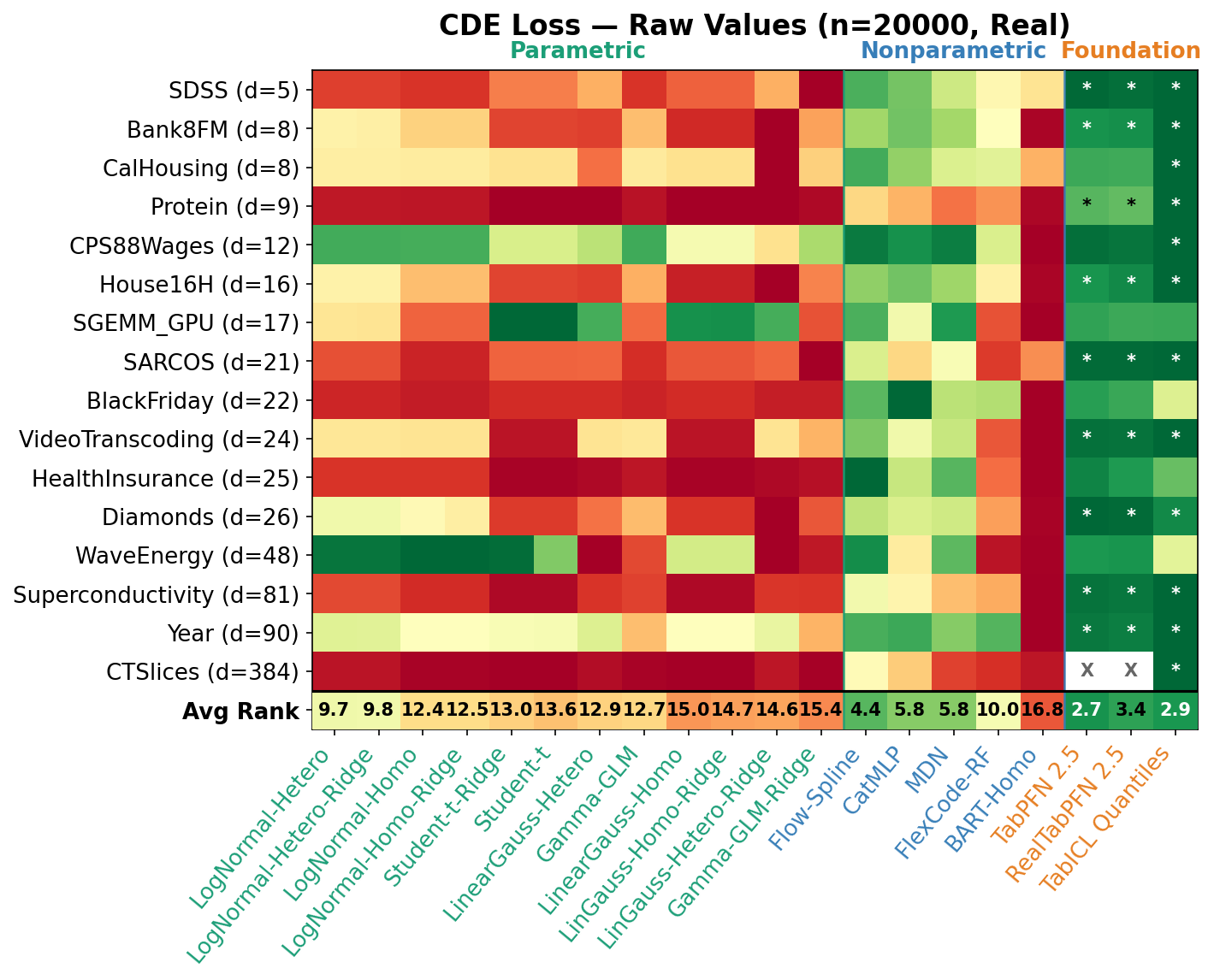}
  \caption{\textbf{CDE loss across real-world datasets at
    $n = 20{,}000$.}
    Per-dataset raw CDE loss values (lower/greener is better).
    Datasets are sorted by $d$.
    A $*$ marks foundation models that significantly outperform all competitors on that dataset; $\times$ marks foundation models that encountered out-of-memory errors.
    All three foundation models occupy the top three average ranks.
    On CTSlices, both \method{TabPFN} variants ran out of memory at this sample size.}
  \label{fig:rankings_cde_n20000}
\end{figure}

Figures~\ref{fig:rankings_cde_n50}--\ref{fig:rankings_cde_n20000} present CDE loss raw-value heatmaps across real-world datasets at $n \in \{50, 1{,}000, 20{,}000\}$, with datasets sorted by covariate dimension $d$ (results at other sample sizes are reported in Appendix~\ref{app:rankings}).
Across all sample sizes, foundation models dominate the top positions in average rank.

At $n=50$ (Figure~\ref{fig:rankings_cde_n50}), \method{RealTabPFN-2.5} achieves an average rank of 2.2 across the 39 datasets, with \method{TabPFN-2.5} (2.3) close behind.
The third-best average rank, however, is held by the non-foundation \method{Student-t-Ridge} (6.6), which narrowly outperforms \method{TabICL-Quantiles} (7.0); thus at this smallest sample size two of the three foundation model variants occupy the top two positions, but \method{TabICL-Quantiles} lags behind the best parametric baseline.
Among the remaining non-foundation competitors, \method{FlexCode-RF} (7.7) and \method{LogNormal-Homo-Ridge} (7.8) follow.
Despite \method{TabICL-Quantiles}' weaker ranking at $n=50$, a foundation model achieves the best CDE loss on 32 out of 39 datasets (82\%), indicating broad dominance even at this extremely small sample size.
Significance tests confirm this pattern: a $*$ in the heatmap marks cases where a foundation model significantly outperforms all non-foundation competitors. 

At $n=1{,}000$ (Figure~\ref{fig:rankings_cde_n1000}), all three foundation models occupy the top three average ranks: \method{RealTabPFN-2.5} leads at 2.1, followed by \method{TabPFN-2.5} (2.3) and \method{TabICL-Quantiles} (2.6).
The gap to the best non-foundation baselines is substantial: \method{Flow-Spline} (8.4), \method{FlexCode-RF} (8.7), and \method{Quantile-Tree} and \method{CatMLP} (both 9.1) trail by a wide margin.
Foundation models achieve the best CDE loss on 36 out of 39 datasets (92\%) at this sample size, with significant wins ($*$) on many datasets.

At $n=20{,}000$ (Figure~\ref{fig:rankings_cde_n20000}), which covers the 16 datasets large enough for this subsample, all three foundation models again hold the top three average ranks: \method{TabPFN-2.5} (2.7), \method{TabICL-Quantiles} (2.9), and \method{RealTabPFN-2.5} (3.4). The best non-foundation competitor is \method{Flow-Spline} (4.4), with \method{CatMLP} and \method{MDN} tied at 5.8.
The gap between foundation models and the best nonparametric baselines narrows as $n$ grows (compare the 6+ rank-point gap at $n = 1{,}000$ with the $\approx$1--2 rank-point gap at $n = 20{,}000$), consistent with a sample-efficiency advantage that diminishes as non-foundation methods receive more training data.
Foundation models achieve the best CDE loss on 12 out of 16 datasets (75\%) at this sample size. 
 However, a notable practical limitation emerged at this sample size: on the CTSlices dataset ($d = 384$), both \method{TabPFN} variants encountered out-of-memory errors and could not produce predictions (marked with $\times$ in Figure~\ref{fig:rankings_cde_n20000}) on a NVIDIA GeForce RTX 5070 Ti GPU with 16303 MiB of VRAM. Only \method{TabICL-Quantiles} ran successfully on this dataset, achieving rank~1. This indicates that the combination of large sample size and high covariate dimension can exceed the memory capacity of current \method{TabPFN} implementations, a constraint that users should be aware of in high-dimensional settings.

The ranking pattern under log-likelihood and CRPS (Appendix~\ref{app:perf_vs_n}) is broadly consistent with the CDE loss results, though with some differences for \method{TabICL-Quantiles}. Under log-likelihood, the two \method{TabPFN} variants lead at every sample size (e.g., average ranks of 1.9 at $n=1{,}000$), but \method{TabICL-Quantiles} ranks further behind (3.8 at $n = 1{,}000$; 8.2 at $n=50$; 4.3 at $n = 20{,}000$). At $n = 20{,}000$, \method{Flow-Spline} (3.9) surpasses \method{TabICL-Quantiles} under log-likelihood, so that only two of the three foundation models remain in the top three for this metric.
Under CRPS, all three foundation models consistently occupy the top three ranks at every sample size, and \method{TabICL-Quantiles} performs particularly well, achieving average ranks of 2.1 at $n = 1{,}000$ and 1.7 at $n = 20{,}000$---the best overall among all methods. A likely explanation is that \texttt{TabICL} outputs predictive quantiles, which must be converted to densities via interpolation and numerical differentiation. CRPS can be computed directly from the quantile function and is thus unaffected by this conversion, whereas CDE loss and log-likelihood evaluate the density pointwise and are sensitive to the roughness it introduces. This suggests that \texttt{TabICL-Quantiles}' predictive distributions are high quality, but that the density conversion  penalizes density-specific metrics.

Calibration results based on the PIT KS statistic (Appendix~\ref{app:perf_vs_n}) paint a more nuanced picture. 
Foundation models achieve among the best calibration ranks at small sample sizes (e.g., \method{TabPFN} variants rank 4.4–4.6 at $n = 50$), and remain in the upper half of methods at $n = 1,000$. However, at $n \geq 5,000$, their calibration ranks deteriorate to mid-table (ranks 7–9 for the \method{TabPFN} variants), while \method{MDN} typically achieves the best calibration. This suggests that foundation models' in-context learning mechanism, while highly effective for density accuracy, does not scale as well for calibration as end-to-end training on larger datasets. 
The 90\% empirical coverage results (Appendix~\ref{app:rankings}) show a similar trend. This confirms that strong performance on proper scoring rules does not automatically guarantee the best calibration across all settings.

\begin{figure}[H]
  \centering
  \includegraphics[scale=0.25]{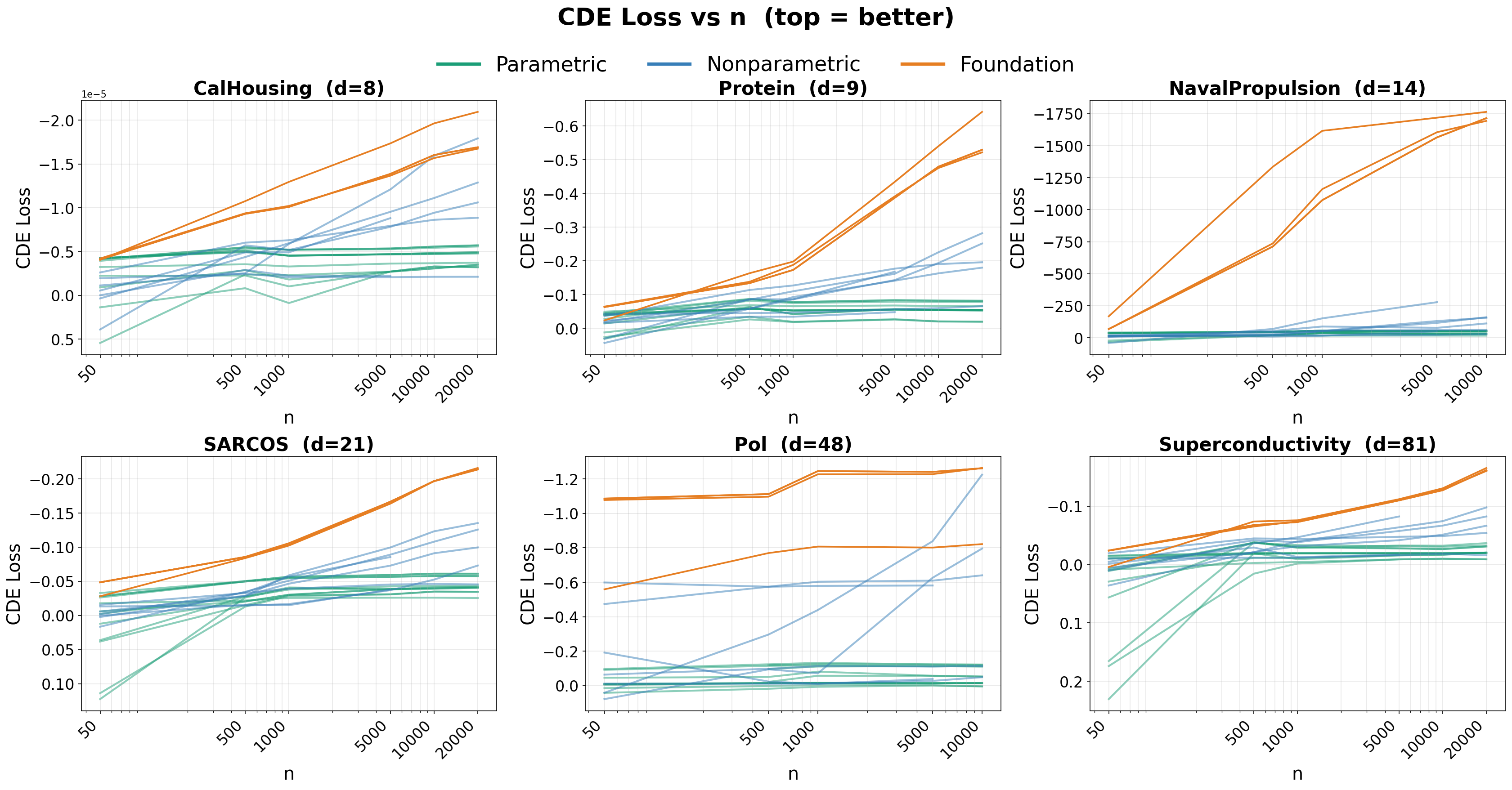}
  \caption{\textbf{CDE loss vs.\ sample size for selected real-world datasets.} Each panel shows one dataset, arranged by increasing $d$. Foundation models (orange, bold) are contrasted against parametric (green, faded) and nonparametric (blue, faded) baselines; the $y$-axis is oriented so that higher values are better. Foundation models are already competitive at $n=50$ and often exhibit the steepest improvement with growing $n$. Parametric baselines plateau early and rarely reach the top regardless of sample size. Nonparametric methods show greater variability, occasionally matching foundation models in higher-dimensional settings but without consistency.}
  \label{fig:perf_vs_n_real}
\end{figure}

Figure~\ref{fig:perf_vs_n_real} displays CDE loss as a function of sample size $n$ across six real-world datasets, arranged by increasing covariate dimension, with methods grouped into parametric, nonparametric, and foundation-model families. Foundation models are already competitive at $n = 50$, and their advantage generally widens as $n$ grows---most dramatically in NavalPropulsion ($d = 14$), where they far surpass all other methods at large~$n$. The foundation-model advantage persists in higher dimensions: at $d = 81$ (Superconductivity) these models remain the best performers across all sample sizes. At $d = 48$ (Pol), the picture is more mixed, with individual nonparametric methods occasionally matching foundation models at intermediate sample sizes, but no single nonparametric estimator consistently doing so across datasets. Parametric methods are tightly clustered and rarely reach the top regardless of sample size.

\begin{figure}[H]
  \centering
  \includegraphics[scale=0.33]{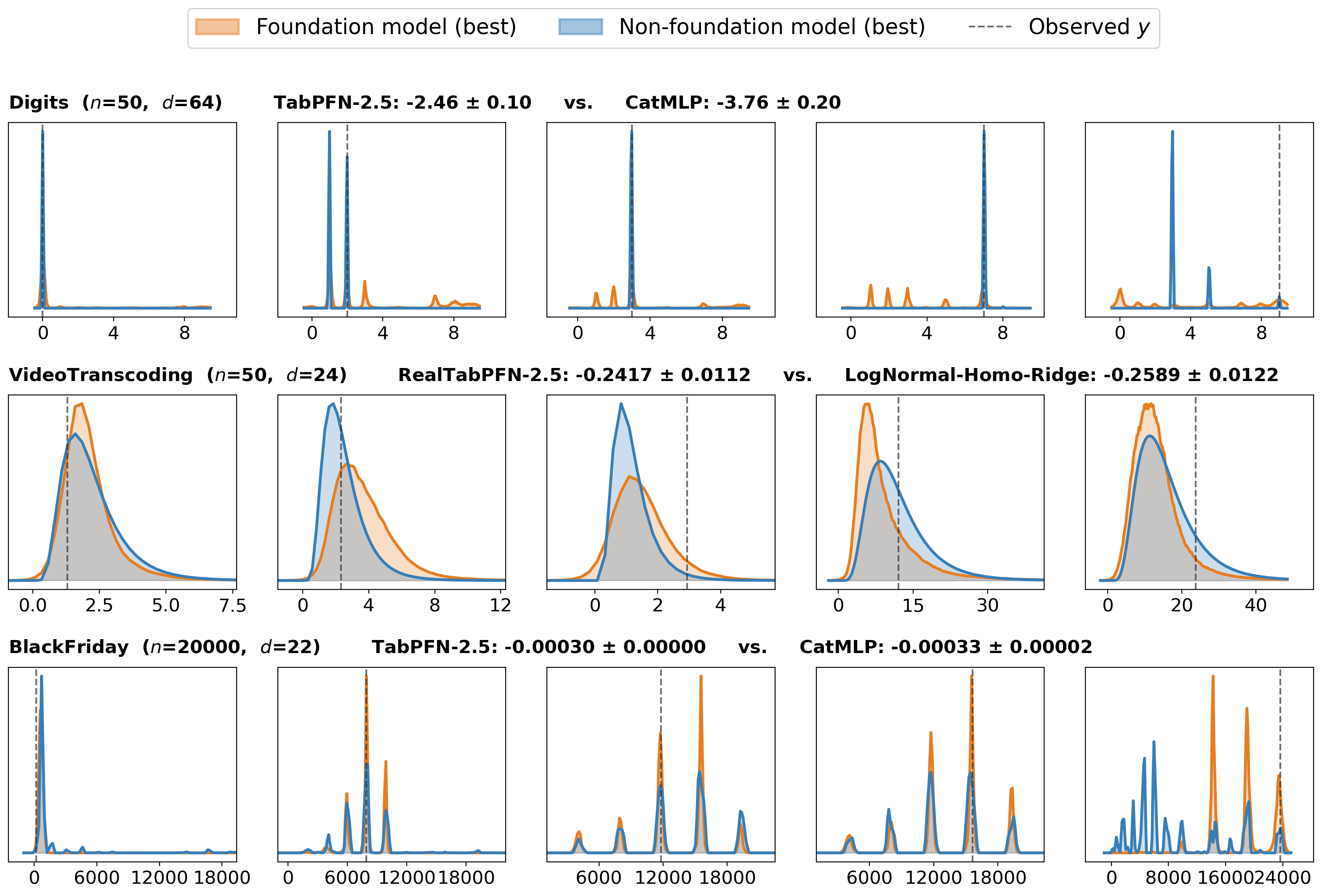}
\caption{\textbf{Diagnostic examples of foundation-model failures on real data.} Each row shows a task where the best foundation model (orange) is beaten in CDE loss by the best non-foundation method (blue); five test cases are shown per row, with dashed lines marking observed outcomes. On Digits, \method{CatMLP} captures the discrete support more precisely than \method{TabPFN-2.5}. On VideoTranscoding, \method{LogNormal-Homo-Ridge} produces sharper tail densities. On BlackFriday, \method{CatMLP} better captures the quasi-discrete response structure---a gap that persists even at the largest available sample size.}
  \label{fig:examples}
\end{figure}

Figure~\ref{fig:examples} highlights counterexamples where foundation models lose to specialized methods. On Digits ($n=50$, $d=64$), where responses take only a few discrete values, \method{CatMLP} outperforms \method{TabPFN-2.5} by placing probability mass exactly on the support points, though this gap disappears by $n=500$, suggesting a small-sample issue rather than a fundamental limitation. On VideoTranscoding ($n=50$, $d=24$), a simple \method{LogNormal-Homo-Ridge} model beats \method{RealTabPFN-2.5} because its strong parametric bias yields better estimates with little data, especially in the right tail; again, the advantage vanishes as $n$ grows. On BlackFriday ($n=20{,}000$, $d=22$), however, \method{CatMLP} also outperforms \method{TabPFN-2.5}, and the gap persists even at the largest sample size, indicating a structural disadvantage for the foundation model on quasi-discrete outcomes.
These examples suggest that foundation models can be outperformed when the response has structure that is naturally captured by a specialized model---such as discrete or quasi-discrete support, or a well-matched parametric family whose inductive bias doubles as regularization---though in most such cases the advantage is confined to small sample sizes.
 Notice however that these cases are the exception rather than the rule: across all 39 datasets, only 3--7 (depending on sample size) see a non-foundation method rank first.

\section{Case Study: Photometric Redshift Estimation with SDSS}
\label{sec:sdss}

The benchmark results in Section~\ref{sec:results} evaluate all methods at
sample sizes up to $n = 20{,}000$, which aligns with the  
context-length limits of current tabular foundation models.
An important  question is how these models behave in a 
large-scale  setting, and in particular how their performance
at moderate $n$ compares to classical methods that can exploit  larger
training sets.
To address this, we conduct a  scaling study on photometric
redshift estimation using the Sloan Digital Sky Survey (SDSS) Data
Release~18 \citep{sdss_dr18}.

The SDSS dataset contains approximately 500{,}000 galaxies with
$ugriz$ broadband photometric magnitudes as covariates and spectroscopic
redshift $z$ as the response.
Estimating the full conditional density $f(z \mid \text{photometry})$
is a standard  problem in astrophysics, as downstream
cosmological analyses depend on accurate predictive distributions
rather than point estimates
\citep{schmidt2020evaluation,izbicki2017photo}.
This makes SDSS a natural testbed for evaluating conditional density
estimators at scale.

We construct a sequence of nested subsamples
$n \in \{500, 1\text{k}, 10\text{k}, 50\text{k}, 100\text{k}, 250\text{k}, 500\text{k}\}$,
where each smaller dataset is a strict subset of the next.
All experiments are repeated over 5 independent train/test splits.
Tabular foundation models are constrained by context length:
\method{TabPFN} variants are evaluated up to $n = 50{,}000$ (on GPU),
while \method{TabICL-Quantiles} is evaluated up to $n = 10{,}000$
under default configurations.
All other methods are run at all sample sizes up to $500{,}000$ whenever
the total fit-and-predict time remains below one hour.
 In addition to the methods considered in the main benchmark, we include
\method{FlexZBoost}, a sharpening-enhanced variant of \method{FlexCode}
based on gradient-boosted regressors \citep{dalmasso2020cdetools},
 developed specifically for the   photometric redshift estimation
task \citep{schmidt2020evaluation}.

\begin{figure}[H]
  \centering
  \includegraphics[scale=0.26]{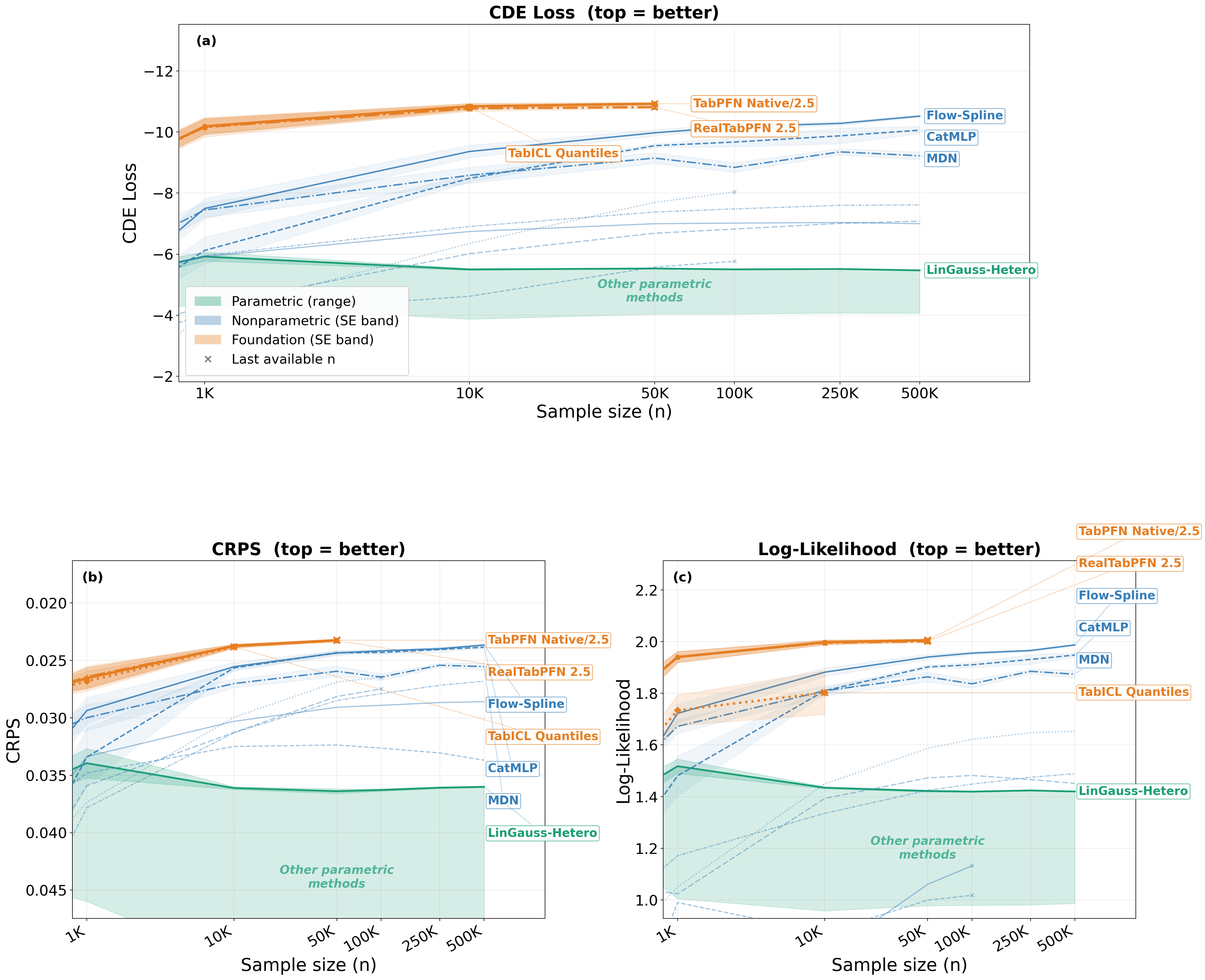}
  \caption{
  \textbf{ CDE loss (panel a), CRPS (panel b), and log-likelihood (panel c) vs.\ training sample size on the SDSS photometric redshift dataset.} All $y$-axes are oriented so that top = better (CDE loss and CRPS are inverted; log-likelihood is shown in natural order).
Parametric methods (green): the shaded band spans the full range across all parametric models; the best-performing parametric model (\method{LinearGauss-Hetero}) is highlighted as a solid line. Ridge-regularized variants are omitted from the band as they are uniformly worse and would distort the $y$-axis scale.
Nonparametric methods (blue): each method is shown as a separate line; shaded $\pm 1$ SE bands are displayed only for the three best-performing nonparametric methods to avoid visual clutter.
A terminal $\times$ marker indicates the last sample size at which a method was evaluated.
Key finding: across all three metrics, \method{TabPFN} trained on 50k galaxies outperforms all  methods trained on 500k galaxies, demonstrating a ten-fold data efficiency advantage. }
  \label{fig:sdss_scaling}
\end{figure}

\begin{figure}[H]
  \centering
  \includegraphics[width=\linewidth]{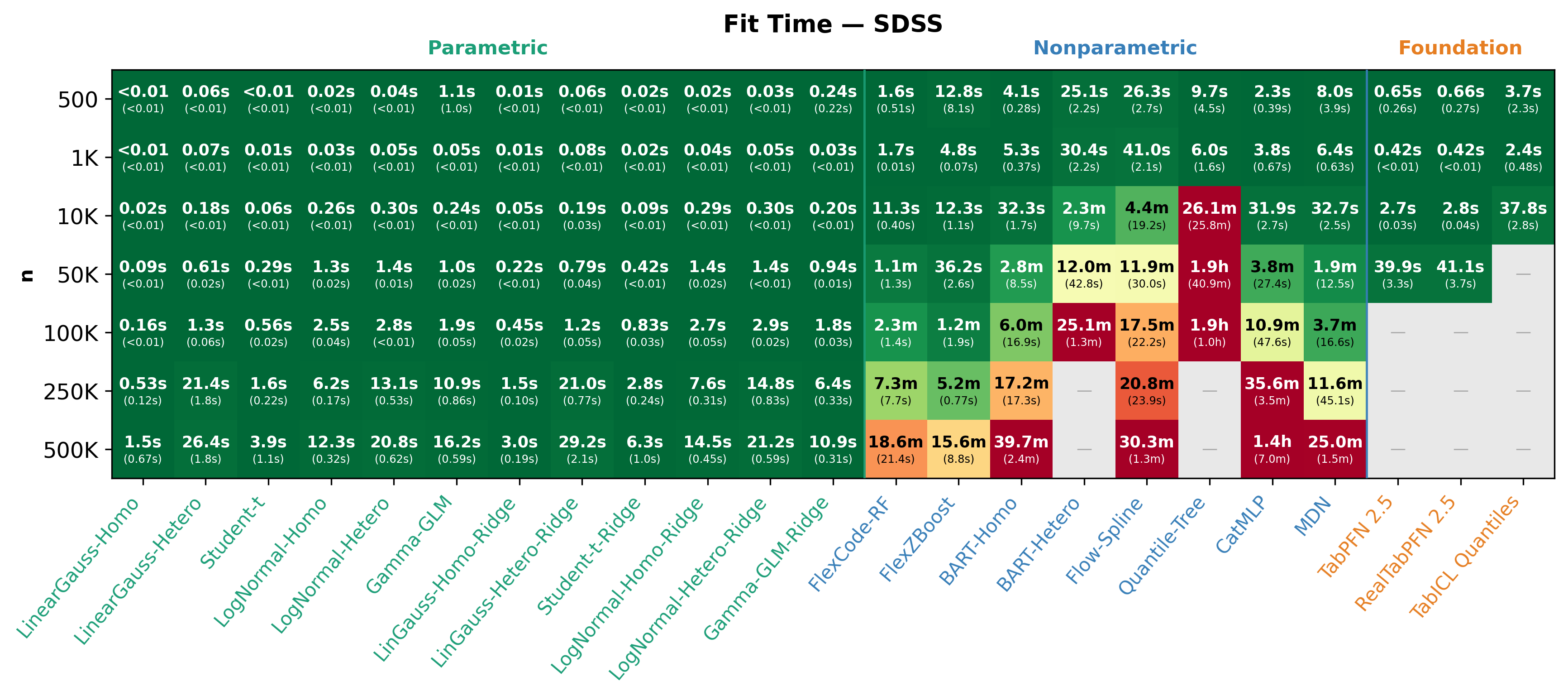}
\caption{\textbf{Fit time (training + prediction) for each method across sample sizes~$n$ on the SDSS photometric redshift dataset.}
Each cell reports the mean fit time (top) and standard error in
parentheses.
Cells are colour-coded within each method column from green (fastest)
to red (slowest); grey cells indicate that the method was not run
at that sample size due to computational limits.
Fit time includes both training and inference on the test set.
  Key finding: Foundation models are substantially faster than neural baselines (\method{Flow-Spline}, \method{MDN}) at the same sample sizes, while parametric baselines are the cheapest overall but sacrifice density accuracy.}
  \label{fig:sdss_scalingTime}
\end{figure}

Figure~\ref{fig:sdss_scaling} reports CDE loss, CRPS, and log-likelihood
as a function of training set size.
The dominant pattern is the strong sample efficiency of tabular foundation
models.
At $n = 50{,}000$—the largest dataset they can process—the
\method{TabPFN} variants achieve a CDE loss of approximately $-10.8$,
surpassing all baselines trained on the full $n = 500{,}000$ dataset.
The same qualitative behavior is observed for CRPS and log-likelihood. 
This comparison isolates an important trade-off:
foundation models operate under strong context constraints,
yet extract substantially more predictive signal per training example.
In this setting, conditioning on only 10\% of the available data yields
better conditional density estimates than conventional methods trained
on the entire dataset.
Calibration diagnostics tell a similar story.
PIT-based KS statistics and empirical coverage (Appendix~\ref{app:sdss})
show that foundation models at $n = 50{,}000$ remain competitive with
the best baselines, indicating that their gains in accuracy are not
achieved at the expense of probabilistic calibration.
Finally,  Figure~\ref{fig:sdss_scalingTime} shows that tabular foundational models have low computation time when compared to competitive nonparametric models.
For instance, \method{TabPFN-2.5} processes 50{,}000 galaxies in approximately 39 seconds on GPU, while \method{Flow-Spline}—the strongest non-foundation competitor at $n = 500{,}000$—requires roughly 19 minutes on CPU and still achieves a weaker CDE loss. The foundation model thus achieves superior CDE performance in a fraction of both the data and the wall-clock time, though this comparison involves different hardware (GPU vs.\ CPU).

From a practical perspective, these results suggest that tabular
foundation models can substantially reduce data requirements in
 workflows where labeled data is expensive.
In this case, achieving comparable CDE performance with
classical methods would require roughly an order of magnitude more data.

\section{Final Remarks}
\label{sec:final}

We presented a broad empirical benchmark of tabular foundation models for conditional density estimation in tabular regression. Across 39 real-world datasets, training sizes from 50 to 20{,}000, and six complementary evaluation metrics, these models proved to be the strongest off-the-shelf approach for density estimation. The \method{TabPFN} variants achieved the best results on CDE loss, log-likelihood, and CRPS on most datasets, while \method{TabICL-Quantiles} also ranked among the top-performing methods on most metrics, despite some loss in density-based performance that may stem from the quantile-to-density conversion step. Calibration results were more mixed, with foundation models performing especially well at small sample sizes but sometimes being overtaken at larger \(n\) by task-specific neural baselines such as \method{MDN}. Importantly, this performance was achieved without any dataset-specific fine-tuning or changes to the pretraining objective. The SDSS photometric redshift case study further demonstrates the practical value of these models: using only 50{,}000 training galaxies, \method{TabPFN} outperformed all competing methods trained on the full 500{,}000-example dataset, pointing to a clear sample-efficiency advantage for uncertainty-aware regression.

\textbf{Limitations.} Despite these strong results, several caveats apply.
First, all \method{TabPFN} variants are subject to
       context-length constraints (here capped at $n = 50{,}000$)
       and typically require GPU for such sample sizes.
       Moreover, the combination of large $n$ and high covariate
       dimension can trigger out-of-memory errors well below the
       context-length limit: in our experiments, both \method{TabPFN}
       variants failed on CTSlices ($d = 384$) already at
       $n = 20{,}000$.
         These constraints limit direct applicability
       to very large or high-dimensional datasets without subsampling
       or batching strategies.
 Second, while foundation models are broadly competitive on marginal calibration, we observed that for some datasets their calibration is worse than some baselines. Third, our benchmark is restricted to the univariate-response setting; it is unclear how these conclusions extend to multivariate or structured outputs.
Finally, as illustrated by the diagnostic counterexamples in Figure~\ref{fig:examples}, when the response has structure that is naturally captured by a specialized model—such as discrete support or a well-matched parametric family—the task-specific inductive bias can be more informative than the general-purpose prior acquired during pre-training. In such regimes,   current foundation models' pre-training priors, while broadly effective, do not yet fully subsume the knowledge encoded in well-chosen structural assumptions about the response.
\vspace{2mm}

\textbf{Future directions.} This work opens several avenues for further investigation. Combining tabular foundation models with post-hoc recalibration (such as \citealt{dey2025towards}) may address the calibration gaps observed on some datasets.  Moreover, extending the evaluation to multivariate conditional density estimation, structured outputs, and downstream decision-making tasks that depend on the full predictive distribution is a natural next step.

We hope this benchmark helps shift the evaluation of tabular learning beyond point prediction and toward a fuller assessment of distributional quality. Code to reproduce all experiments, as well as tables with all metrics and standard errors, is available at \url{https://github.com/rizbicki/tabDensityComparisons}.

% ===================================================================
%  ACKNOWLEDGEMENTS
% ===================================================================
\section*{Acknowledgements}
RI is grateful for the financial support of CNPq (422705/2021-7, 305065/2023-8 and 403458/2025-0) and FAPESP (grant 2023/07068-1).
PLCR was supported by a national grant managed by the French National Research Agency (Agence Nationale de la Recherche) attributed to the SBI4C project of the MIAI AI Cluster, under the reference ANR-23-IACL-0006.

% ===================================================================
%  REFERENCES
% ===================================================================
\bibliography{references}
\bibliographystyle{plainnat}

% ===================================================================
%  APPENDIX
% ===================================================================
\appendix

\section{Implementation Details}
\label{app:implementation}

\subsection{From Foundation-Model Outputs to Densities}
\label{app:tfm_details}

\paragraph{\method{TabPFN-2.5}.}
This variant uses the explicit v2.5 default regression checkpoint from the
\method{TabPFN} release \citep{hollmann2025tabpfn}.   In regression mode, it outputs a bar distribution:
a discrete probability mass over bins that partition the response range.  We convert this bar distribution to a conditional density as follows: for each bin, the probability mass is divided by the bin width to obtain a density value at the bin center; these bin-center density values are then linearly interpolated onto a fine evaluation grid of 200 equally spaced points spanning the training range (plus a 5\% margin on each side); finally, the interpolated density is clamped to be non-negative and renormalized to integrate to one. This procedure yields a smooth approximation to the underlying piecewise-constant bar density rather than evaluating the exact step function.

\paragraph{\method{RealTabPFN-2.5}.}
This variant uses the \method{RealTabPFN} regression checkpoint from the
\method{TabPFN-2.5} release \citep{hollmann2025tabpfn}. The prediction
interface is the same as for \method{TabPFN-2.5}, but the model is additionally
fine-tuned on curated real-world tabular datasets. As with the other
\method{TabPFN} variants, we use its native bar distribution as the starting
point for the CDE output, following the same interpolation procedure described above.

\paragraph{\method{TabICL-Quantiles}.}
\method{TabICL} also follows an in-context learning paradigm for tabular
prediction, but uses a different architecture and pretraining strategy
\citep{qu2025tabicl,qu2026tabiclv2}. We use an ensemble of four estimators. Its
regression variant outputs predictive quantiles. We extract 199 quantiles at
levels $\alpha \in \{0.005, 0.010, \ldots, 0.995\}$, enforce monotonicity by
sorting, and interpolate the resulting quantile function onto a regular grid to
obtain the CDF. The conditional density is then computed as the numerical
gradient of the interpolated CDF, clamped to be non-negative and renormalized.

\subsection{Baseline Implementations}
\label{app:baseline_details}

\subsubsection{Parametric baselines}
\label{app:parametric_details}

These methods assume a parametric family for $Y\mid X=x$ and allow one or more
distributional parameters to depend on the covariates.

\paragraph{\method{LinearGauss-Homo}.}
A Gaussian linear model with mean linear in $x$ and constant variance:
$Y\mid X=x\sim\mathcal N(\beta^\top x,\sigma^2)$, where $\hat\sigma^2$ is
estimated from training residuals with a degrees-of-freedom correction
\citep{rigby2005generalized,kneib2023rage}.

\paragraph{\method{LinearGauss-Hetero}.}
A heteroscedastic Gaussian model in which both the mean and the log-variance
depend on $x$:
$Y \mid X = x \sim \mathcal{N}(x^\top\beta,\,\exp(x^\top\gamma))$
\citep{rigby2005generalized,kneib2023rage}. The parameters $\beta$ and $\gamma$
are estimated \emph{jointly} by maximizing the Gaussian log-likelihood
\[
  \ell(\beta,\gamma)
  = -\tfrac{1}{2}\sum_{i=1}^{n}\bigl[x_i^\top\gamma
    + (y_i - x_i^\top\beta)^2\,\exp(-x_i^\top\gamma)\bigr],
\]
via L-BFGS-B (with analytic gradient), initialized from two-step OLS estimates
of $\beta$ and a least-squares fit of log-squared residuals for $\gamma$.
The predictive density is
$\hat f(y\mid x)=\mathcal N\!\bigl(y;\,x^\top\hat\beta,\,
\exp(x^\top\hat\gamma)\bigr)$.

\paragraph{\method{Student-t}.}
A linear model with Student-$t$ conditional distribution
\citep{rigby2005generalized,kneib2023rage,klein2024distributional}. The
conditional mean is estimated by linear regression,
$\hat\mu(x)=x^\top\hat\beta$. The degrees of freedom $\hat\nu$ and scale
$\hat\sigma$ are then estimated by profile maximum likelihood on the training
residuals, optimizing $\nu$ over $[2.01,200]$ and setting
$\hat\sigma=\sqrt{\overline{\varepsilon^2}(\hat\nu-2)/\hat\nu}$ when
$\hat\nu>2$.

\paragraph{\method{LogNormal-Homo} and \method{LogNormal-Hetero}.}
These models assume a log-normal conditional distribution with linear structure
in the log-scale mean \citep{rigby2005generalized,kneib2023rage}. When the
response is not strictly positive, all training values are shifted by a
constant
$c=-\min(y_{\text{train}})+0.01\cdot\mathrm{range}(y_{\text{train}})$
so that the shifted response $\tilde Y=Y+c>0$. The model is then
$\log \tilde Y\mid X=x \sim \mathcal N(x^\top\beta,\sigma^2)$, with either
constant $\sigma$ (\method{-Homo}) or input-dependent
$\log\sigma^2(x) = x^\top\gamma$ estimated jointly with the mean via L-BFGS-B
(\method{-Hetero}), analogously to \method{LinearGauss-Hetero}.
The density is evaluated on the original scale and renormalized.

\paragraph{\method{Gamma-GLM}.}
A gamma generalized linear model with a log link for the conditional mean
\citep{rigby2005generalized,kneib2023rage,klein2024distributional}. As with
the log-normal models, when the response is not strictly positive the training
values are shifted by
$c=-\min(y_{\text{train}})+0.01\cdot\mathrm{range}(y_{\text{train}})$. On the
shifted data, $\log \hat\mu(x)=x^\top\hat\beta$ is estimated by regressing
$\log\tilde y$ on $x$, and a constant shape parameter $\hat a$ is estimated
from the variance of the log-scale residuals as
$\hat a=1/\mathrm{Var}(\log\tilde y-x^\top\hat\beta)$. The predictive density
is
$\hat f(\tilde y\mid x)=\mathrm{Gamma}(\tilde y;\,a=\hat a,\,
\mathrm{scale}=\hat\mu(x)/\hat a)$, evaluated on the original scale and
renormalized.

For each parametric model above, we also include a ridge-regularized variant
(\method{-Ridge} suffix), where the regression coefficients are estimated by
ridge regression with the regularization strength selected by leave-one-out
cross-validation over 20 log-spaced values in $[10^{-4},10^4]$.
For the heteroscedastic variants, the same ridge penalty is applied to both
$\beta$ and $\gamma$ in the joint likelihood.

\subsubsection{Nonparametric and flexible baselines}
\label{app:nonparametric_details}

\paragraph{\method{FlexCode-RF}.}
\method{FlexCode} represents the conditional density through a cosine
orthonormal basis expansion on the normalized response:
$\hat f(z\mid x)=\sum_{i=0}^{I-1}\hat\beta_i(x)\varphi_i(z)$, where
$\varphi_0\equiv 1$ and
$\varphi_i(z)=\sqrt{2}\cos(i\pi z)$ for $i\geq 1$
\citep{izbicki2017converting}. Each coefficient function $\hat\beta_i(x)$ is
estimated by regressing $\varphi_i(Y)$ on $X$ using a random forest with 100
trees and maximum depth 8. The number of basis functions is selected by
cross-validation, and the estimated density is projected to be non-negative and
renormalized.

\paragraph{{\method{FlexZBoost}.}}
\method{FlexZBoost} is a variant of \method{FlexCode} that replaces the random forest regression engine with XGBoost (100 trees, maximum depth 4, learning rate 0.1) and adds a post-hoc sharpening step \citep{dalmasso2020cdetools}. After fitting the basis expansion, the estimated density is raised to a power $\alpha > 0$ and renormalized; the sharpening parameter $\alpha$ is tuned by 5-fold cross-validation minimizing the CDE loss, over a grid of 16 values in $[0.5,\, 2.0]$. This method has been shown to perform well on photometric redshift estimation tasks \citep{schmidt2020evaluation}.

\paragraph{\method{BART-Homo}.}
This method uses XBART to model the conditional mean and pairs it with a
constant residual variance estimated from the posterior mean of the XBART
$\sigma$ draws, averaged over post-burn-in sweeps, inducing a Gaussian
predictive density \citep{chipman2010bart,he2019xbart}.

\paragraph{\method{BART-Hetero}.}
A two-stage XBART-based approach: one XBART model estimates the conditional
mean and a second XBART model is fit to the log-squared residuals to estimate
the input-dependent variance, yielding a Gaussian conditional density with
heteroscedastic variance \citep{chipman2010bart,he2019xbart}.

\paragraph{\method{Quantile-Tree}.}
A gradient-boosted tree model (XGBoost with the quantile error objective) fit
independently at 49 quantile levels
$\alpha\in\{0.02,0.04,\ldots,0.98\}$. The fitted quantiles are sorted to
enforce monotonicity, linearly interpolated to obtain the CDF, and
numerically differentiated to obtain the density, which is then clamped to be
non-negative and renormalized
\citep{koenker1978regression,meinshausen2006quantile,chen2016xgboost}.

\paragraph{\method{Flow-Spline}.}
A conditional neural spline flow \citep{durkan2019nsf,papamakarios2021normalizing}.
The model learns an invertible transformation from a standard Gaussian base
distribution to the conditional distribution of $Y\mid X=x$, parameterized as
a composition of rational-quadratic spline layers. Each layer's spline
parameters are produced by a two-hidden-layer MLP conditioned on $x$.
Training minimizes the negative log-likelihood with Adam in mini-batches
(batch size 512), with gradient clipping (max norm~5.0) and early stopping on a
10\% validation split (patience~12).

\paragraph{\method{MDN}.}
A mixture density network with a tuned number of Gaussian mixture components (selected from $\{2,3,5\}$ by cross-validation)
\citep{bishop1994mixture}. The architecture consists of one hidden layer with
ReLU activation mapping the input to the mixture weights, component means, and
component log-standard deviations. Training minimizes the negative mixture
log-likelihood with Adam, with early stopping based on a 10\% held-out validation split (patience of 30 epochs).

\paragraph{{\method{CatMLP}.}}
A categorical MLP that discretizes the response variable into $n_{\mathrm{bins}}$ equal-width bins and fits a two-hidden-layer MLP (with ReLU activations) to predict the bin probabilities via a softmax output and a cross-entropy training objective. The predicted bin probabilities are converted to a density by dividing by the bin width, then linearly interpolated onto the 200-point evaluation grid and renormalized. Training uses Adam with early stopping on a 10\% validation split (patience~30). Hyperparameters (number of bins, hidden units, learning rate, and training epochs) are tuned by random search.

\subsection{Hyperparameter Search}
\label{app:tuning_details}

The foundation models \method{TabPFN-2.5},
\method{RealTabPFN-2.5}, and \method{TabICL-Quantiles}) are used without any
hyperparameter tuning.

For the parametric baselines, the main tuning choice is whether to use ordinary
least squares or ridge regression. In the ridge variants, the regularization
strength is selected from 20 log-spaced candidates in $[10^{-4},10^4]$ via
leave-one-out cross-validation.

For \method{FlexCode-RF}  {and \method{FlexZBoost}}, the number of cosine basis functions $I$ is selected
by 5-fold cross-validation on the training set, minimizing the CDE loss, with
$
I_{\max}=\min\bigl({30},\max(15,\lfloor\sqrt{n}\rfloor)\bigr).
$ 
 {For \method{FlexZBoost}, the sharpening exponent $\alpha$ is additionally selected by 5-fold cross-validation over 16 values in $[0.5, 2.0]$.}

For the remaining baselines
(\method{MDN}, \method{Flow-Spline}, \method{Quantile-Tree},
\method{BART-Homo}, \method{BART-Hetero},  {and \method{CatMLP}}),
hyperparameters are tuned by random search with 3-fold cross-validation,
drawing 8 configurations and selecting the
configuration that minimizes the mean CDE loss across folds.

\begin{table}[H]
\centering
\caption{\textbf{Random-search spaces for tuned baselines.}}
\label{tab:hyperparams}
\small
\begin{tabular}{lp{9cm}}
\toprule
\textbf{Method} & \textbf{Search space} \\
\midrule
\method{MDN} &
number of components $\in \{2,3,5\}$; hidden units $\in \{16,32,64\}$;
learning rate $\in \{0.005,0.01,0.02\}$; epochs $\in \{300,500,800\}$ \\
\method{Flow-Spline} &
spline bins $\in \{4,8,12\}$; layers $\in \{2,3,4\}$;
hidden units $\in \{32,64,128\}$;
learning rate $\in \{10^{-3},2\times 10^{-3},5\times 10^{-3}\}$;
weight decay $\in \{10^{-6},10^{-5},10^{-4}\}$ \\
\method{Quantile-Tree} &
number of boosting rounds $\in \{50,100,200\}$;
max depth $\in \{3,4,6\}$;
learning rate $\in \{0.05,0.1,0.2\}$ \\
\method{BART-Homo}, \method{BART-Hetero} &
number of trees $\in \{20,30,50\}$;
number of sweeps $\in \{40,60,80\}$ \\
 {\method{CatMLP}} &
{number of bins $\in \{30,50,100\}$; hidden units $\in \{32,64,128\}$;
learning rate $\in \{0.005,0.01,0.02\}$; epochs $\in \{300,500,800\}$} \\
\bottomrule
\end{tabular}
\end{table}

After tuning, the selected configuration is refit on the full training set and
used for final prediction on the corresponding test split.

% ===================================================================
%  APPENDIX: DATASETS
% ===================================================================

\section{Dataset Details}
\label{app:datasets}

Table~\ref{tab:real_datasets} lists the  {39} real-world regression datasets used
in the benchmark, together with their source \citep{vanschoren2013openml,sdss_dr18}, maximum available sample size,
and application domain.  The covariate dimension $d$ reported is the number of features after one-hot encoding of categorical variables. Datasets are included at all sample sizes $n \in \{50, 500, 1{,}000, 5{,}000, 10{,}000, 20{,}000\}$ for which the dataset has at least $n$ observations.

\begin{table}[H]
\centering
\caption{\textbf{Real-world datasets.}}
\label{tab:real_datasets}
\small
\begin{tabular}{llrl}
\toprule
\textbf{Dataset} & \textbf{Source} & \textbf{Max $n$} & \textbf{Domain} \\
\midrule
SpaceGA          & OpenML &    3{,}107 & Spatial analysis \\
Elevators        & OpenML &   16{,}599 & Control systems \\
Kin8nm           & OpenML &    8{,}192 & Robotics \\
Puma8NH          & OpenML &    8{,}192 & Robotics \\
Bank8FM          & OpenML &   22{,}784 & Finance \\
CpuSmall         & OpenML &    8{,}192 & Computer hardware \\
CPUact           & OpenML &    8{,}192 & Computer hardware \\
CalHousing       & OpenML &   20{,}640 & Housing \\
Diamonds         & OpenML &   53{,}940 & Gemology \\
Abalone          & OpenML &    4{,}177 & Marine biology \\
Ailerons         & OpenML &   13{,}750 & Control systems \\
BikeSharing      & OpenML &   17{,}379 & Transportation \\
AmesHousing      & OpenML &    2{,}930 & Housing \\
Digits           & OpenML &    5{,}620 & Digit recognition \\
House16H         & OpenML &   22{,}784 & Housing \\
Sulfur           & OpenML &   10{,}081 & Chemistry \\
BrazilianHouses  & OpenML &   10{,}692 & Housing \\
Pol              & OpenML &   15{,}000 & Politics \\
MercedesBenz     & OpenML &    4{,}209 & Manufacturing \\
Protein          & OpenML &   45{,}730 & Bioinformatics \\
VisualizingSoil  & OpenML &    8{,}641 & Soil science \\
Year             & OpenML &  515{,}345 & Music \\
SGEMM\_GPU       & OpenML &  241{,}600 & GPU performance \\
BlackFriday      & OpenML &  166{,}821 & Retail \\
{Superconductivity} & {OpenML} & {21{,}263} & {Physics} \\
{WaveEnergy}        & {OpenML} & {72{,}000} & {Energy} \\
{VideoTranscoding}  & {OpenML} & {68{,}784} & {Computing} \\
{SARCOS}            & {OpenML} & {44{,}484} & {Robotics} \\
{NavalPropulsion}   & {OpenML} & {11{,}934} & {Engineering} \\
{GridStability}     & {OpenML} & {10{,}000} & {Energy systems} \\
{MiamiHousing}      & {OpenML} & {13{,}932} & {Housing} \\
{HealthInsurance}   & {OpenML} & {22{,}272} & {Healthcare} \\
{CPS88Wages}        & {OpenML} & {28{,}155} & {Economics} \\
{WhiteWine}         & {OpenML} &  {4{,}898} & {Food science} \\
{FIFA}              & {OpenML} & {18{,}063} & {Sports} \\
{Puma32NH}          & {OpenML} &  {8{,}192} & {Robotics} \\
{CTSlices}          & {OpenML} & {53{,}500} & {Medical imaging} \\
{Topo}              & {OpenML} &  {8{,}885} & {Topology} \\
SDSS             & SDSS DR18 & 500{,}000 & Astrophysics \\
\bottomrule
\end{tabular}
\end{table}

\section{Additional Results}

\label{app:rankings}

This section displays
the heatmaps of all metrics for all experiments.

\subsection{Real Datasets Experiments}

\label{app:perf_vs_n}

We start with the results for the real data experiments.

\begin{figure}[p]\centering
  \includegraphics[width=\linewidth]{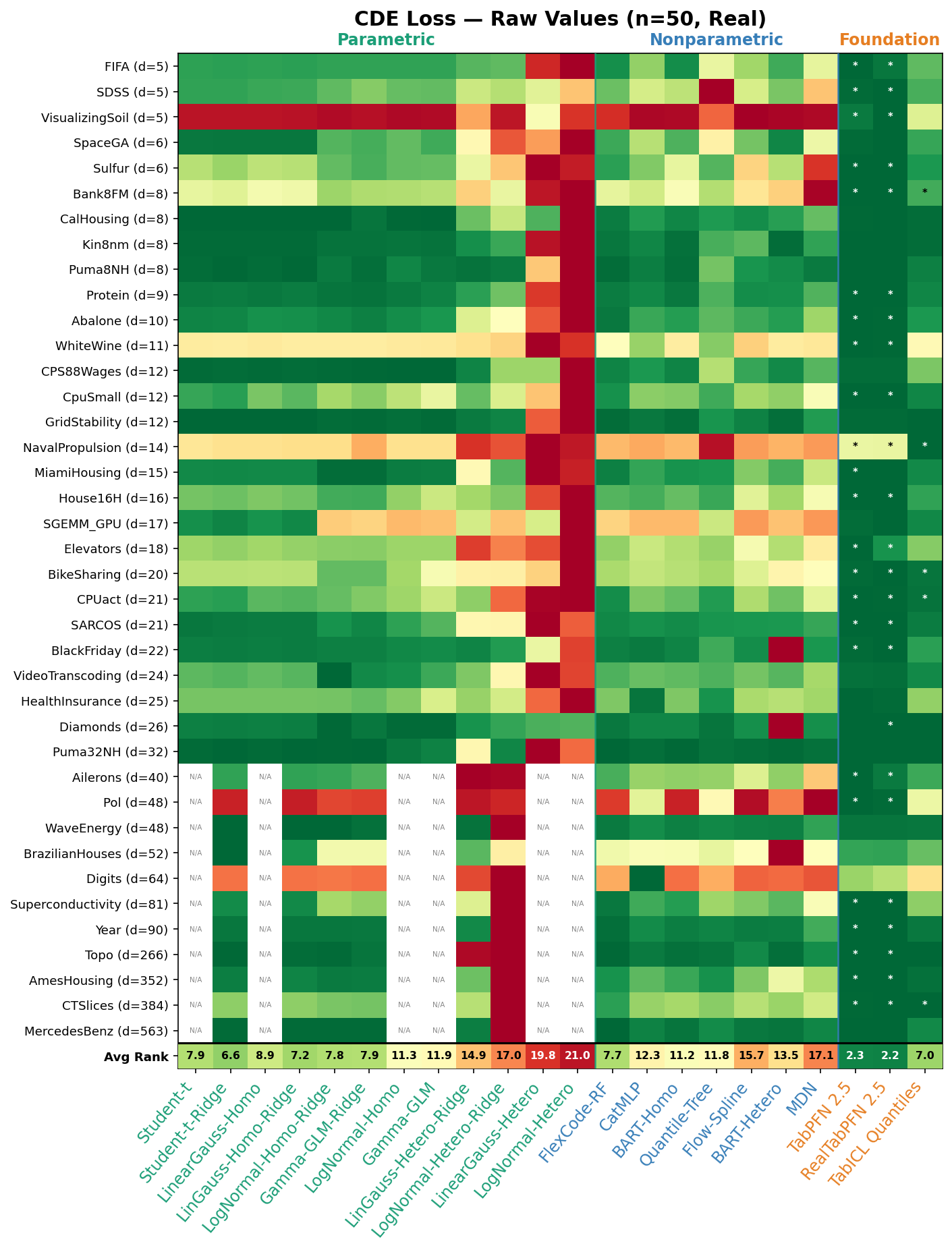}
  \caption{Raw CDE Loss -- $n=50$, real data.}
\end{figure}
\begin{figure}[p]\centering
  \includegraphics[width=\linewidth]{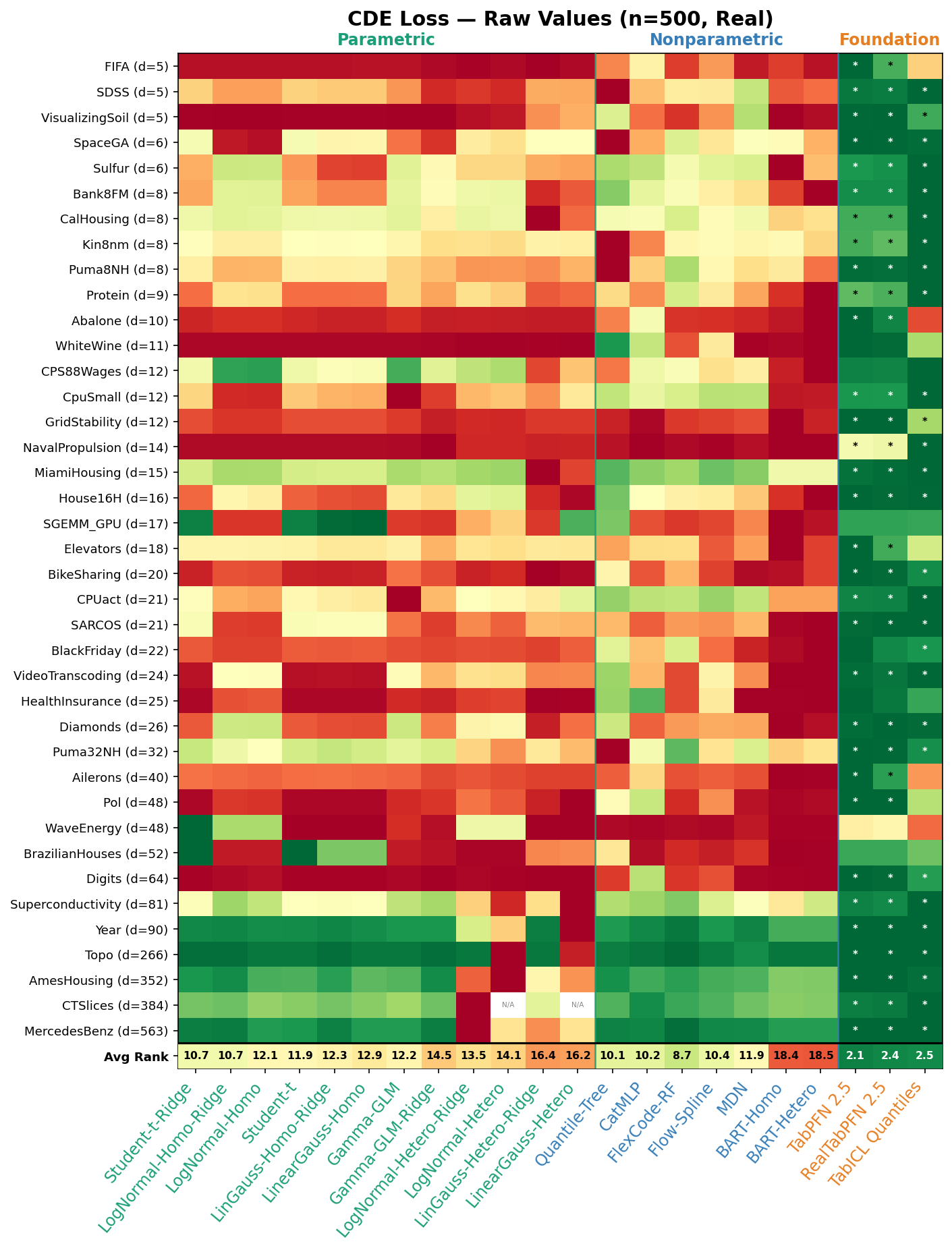}
  \caption{Raw CDE Loss -- $n=500$, real data.}
\end{figure}
\begin{figure}[p]\centering
  \includegraphics[width=\linewidth]{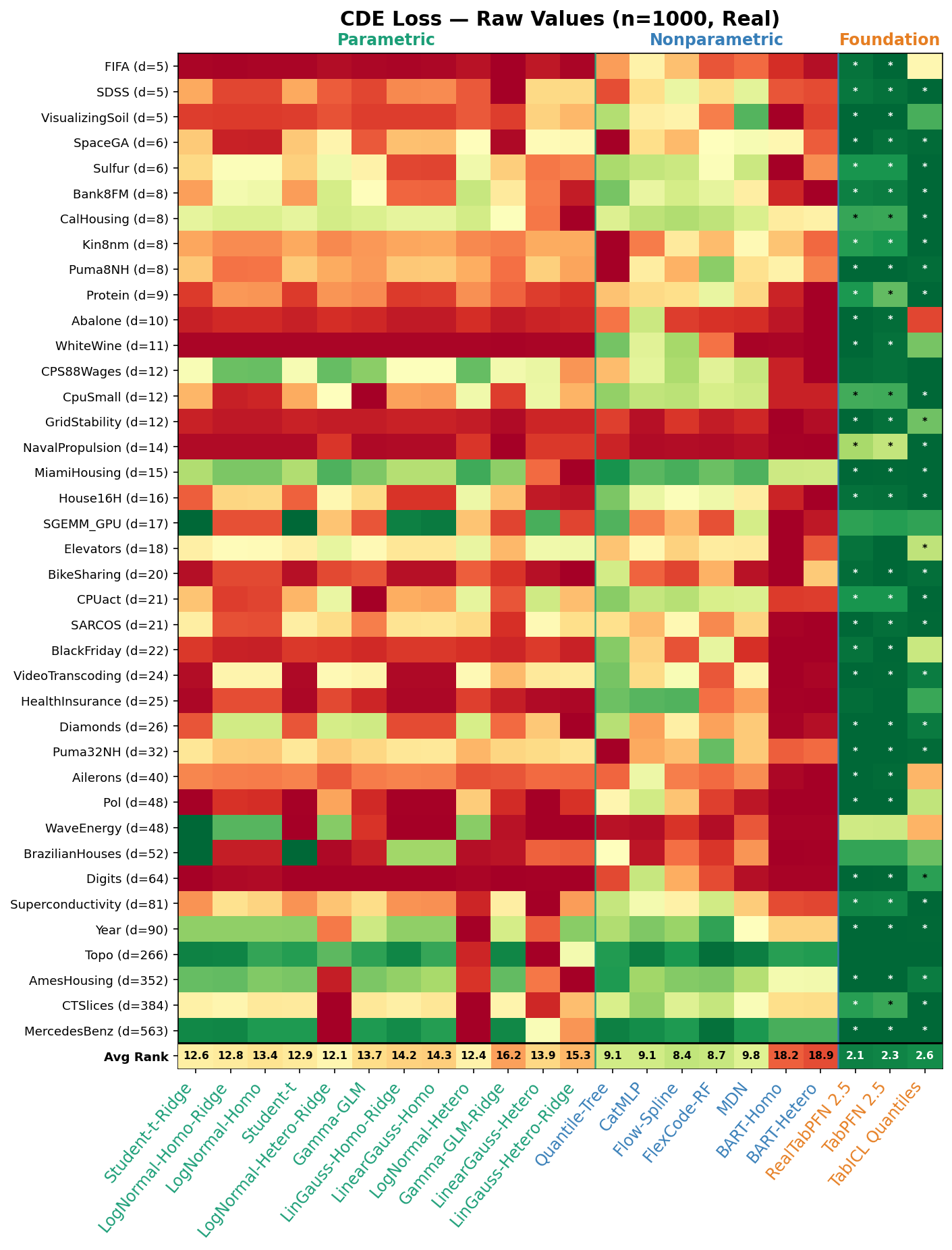}
  \caption{Raw CDE Loss -- $n=1000$, real data.}
\end{figure}
\begin{figure}[p]\centering
  \includegraphics[width=\linewidth]{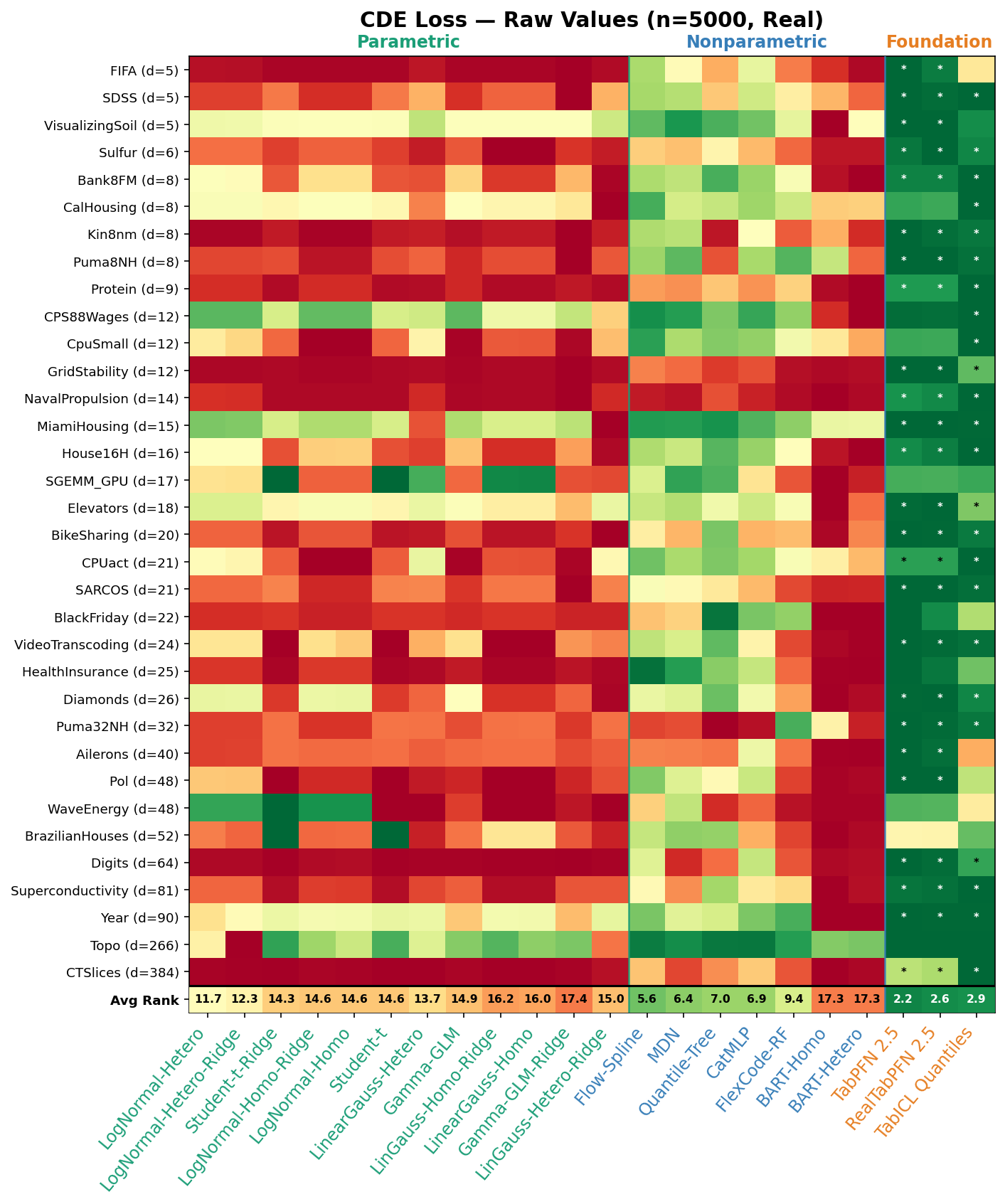}
  \caption{Raw CDE Loss -- $n=5000$, real data.}
\end{figure}
\begin{figure}[p]\centering
  \includegraphics[width=\linewidth]{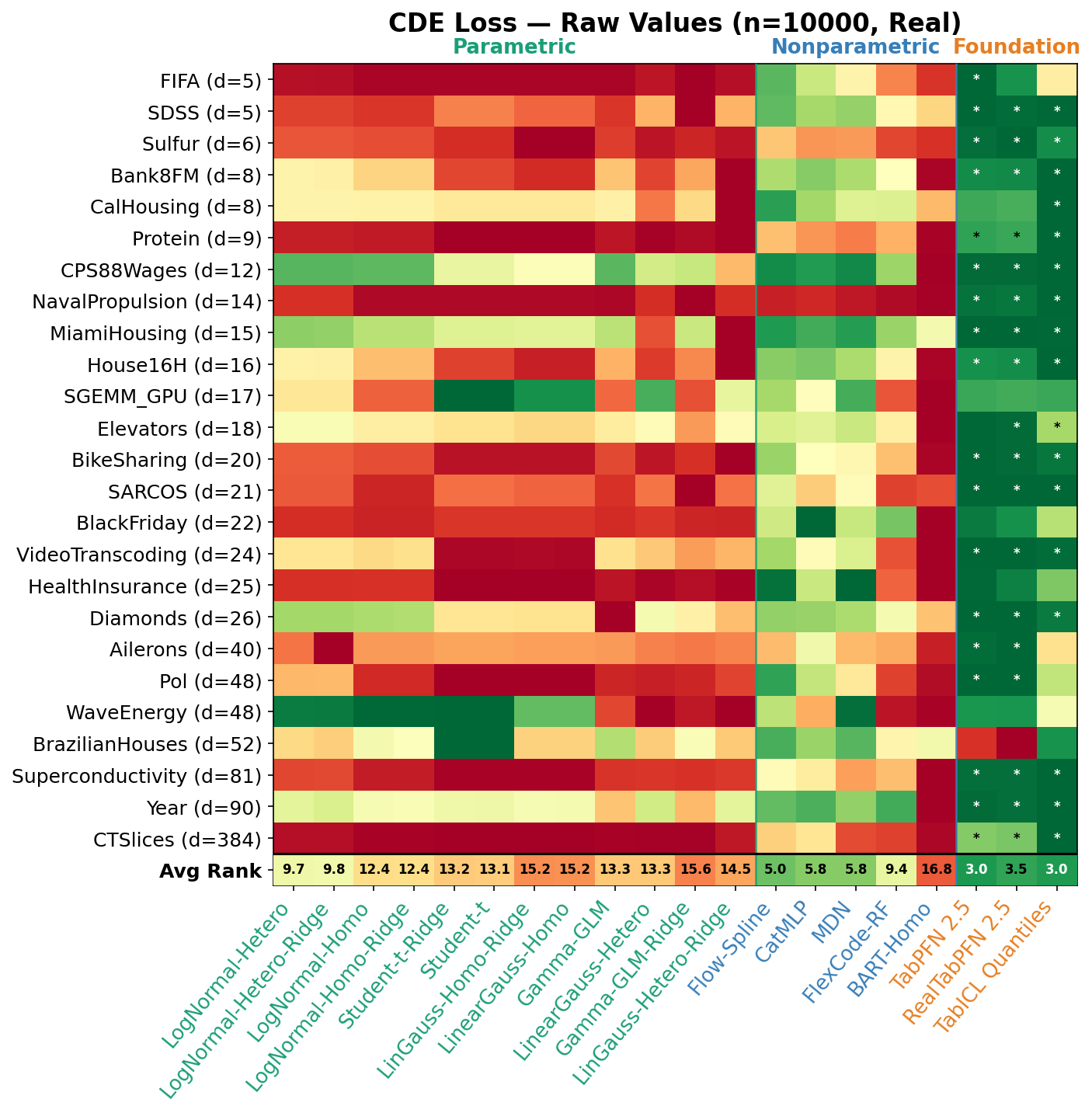}
  \caption{Raw CDE Loss -- $n=10000$, real data.}
\end{figure}
\begin{figure}[p]\centering
  \includegraphics[width=\linewidth]{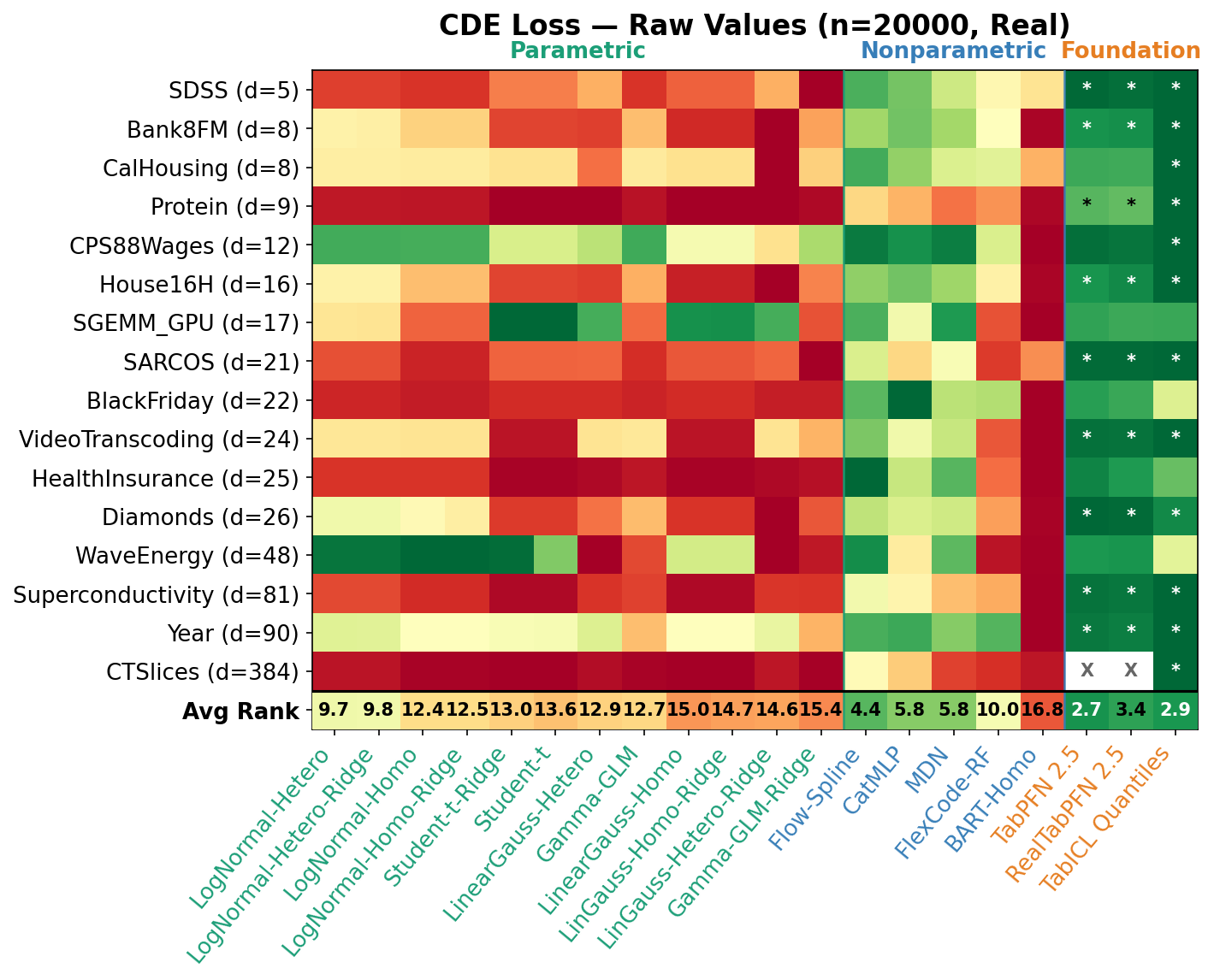}
  \caption{Raw CDE Loss -- $n=20000$, real data.}
\end{figure}

% ── Log-Likelihood ──────────────────────────────────────────────────────────

\begin{figure}[p]\centering
  \includegraphics[width=\linewidth]{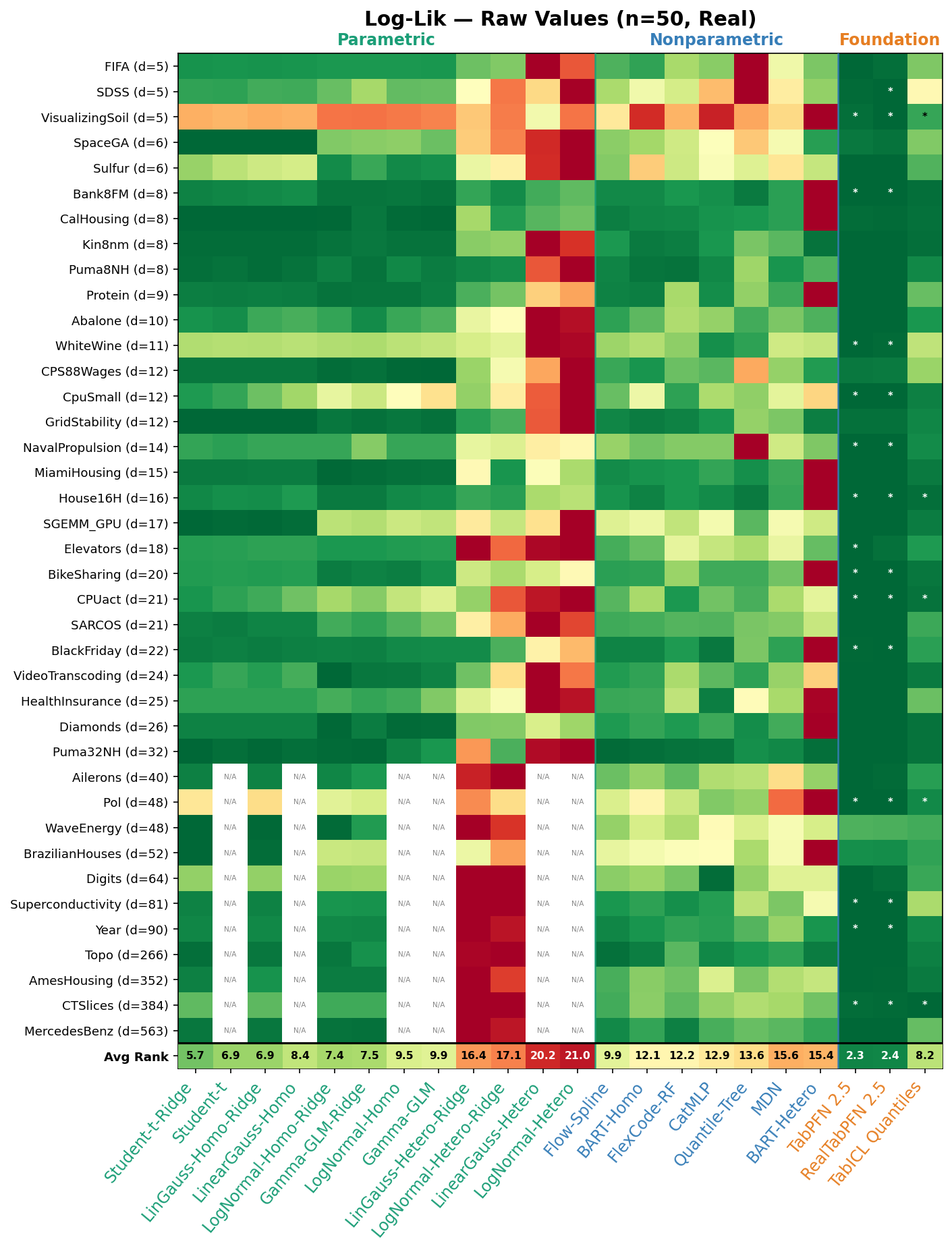}
  \caption{Raw Log-Likelihood -- $n=50$, real data.}
\end{figure}
\begin{figure}[p]\centering
  \includegraphics[width=\linewidth]{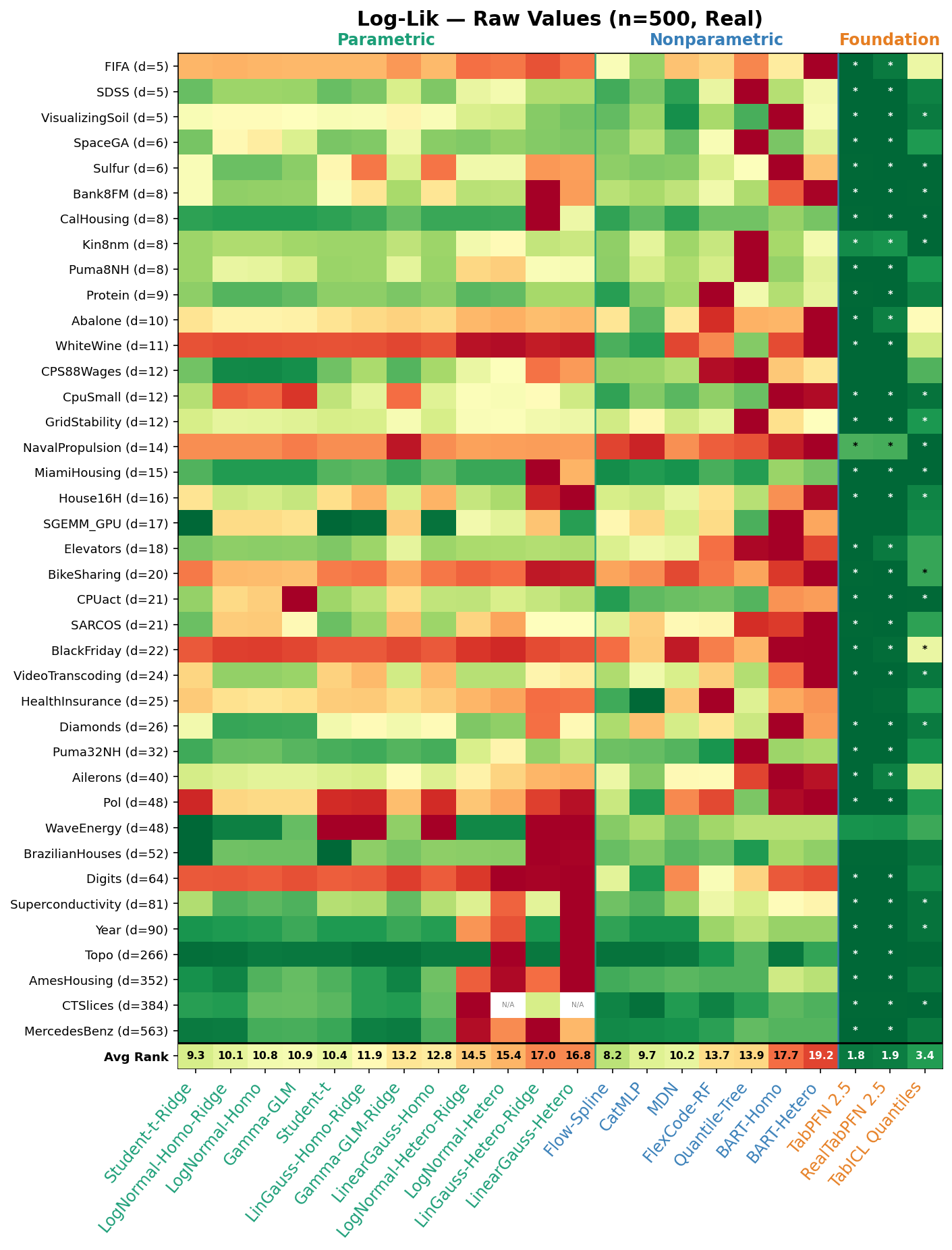}
  \caption{Raw Log-Likelihood -- $n=500$, real data.}
\end{figure}
\begin{figure}[p]\centering
  \includegraphics[width=\linewidth]{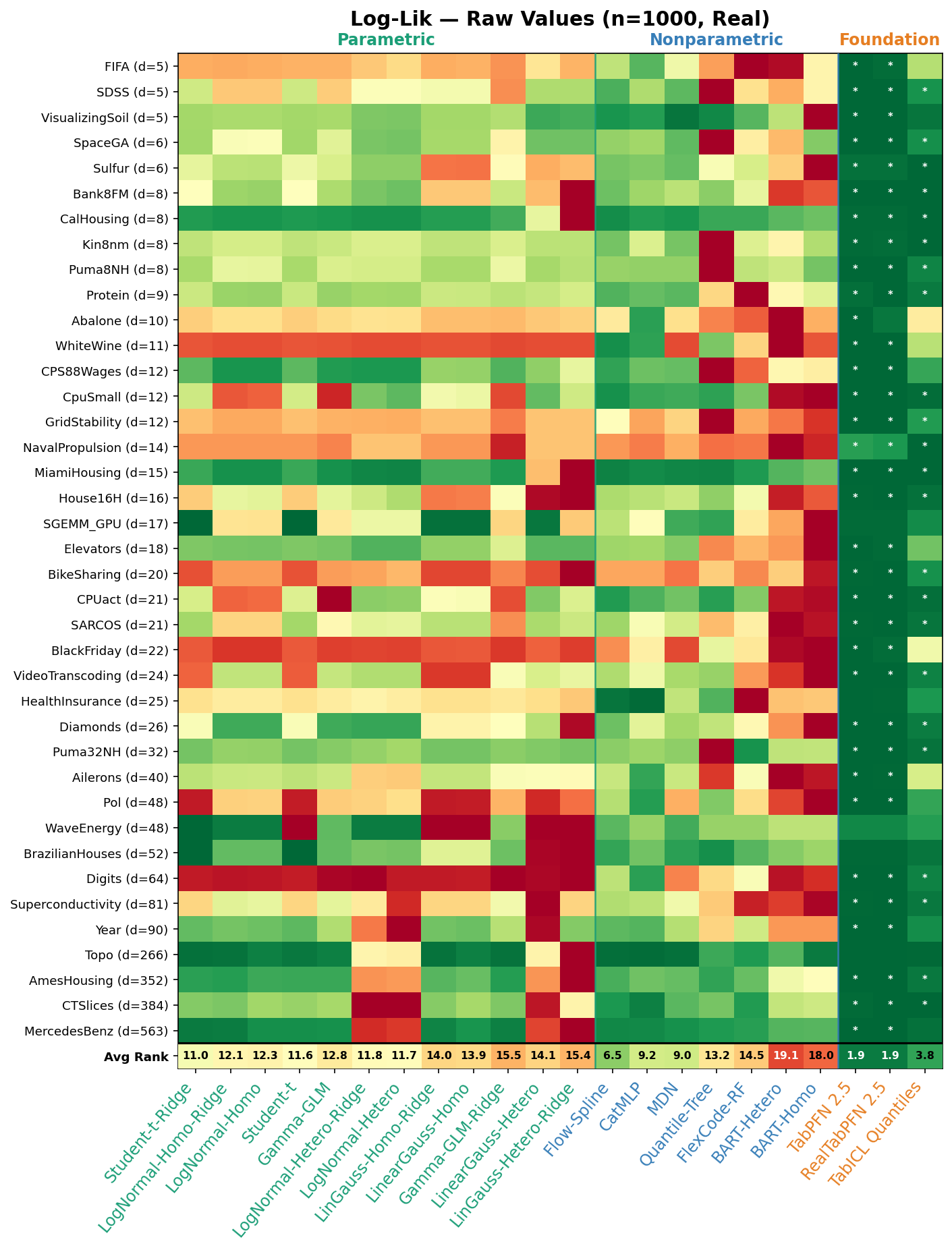}
  \caption{Raw Log-Likelihood -- $n=1000$, real data.}
\end{figure}
\begin{figure}[p]\centering
  \includegraphics[width=\linewidth]{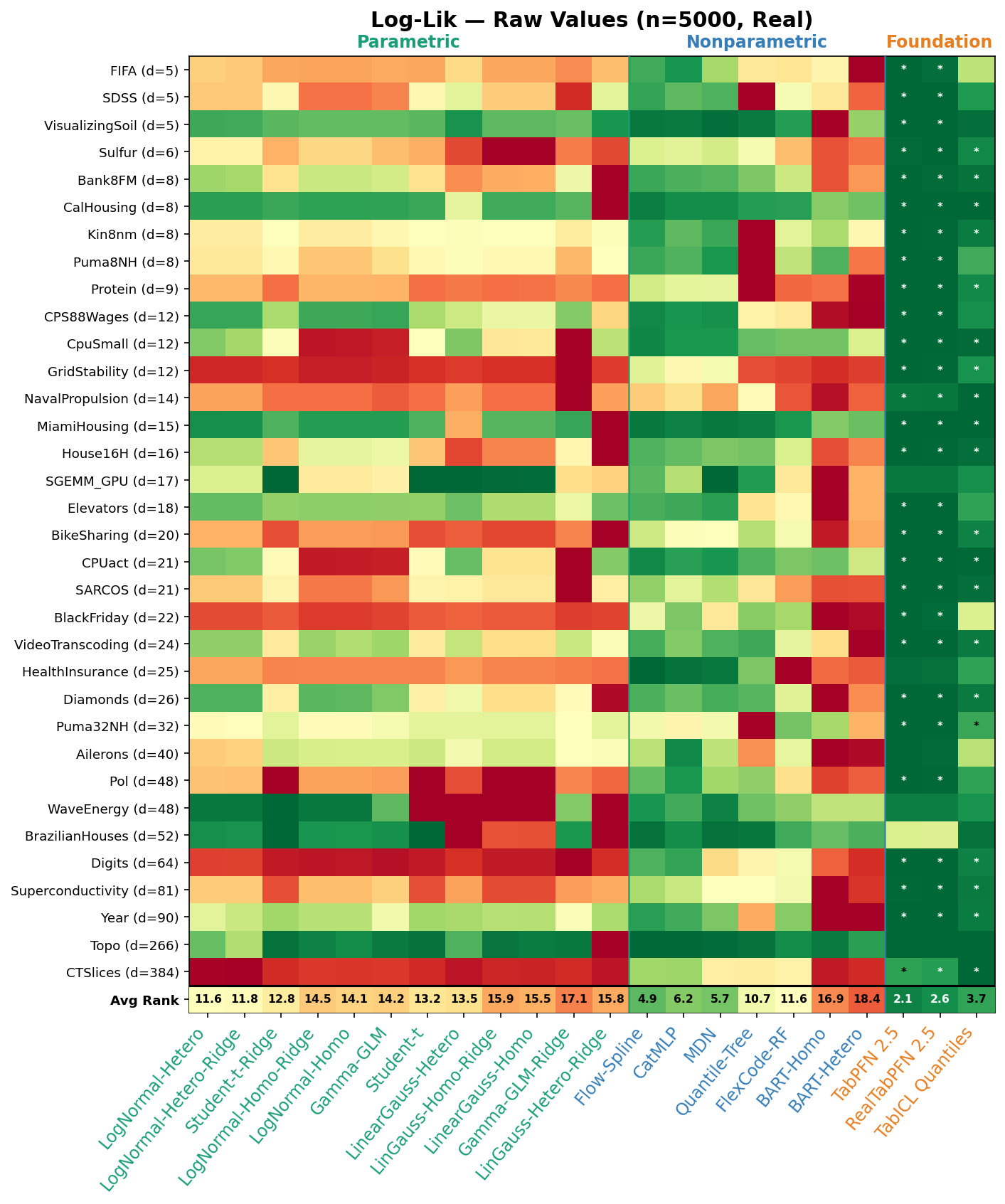}
  \caption{Raw Log-Likelihood -- $n=5000$, real data.}
\end{figure}
\begin{figure}[p]\centering
  \includegraphics[width=\linewidth]{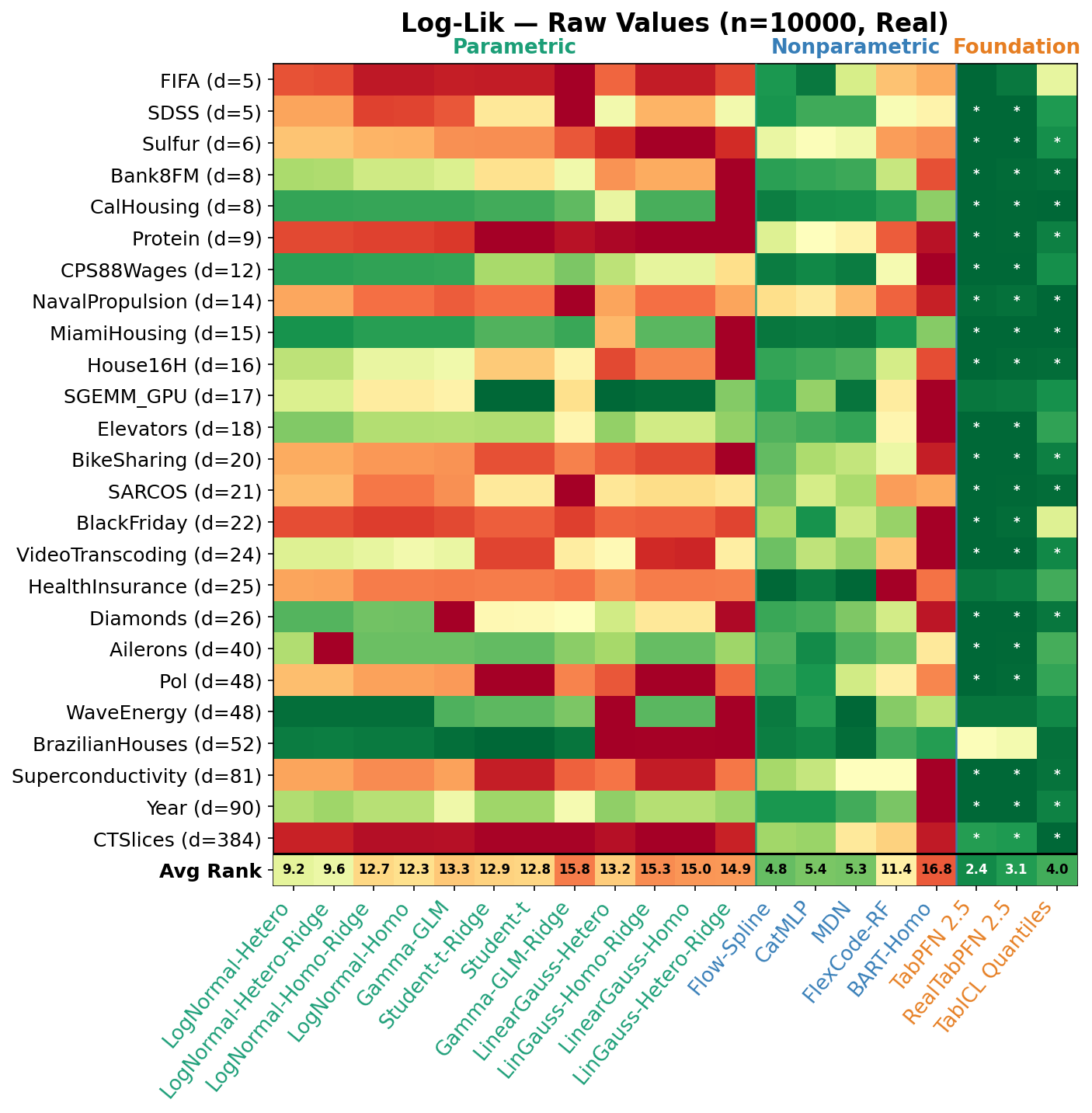}
  \caption{Raw Log-Likelihood -- $n=10000$, real data.}
\end{figure}
\begin{figure}[p]\centering
  \includegraphics[width=\linewidth]{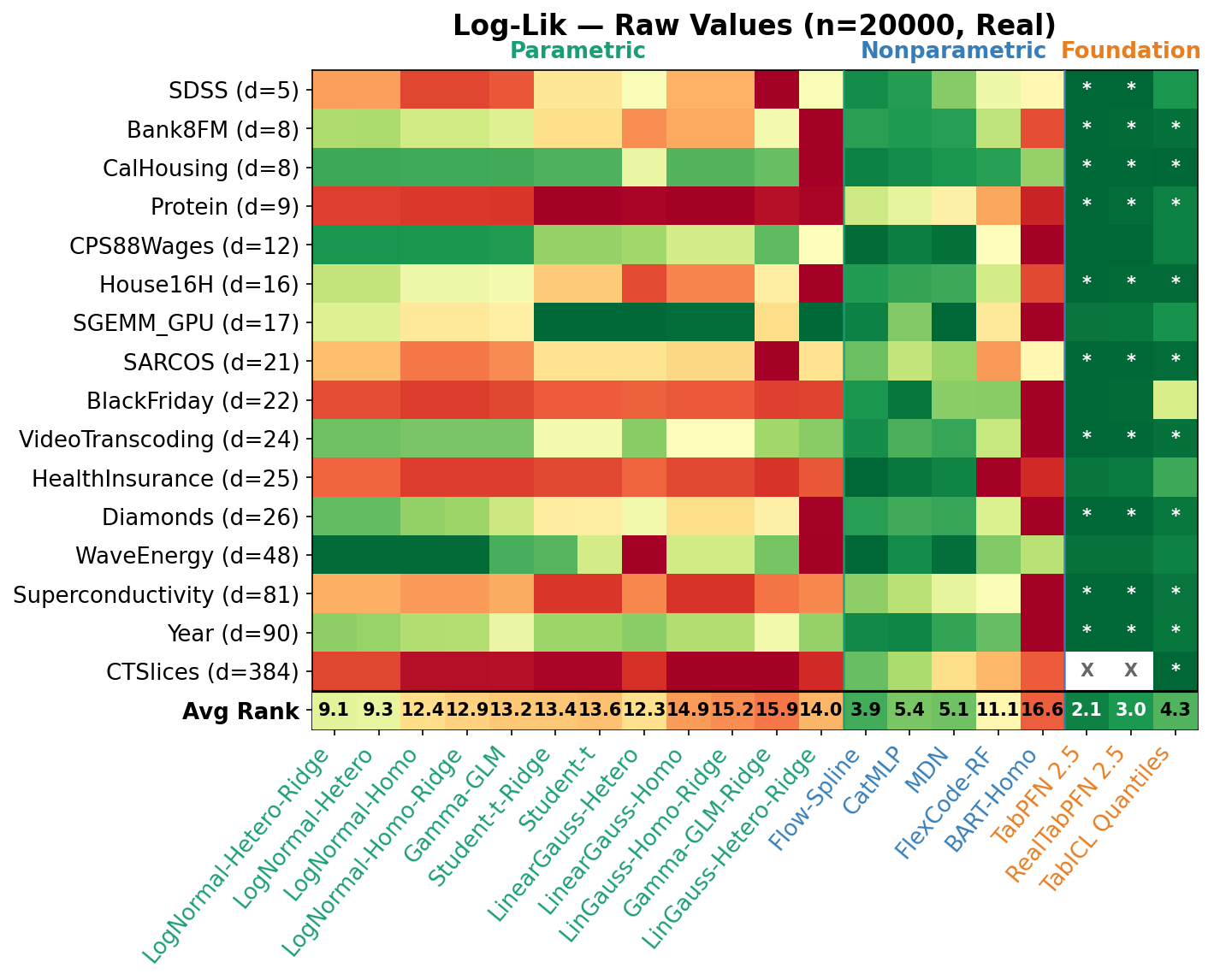}
  \caption{Raw Log-Likelihood -- $n=20000$, real data.}
\end{figure}

% ── CRPS ────────────────────────────────────────────────────────────────────

\begin{figure}[p]\centering
  \includegraphics[width=\linewidth]{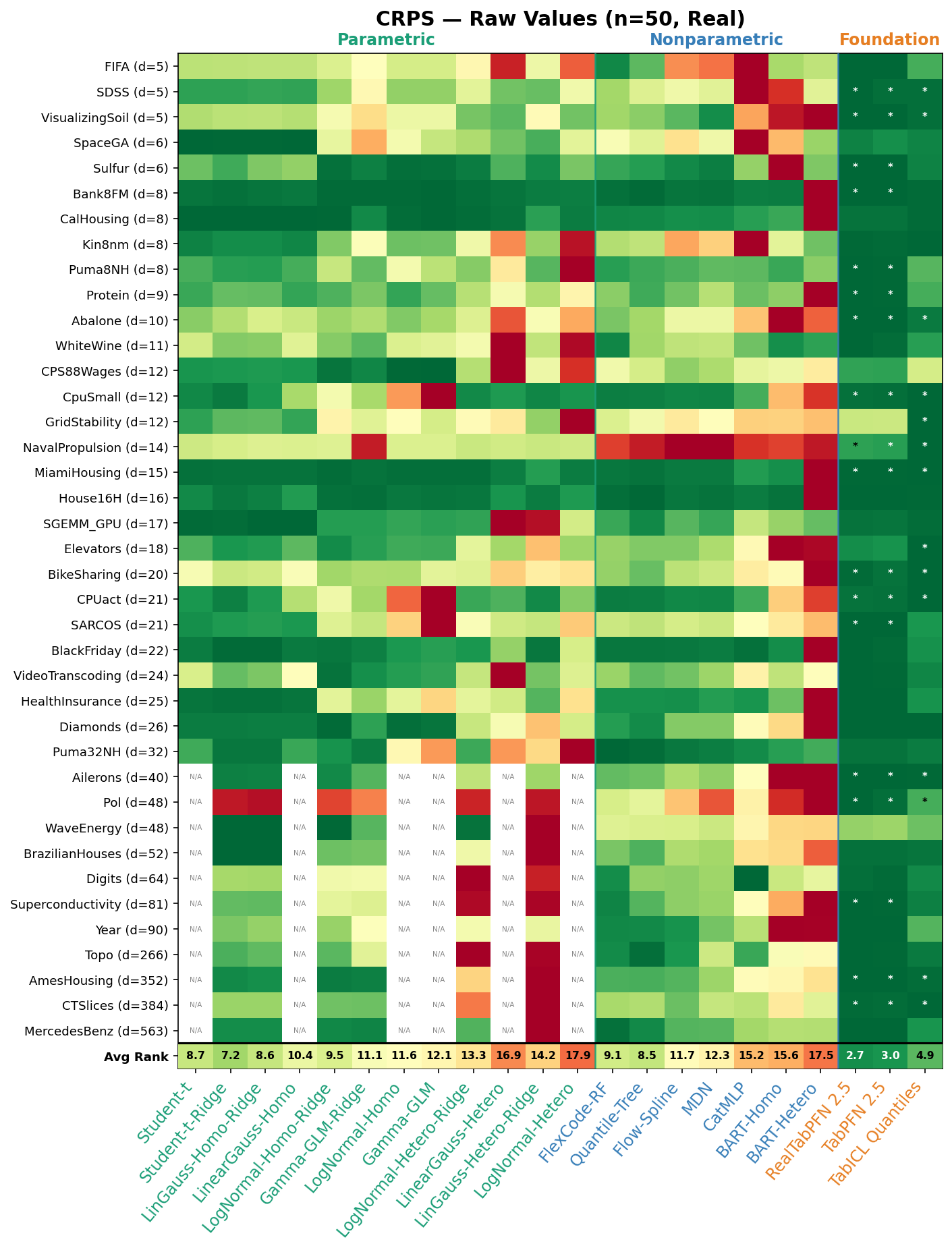}
  \caption{Raw CRPS -- $n=50$, real data.}
\end{figure}
\begin{figure}[p]\centering
  \includegraphics[width=\linewidth]{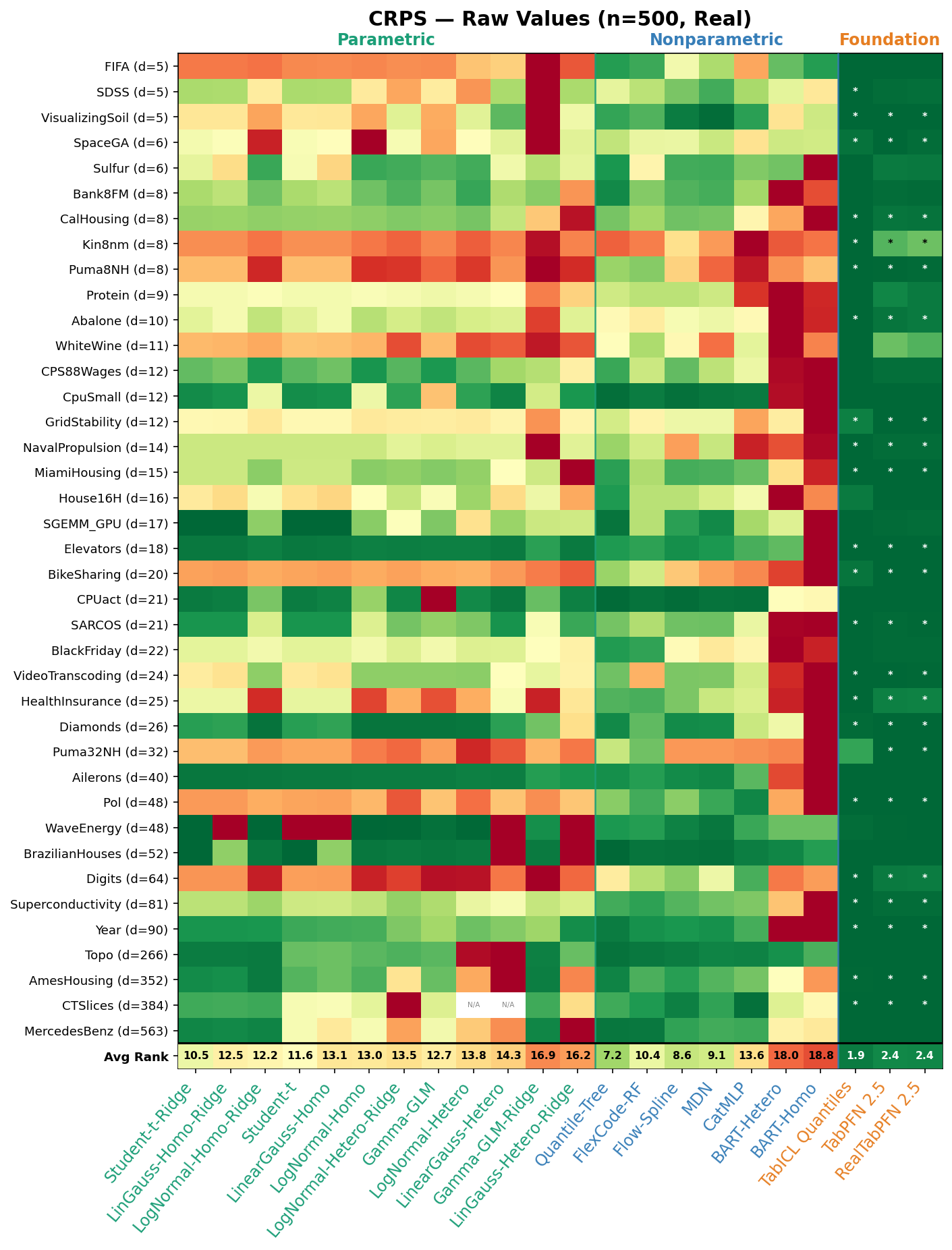}
  \caption{Raw CRPS -- $n=500$, real data.}
\end{figure}
\begin{figure}[p]\centering
  \includegraphics[width=\linewidth]{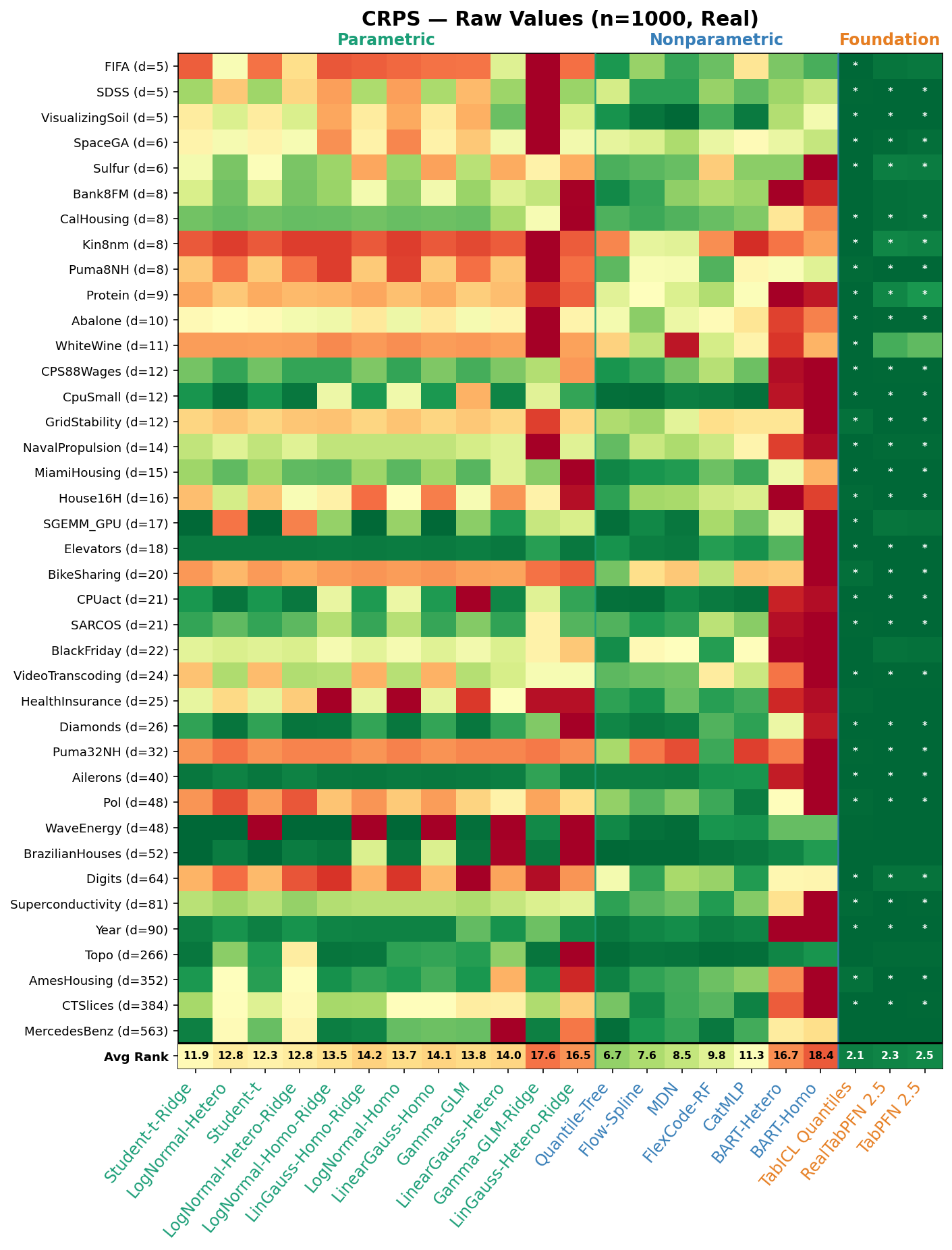}
  \caption{Raw CRPS -- $n=1000$, real data.}
\end{figure}
\begin{figure}[p]\centering
  \includegraphics[width=\linewidth]{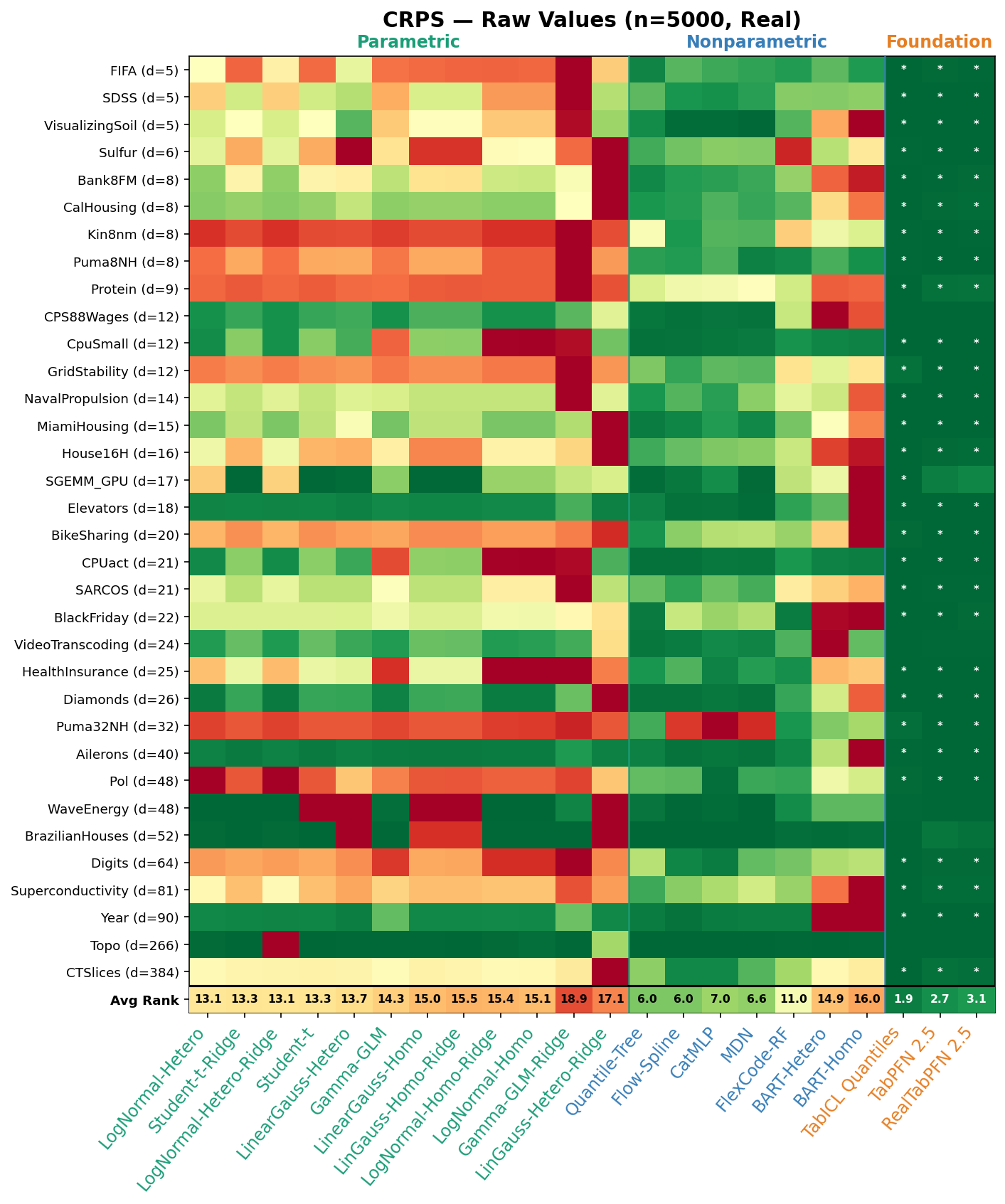}
  \caption{Raw CRPS -- $n=5000$, real data.}
\end{figure}
\begin{figure}[p]\centering
  \includegraphics[width=\linewidth]{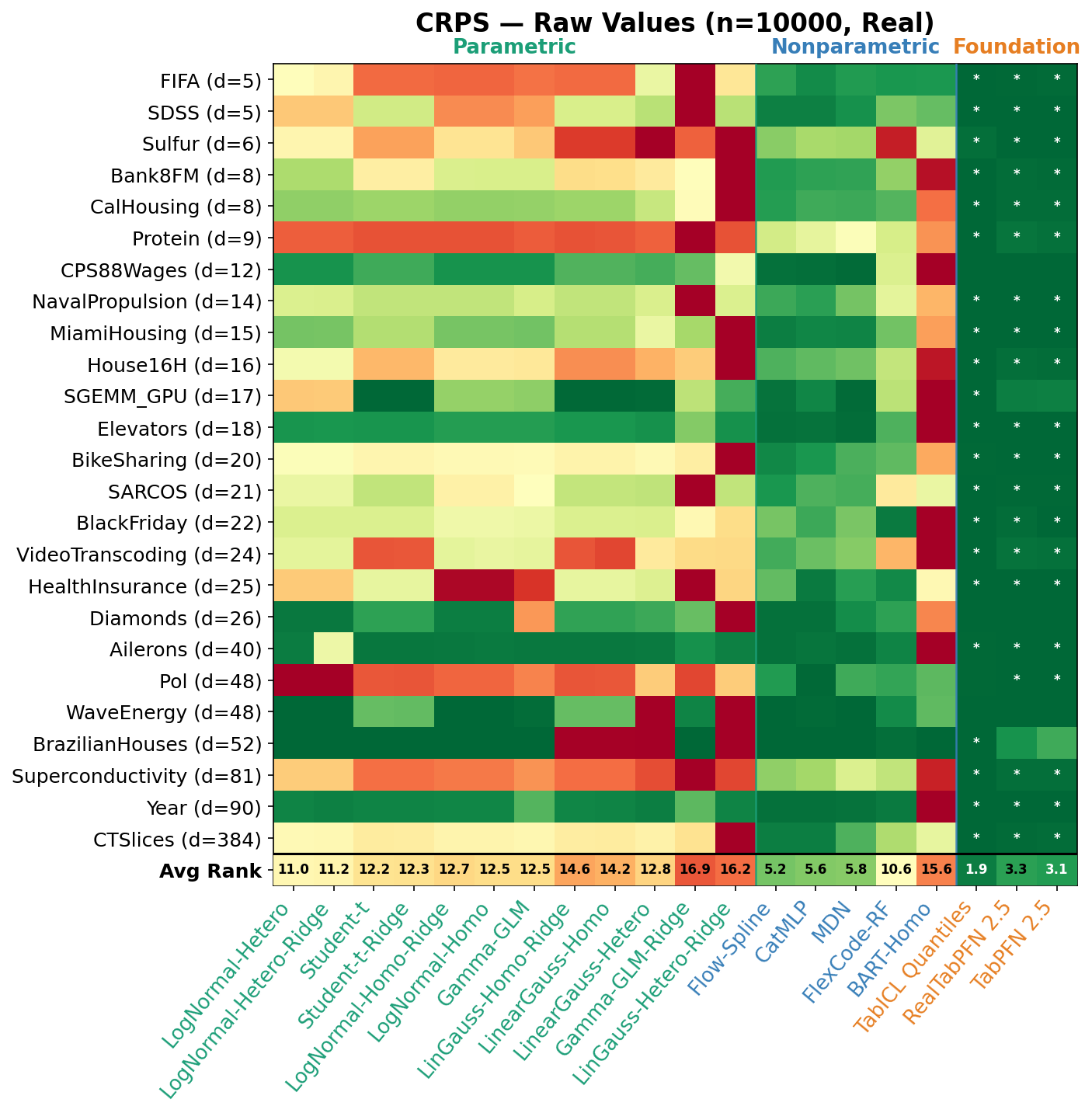}
  \caption{Raw CRPS -- $n=10000$, real data.}
\end{figure}
\begin{figure}[p]\centering
  \includegraphics[width=\linewidth]{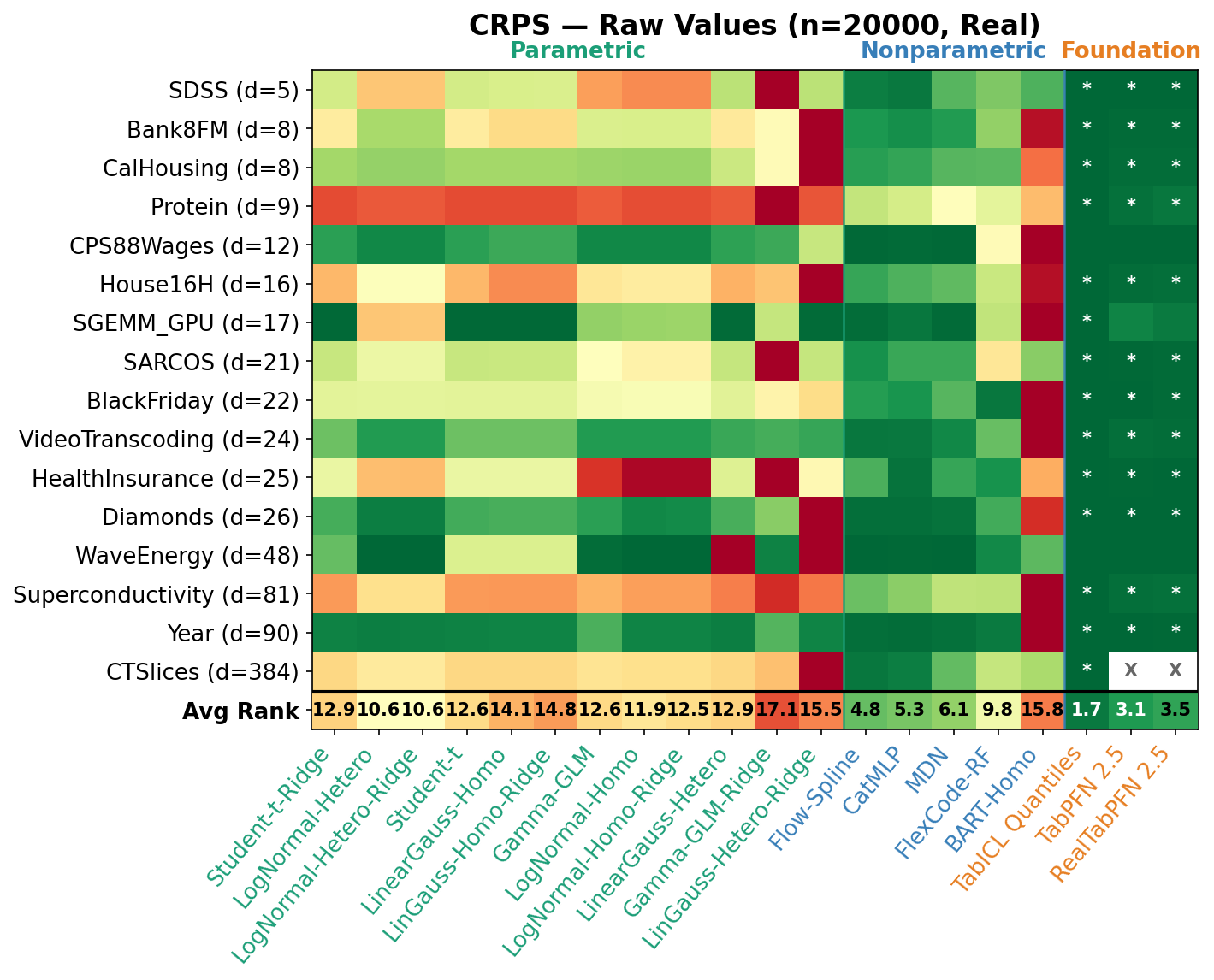}
  \caption{Raw CRPS -- $n=20000$, real data.}
\end{figure}

% ── PIT KS ──────────────────────────────────────────────────────────────────

\begin{figure}[p]\centering
  \includegraphics[width=\linewidth]{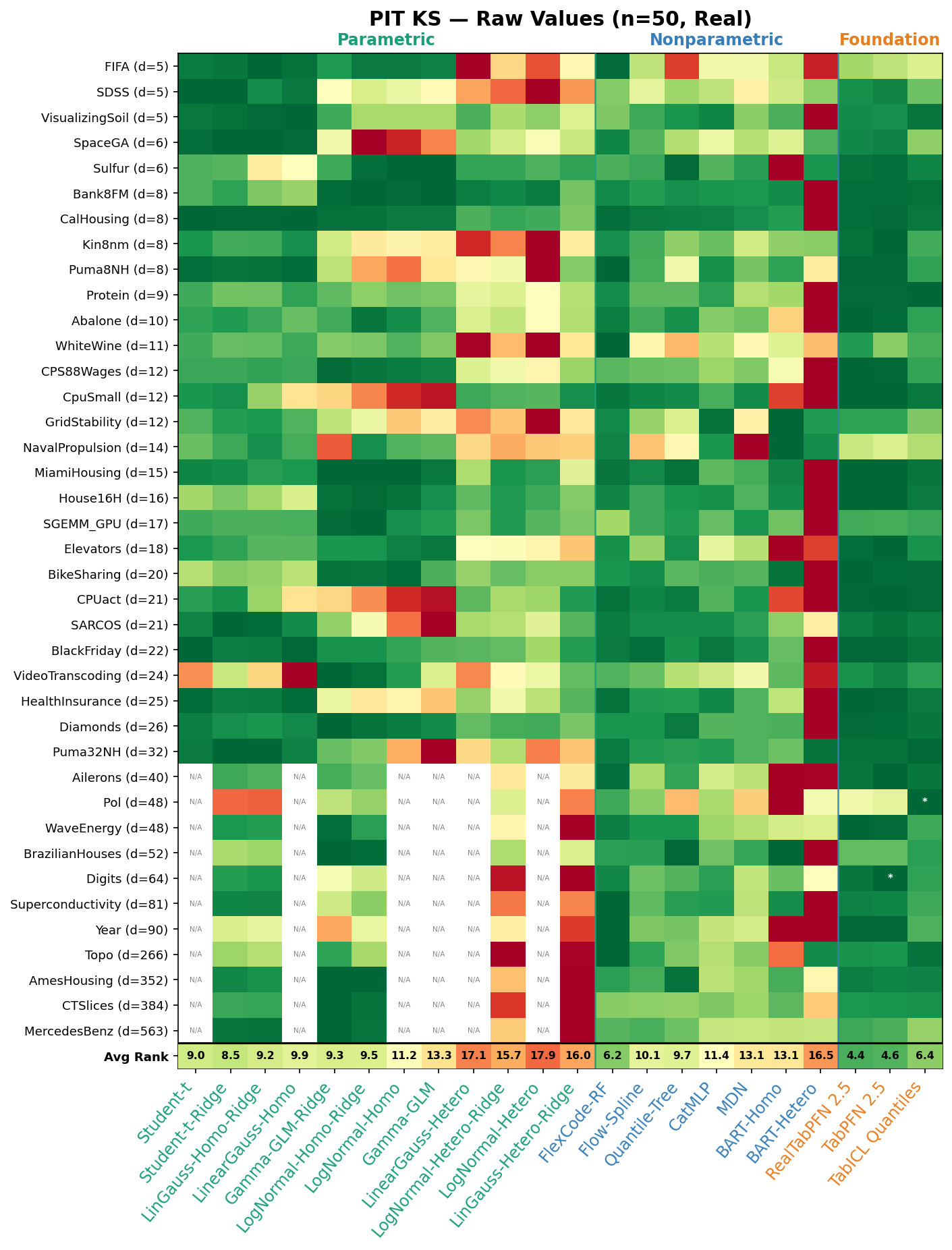}
  \caption{Raw PIT KS -- $n=50$, real data.}
\end{figure}
\begin{figure}[p]\centering
  \includegraphics[width=\linewidth]{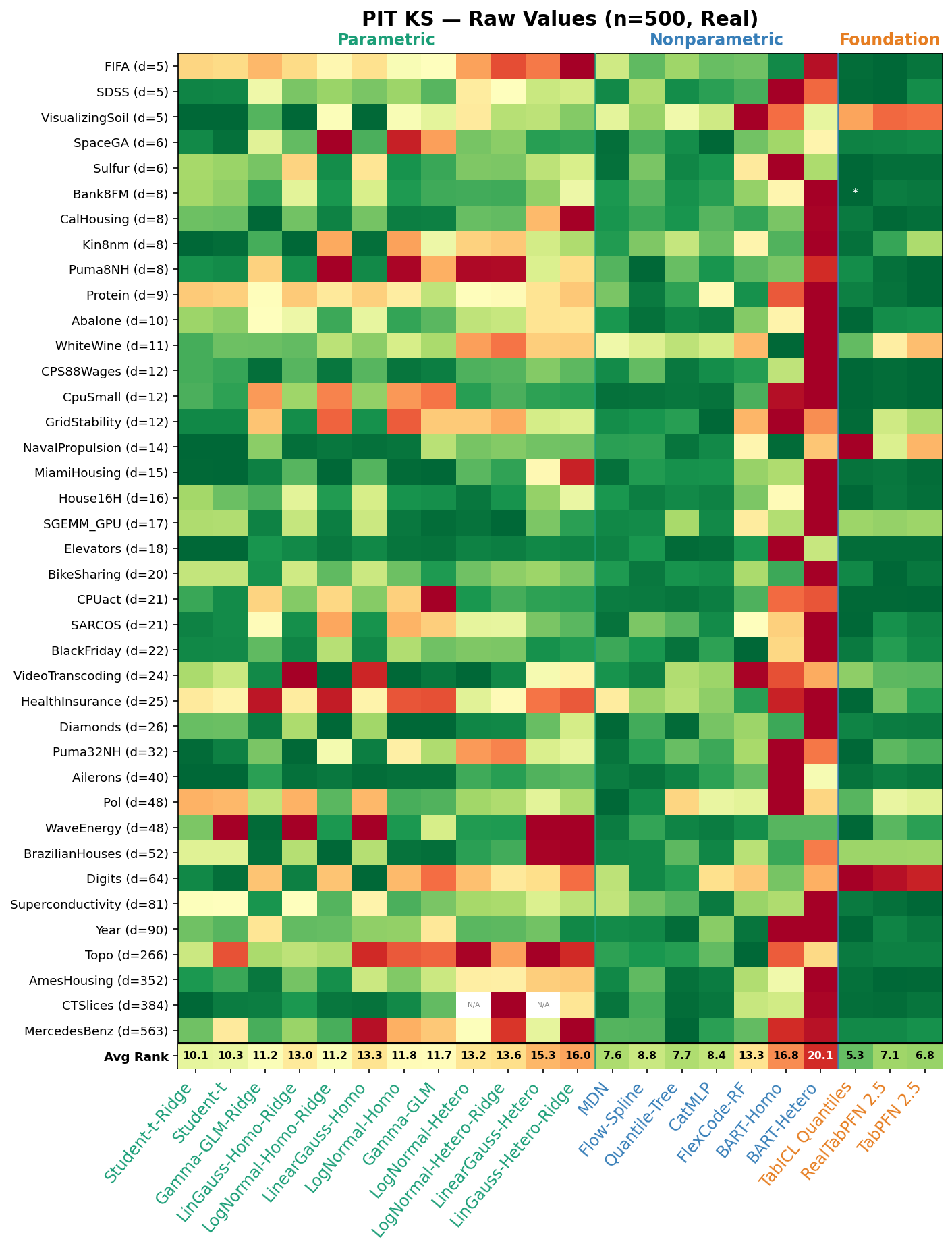}
  \caption{Raw PIT KS -- $n=500$, real data.}
\end{figure}
\begin{figure}[p]\centering
  \includegraphics[width=\linewidth]{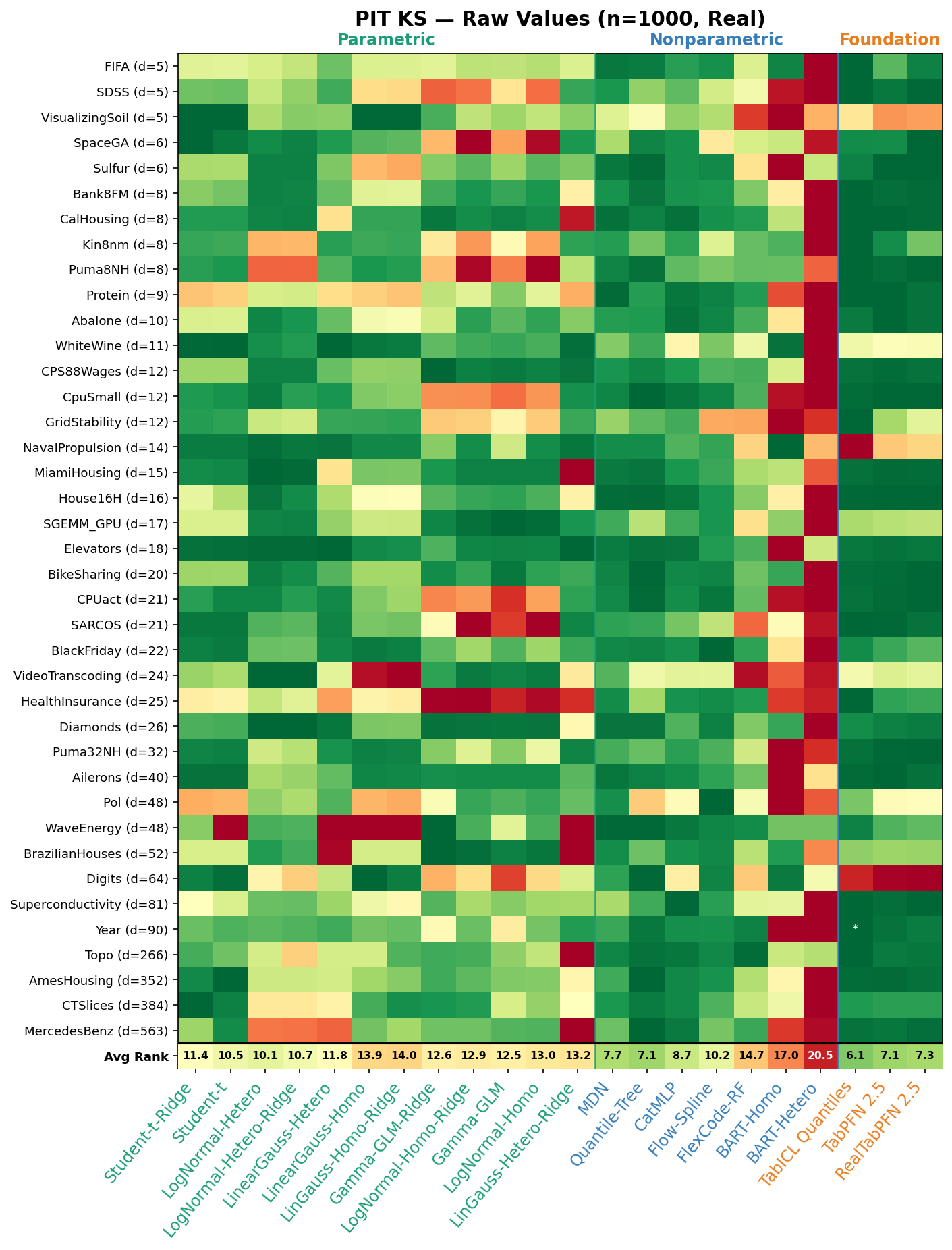}
  \caption{Raw PIT KS -- $n=1000$, real data.}
\end{figure}
\begin{figure}[p]\centering
  \includegraphics[width=\linewidth]{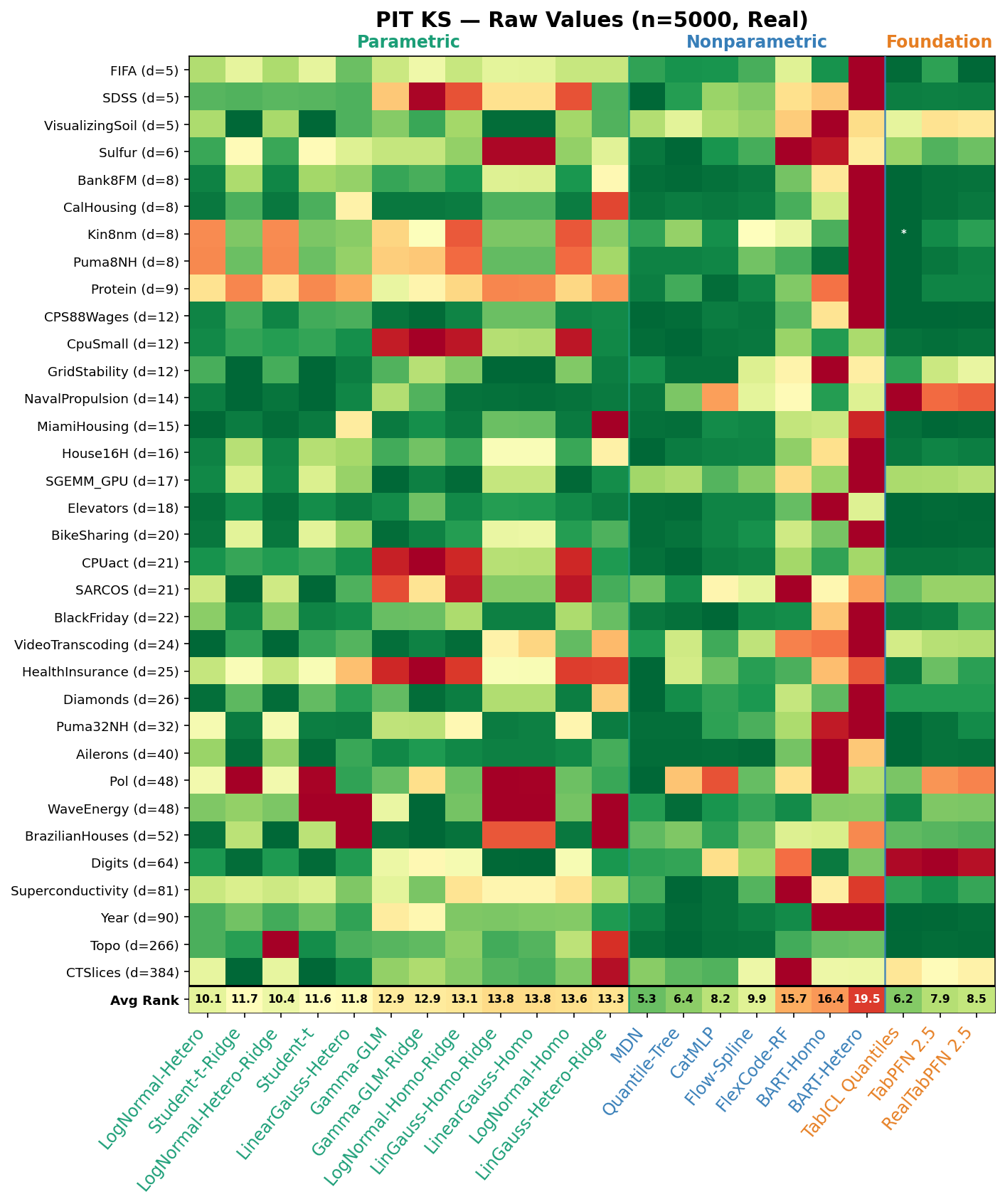}
  \caption{Raw PIT KS -- $n=5000$, real data.}
\end{figure}
\begin{figure}[p]\centering
  \includegraphics[width=\linewidth]{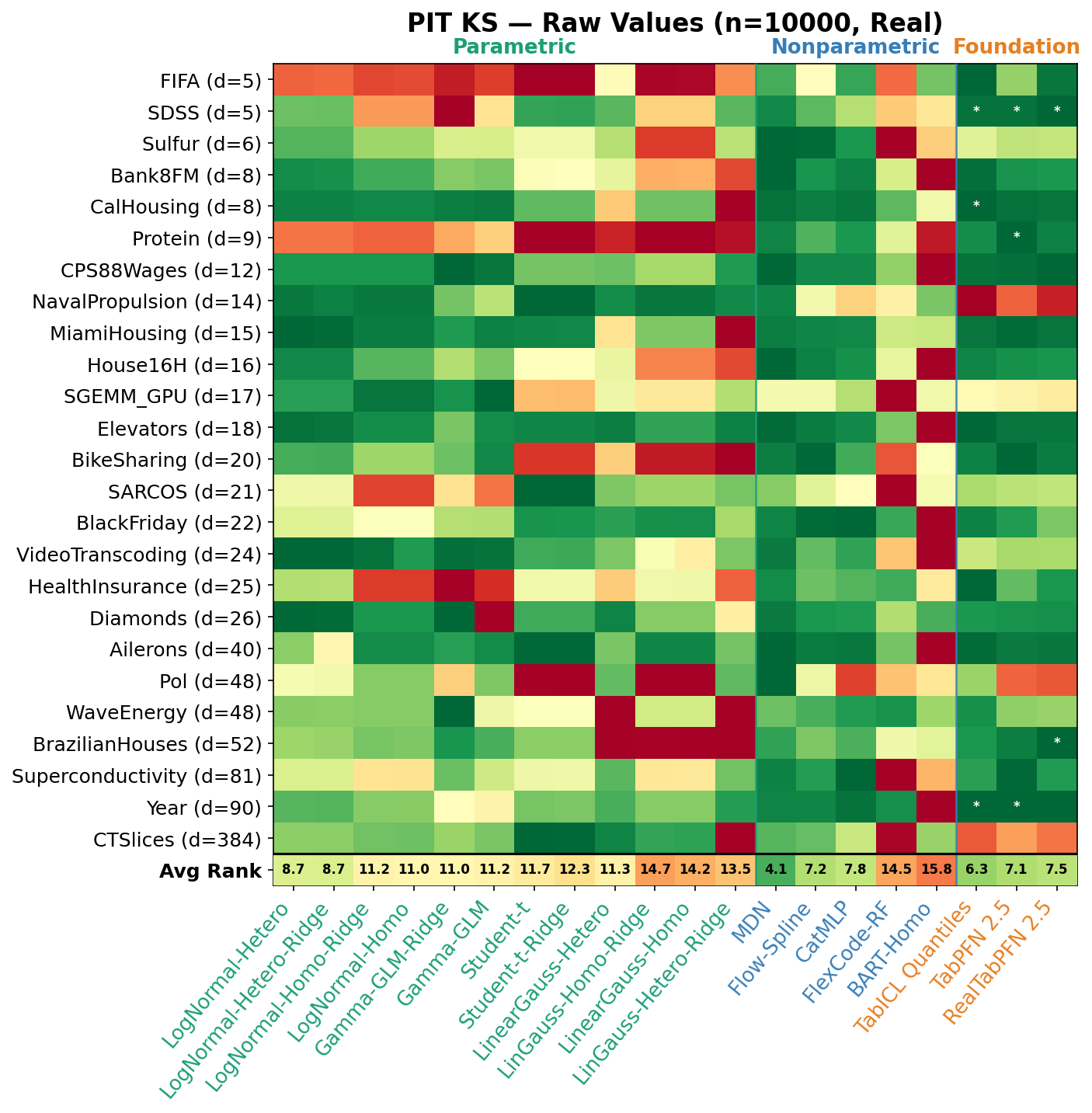}
  \caption{Raw PIT KS -- $n=10000$, real data.}
\end{figure}
\begin{figure}[p]\centering
  \includegraphics[width=\linewidth]{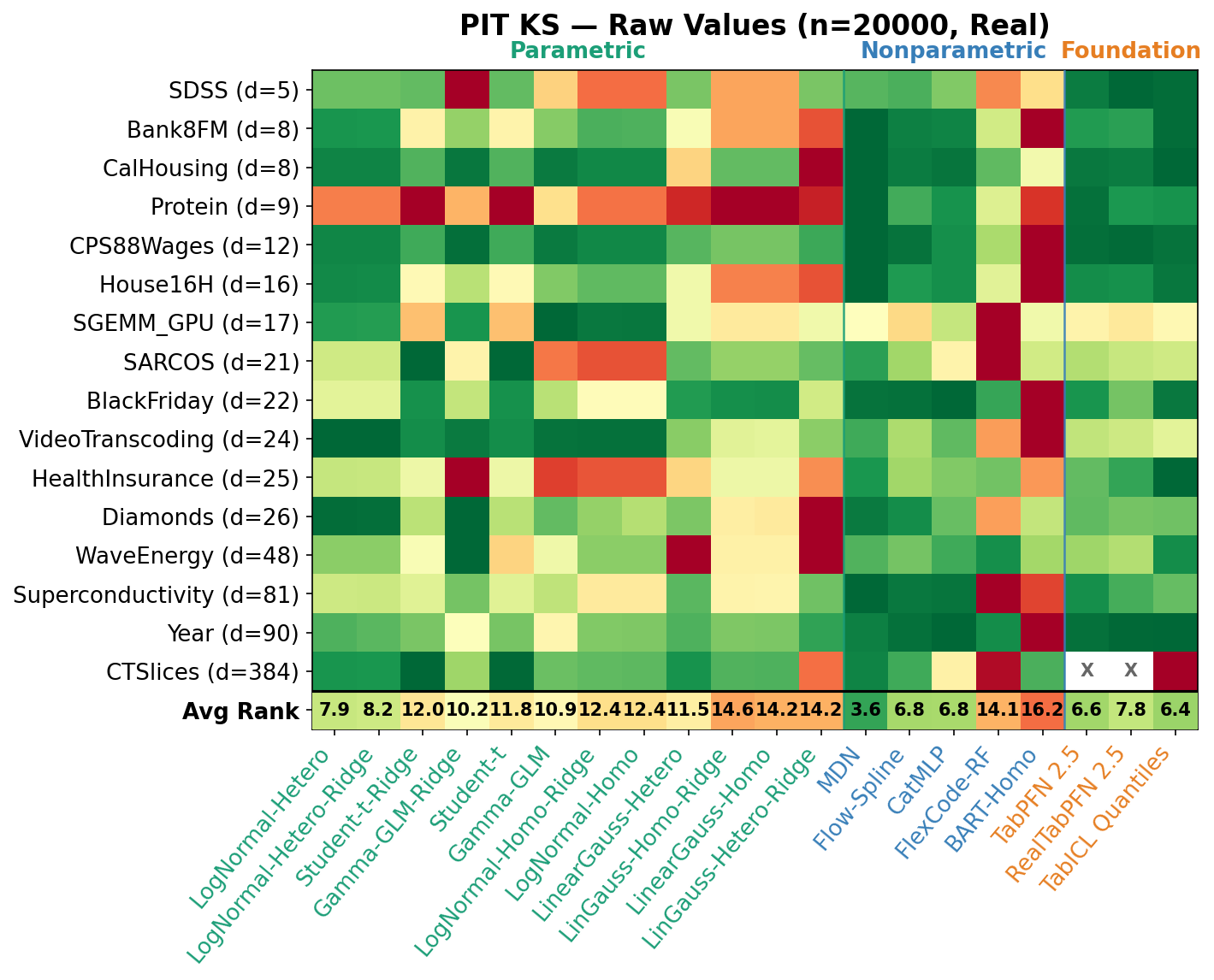}
  \caption{Raw PIT KS -- $n=20000$, real data.}
\end{figure}

% ── 90\% Coverage ───────────────────────────────────────────────────────────

\begin{figure}[p]\centering
  \includegraphics[width=\linewidth]{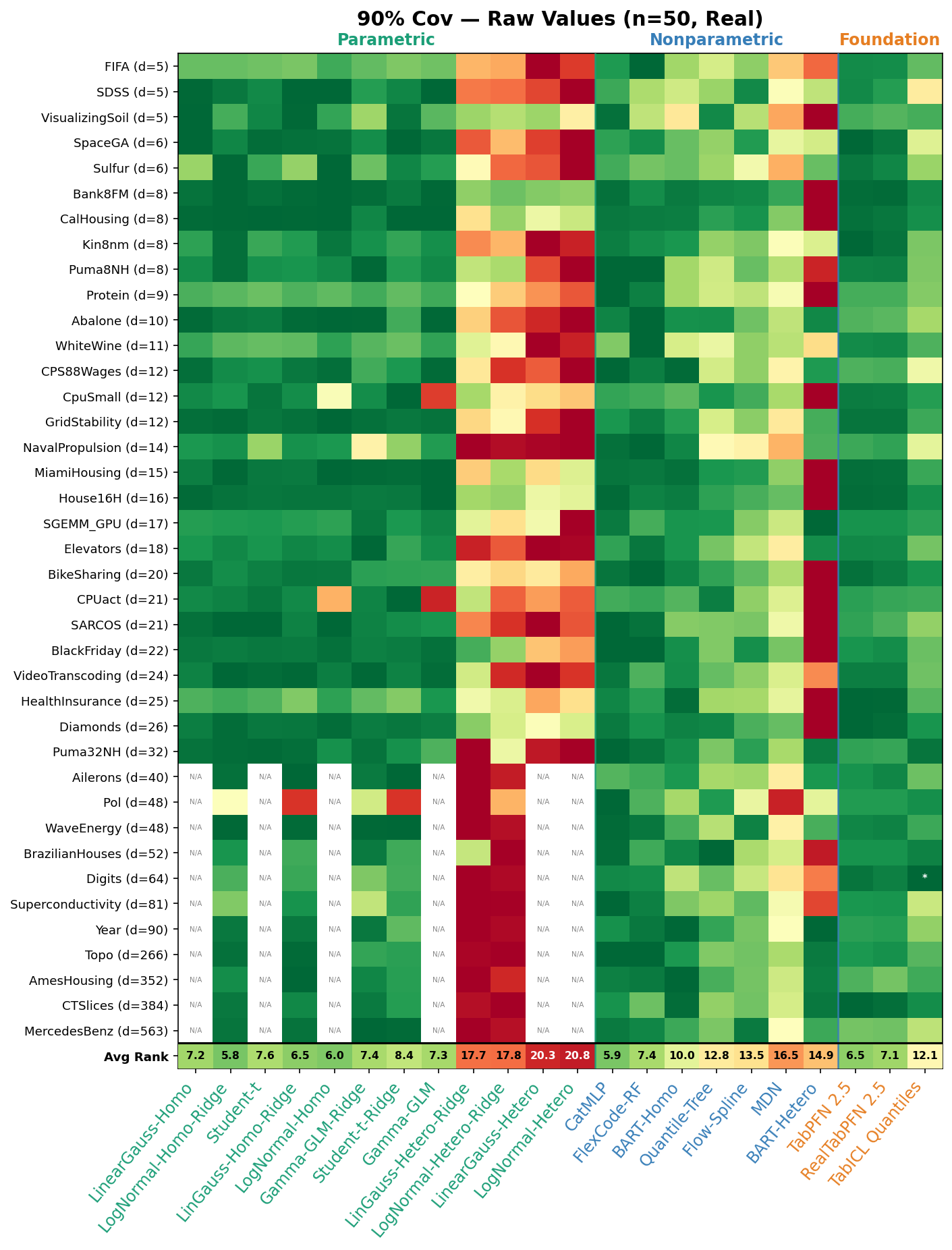}
  \caption{Raw 90\% Coverage -- $n=50$, real data.}
\end{figure}
\begin{figure}[p]\centering
  \includegraphics[width=\linewidth]{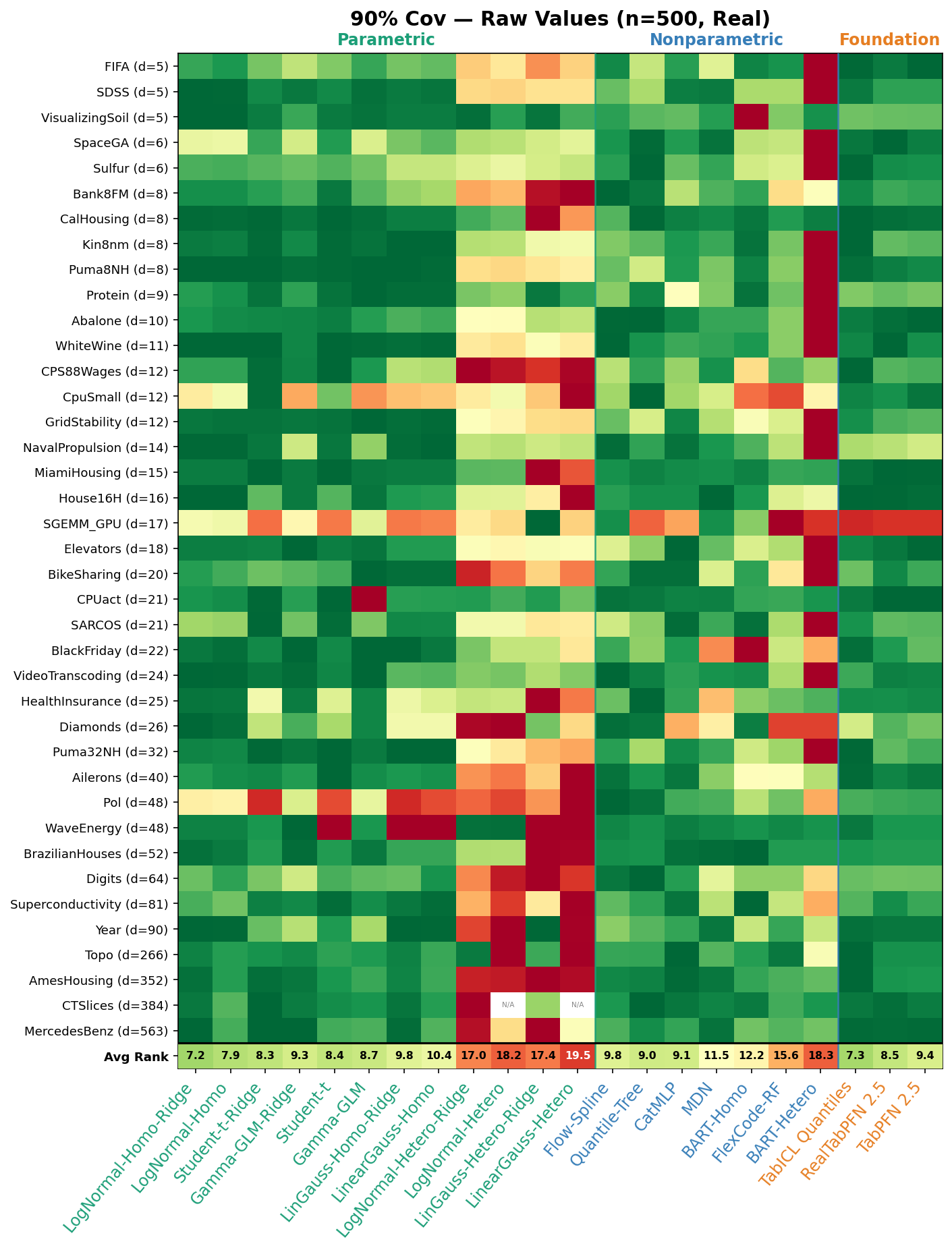}
  \caption{Raw 90\% Coverage -- $n=500$, real data.}
\end{figure}
\begin{figure}[p]\centering
  \includegraphics[width=\linewidth]{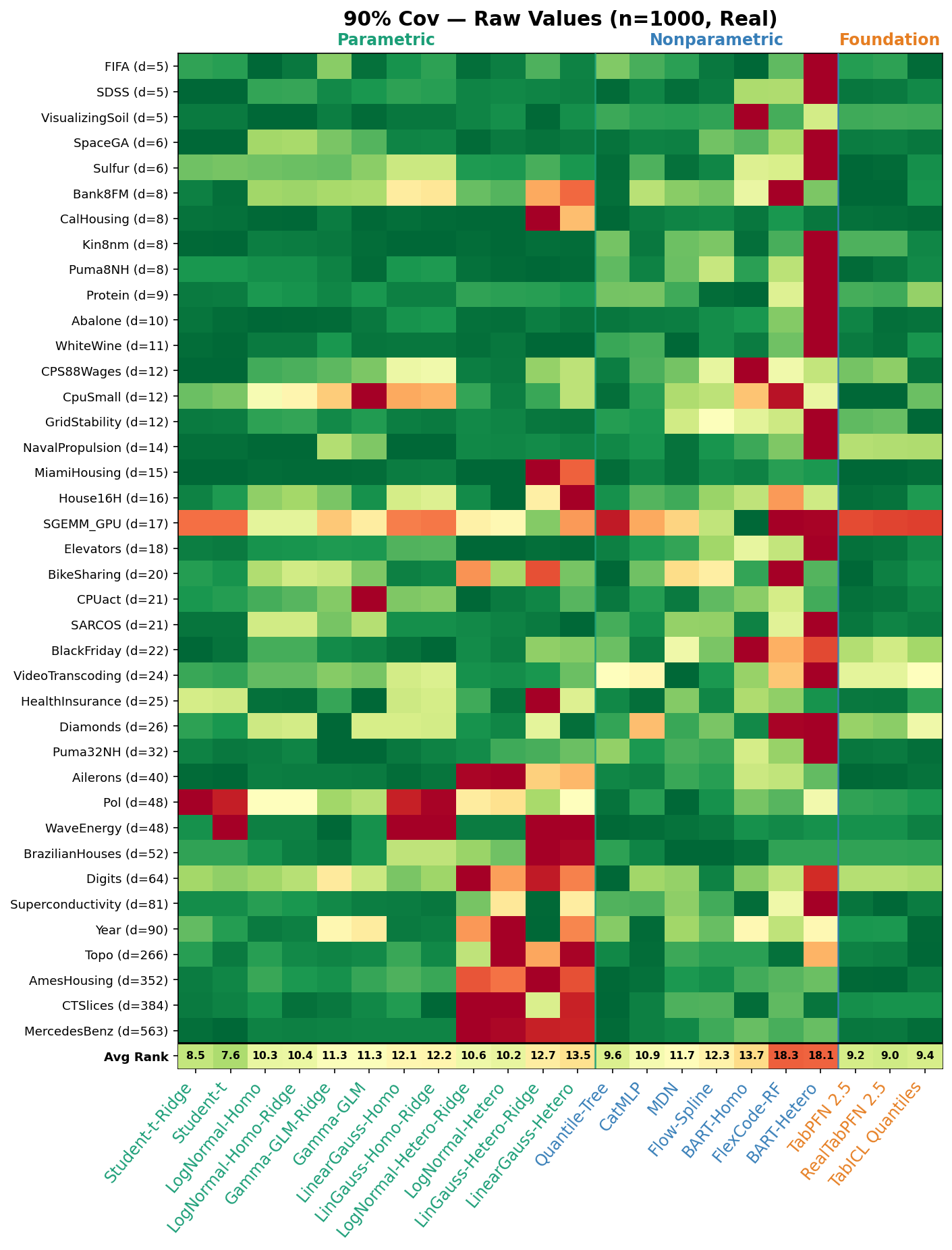}
  \caption{Raw 90\% Coverage -- $n=1000$, real data.}
\end{figure}
\begin{figure}[p]\centering
  \includegraphics[width=\linewidth]{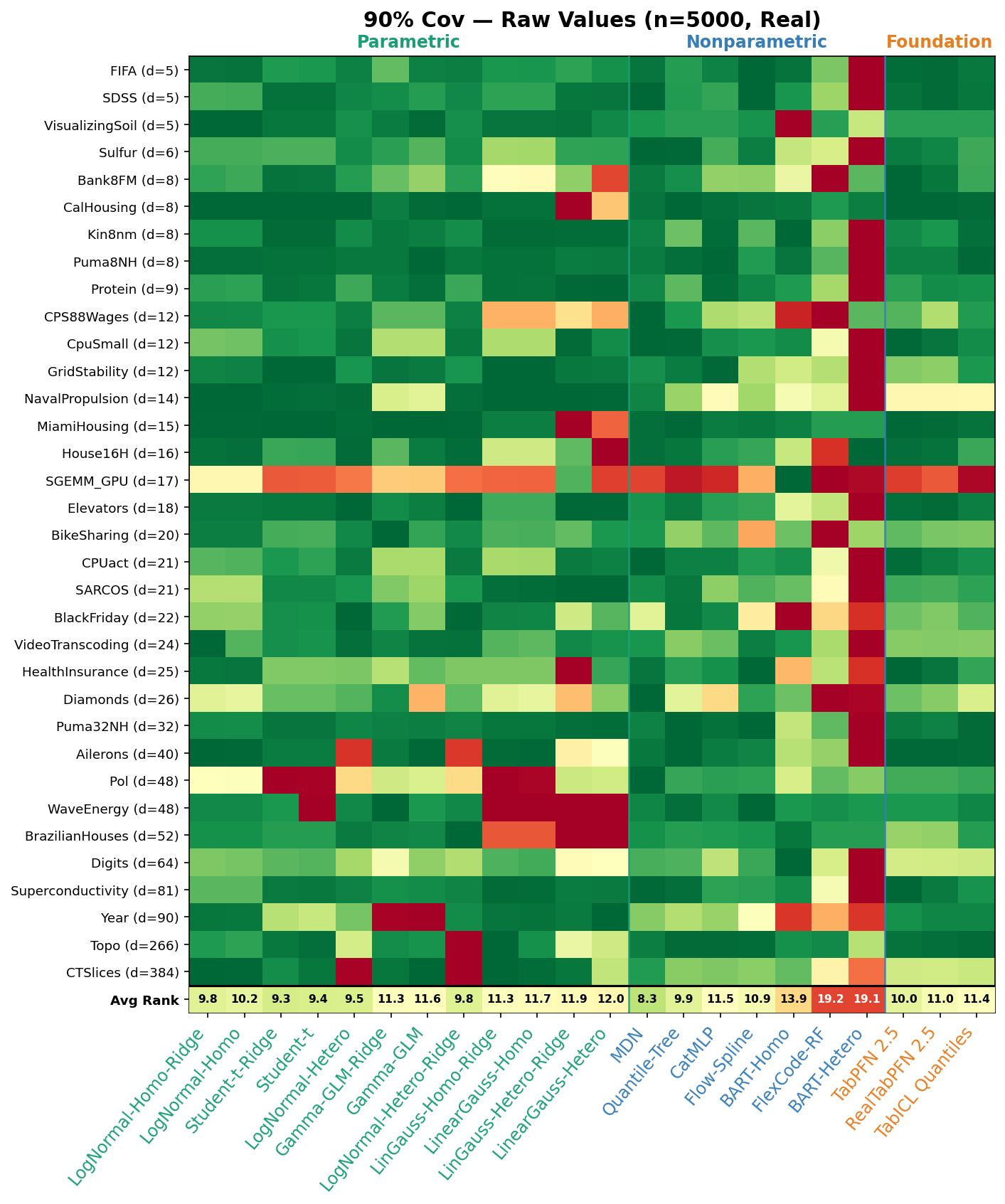}
  \caption{Raw 90\% Coverage -- $n=5000$, real data.}
\end{figure}
\begin{figure}[p]\centering
  \includegraphics[width=\linewidth]{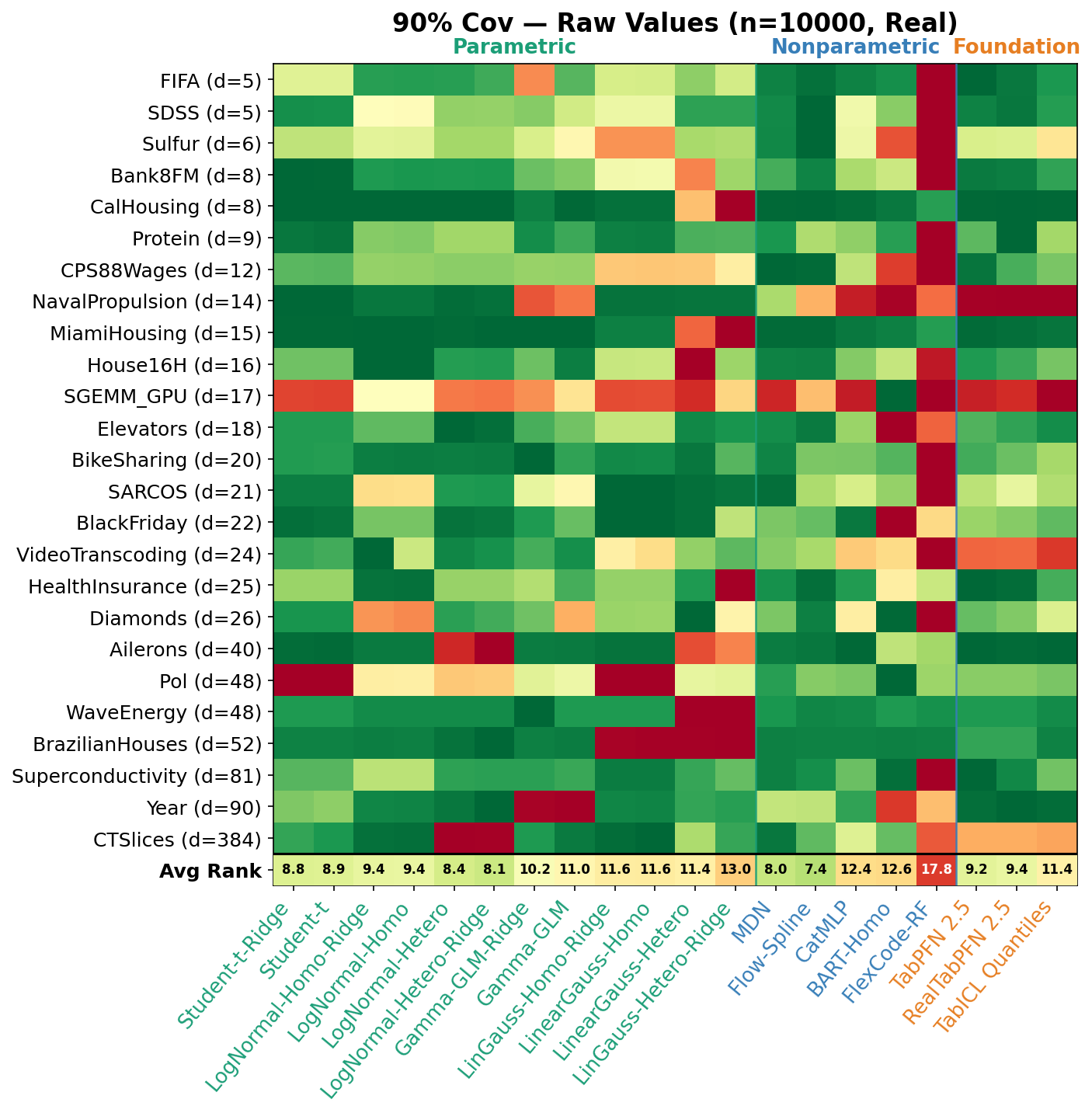}
  \caption{Raw 90\% Coverage -- $n=10000$, real data.}
\end{figure}
\begin{figure}[p]\centering
  \includegraphics[width=\linewidth]{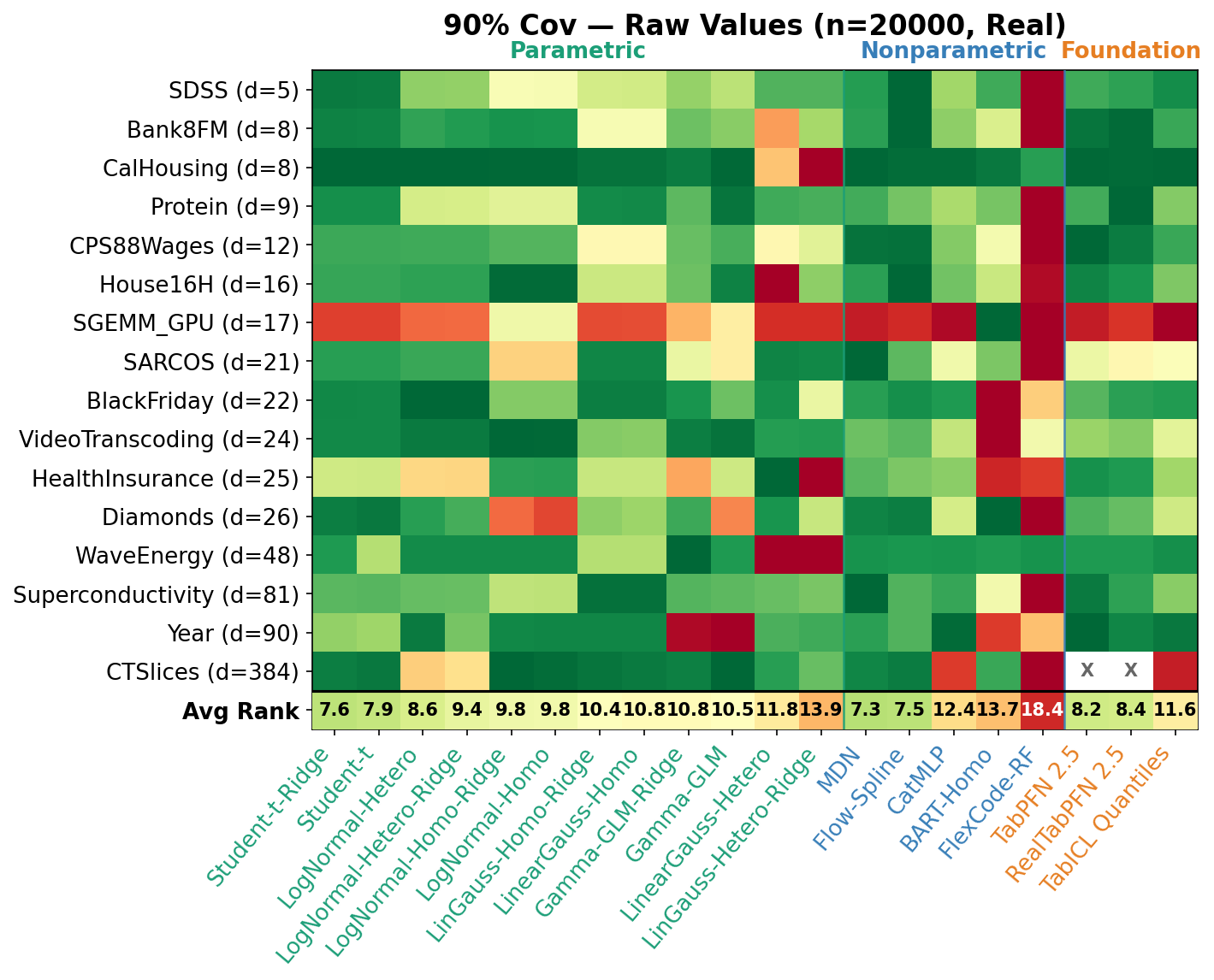}
  \caption{Raw 90\% Coverage -- $n=20000$, real data.}
\end{figure}

% ── Interval Width ──────────────────────────────────────────────────────────

% ── Fit Time ────────────────────────────────────────────────────────────────

\begin{figure}[p]\centering
  \includegraphics[width=\linewidth]{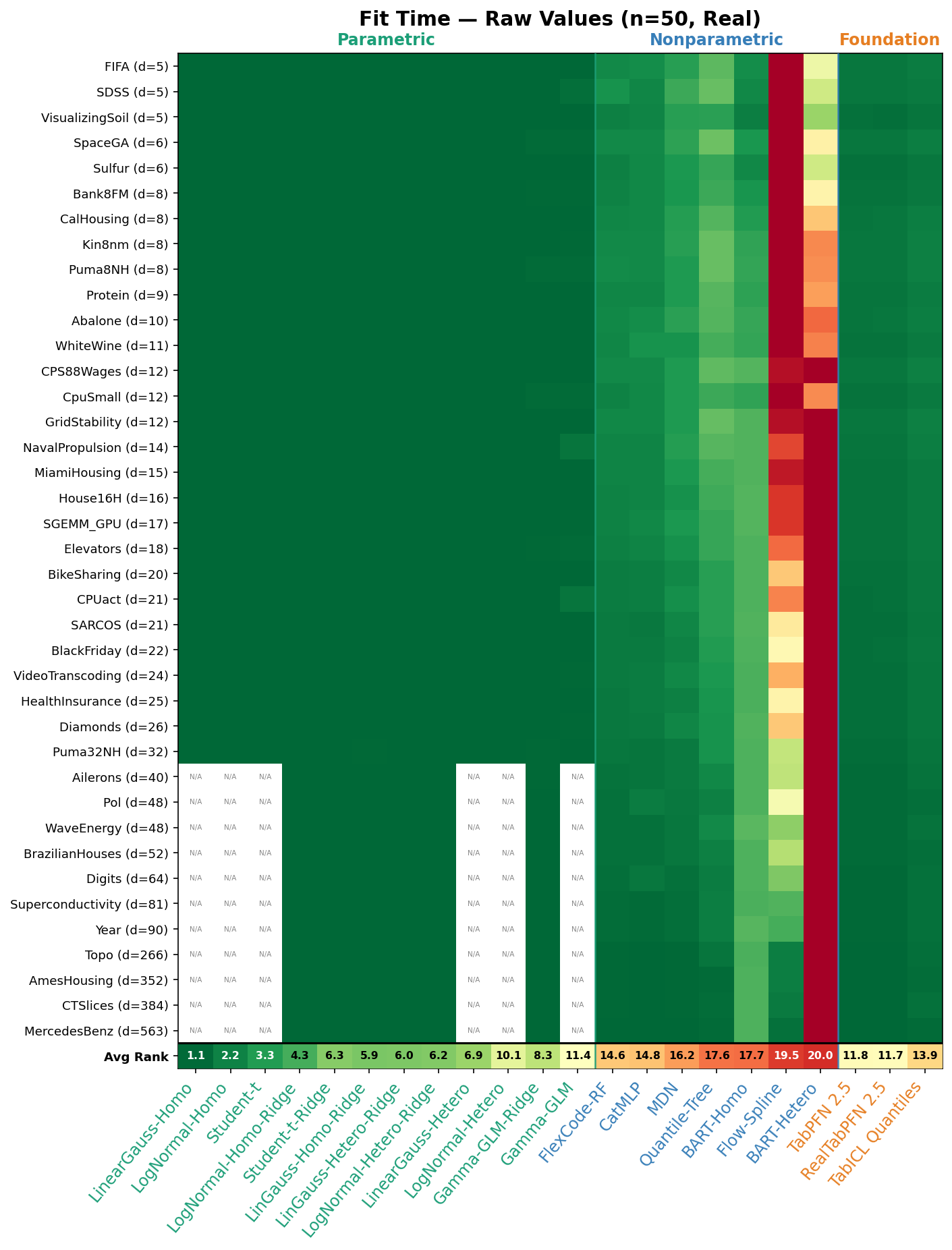}
  \caption{Raw Fit Time -- $n=50$, real data.}
\end{figure}
\begin{figure}[p]\centering
  \includegraphics[width=\linewidth]{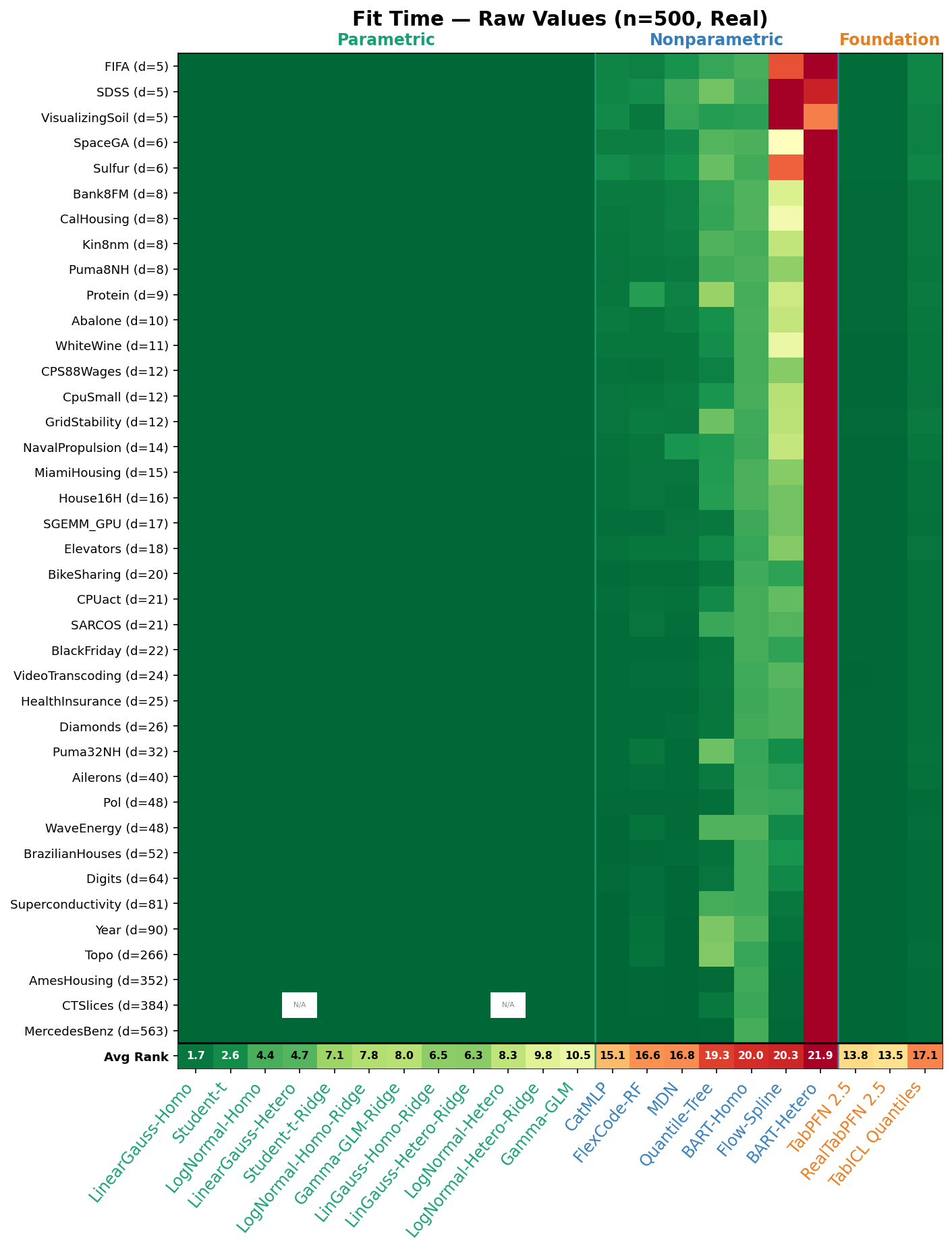}
  \caption{Raw Fit Time -- $n=500$, real data.}
\end{figure}
\begin{figure}[p]\centering
  \includegraphics[width=\linewidth]{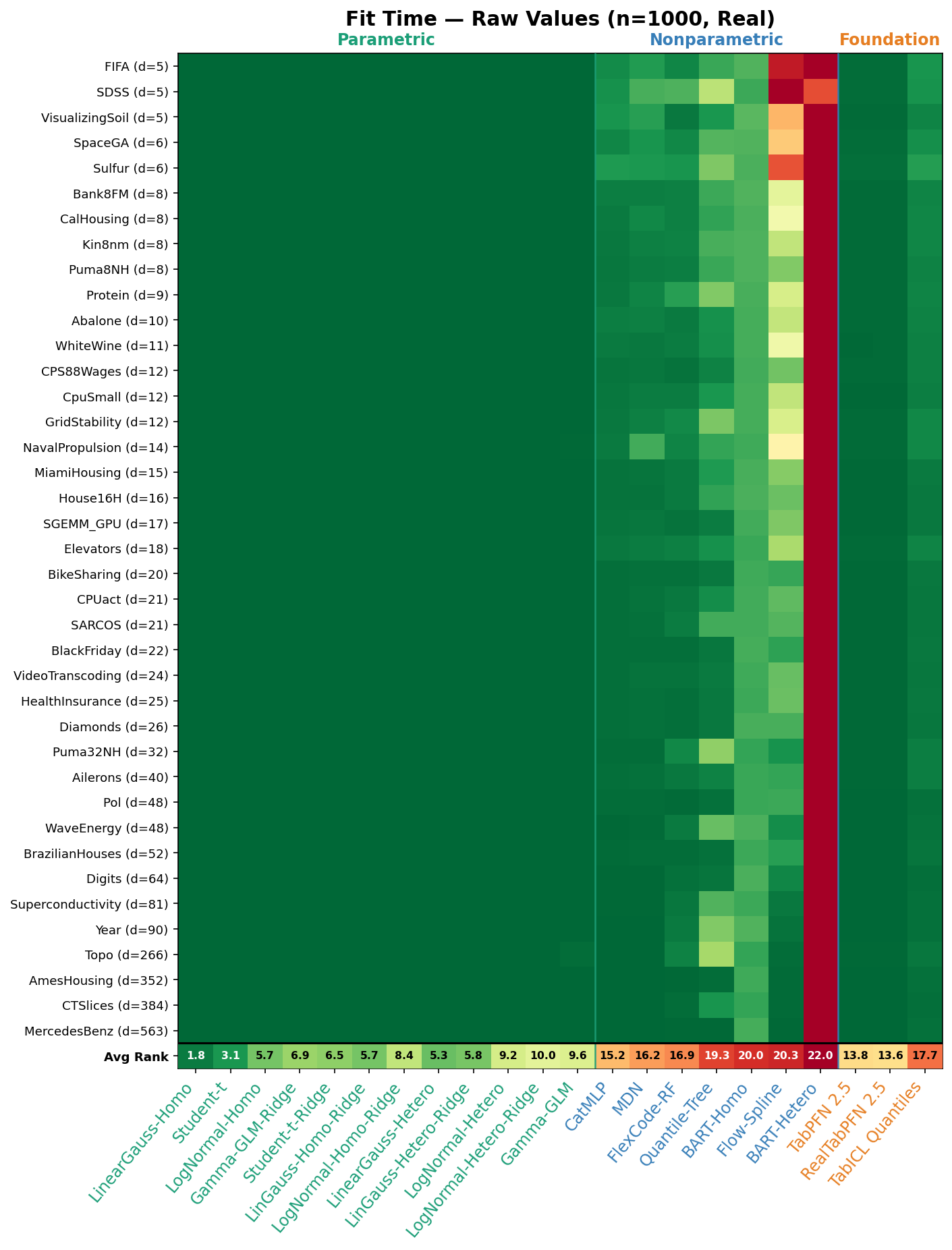}
  \caption{Raw Fit Time -- $n=1000$, real data.}
\end{figure}
\begin{figure}[p]\centering
  \includegraphics[width=\linewidth]{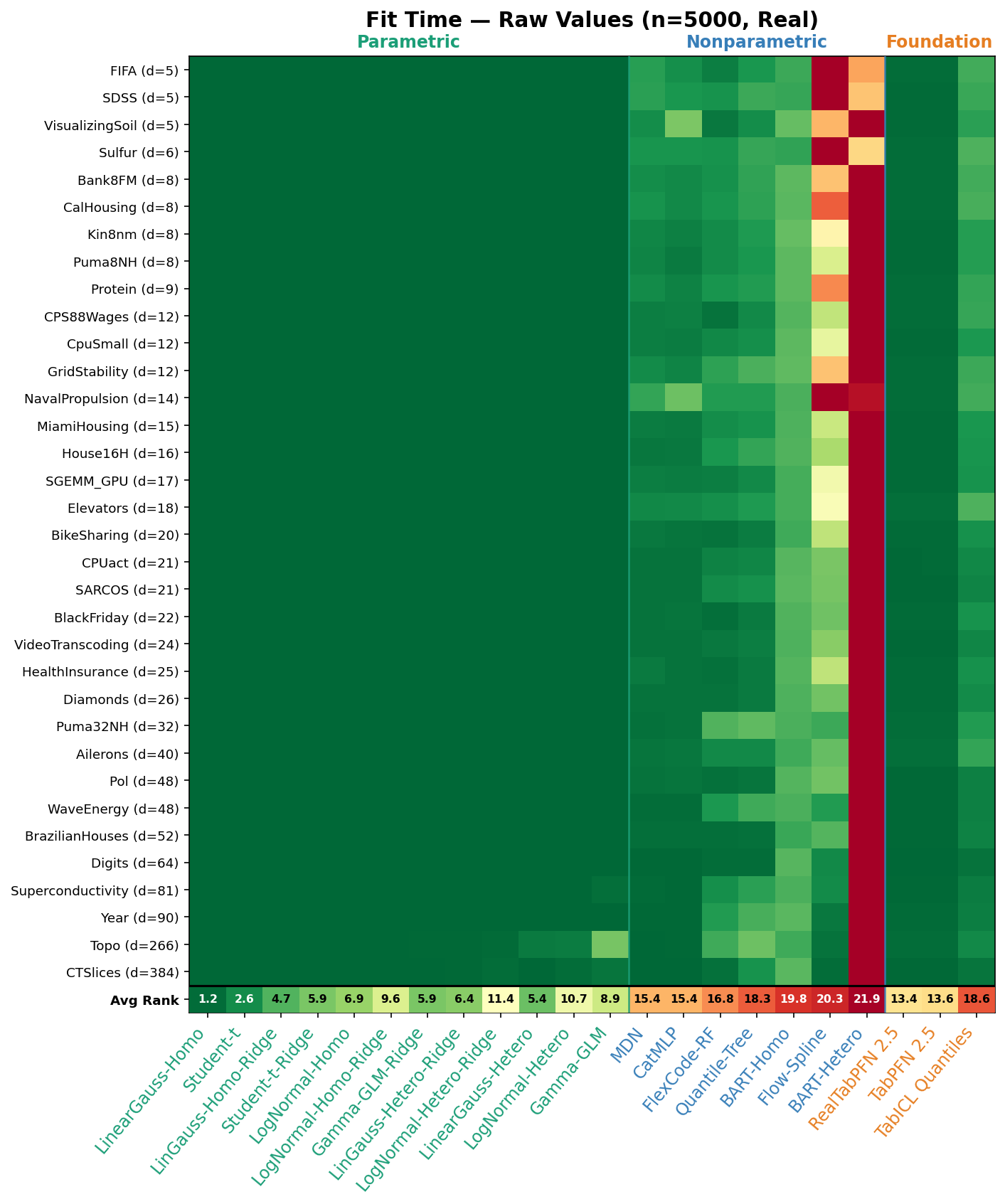}
  \caption{Raw Fit Time -- $n=5000$, real data.}
\end{figure}
\begin{figure}[p]\centering
  \includegraphics[width=\linewidth]{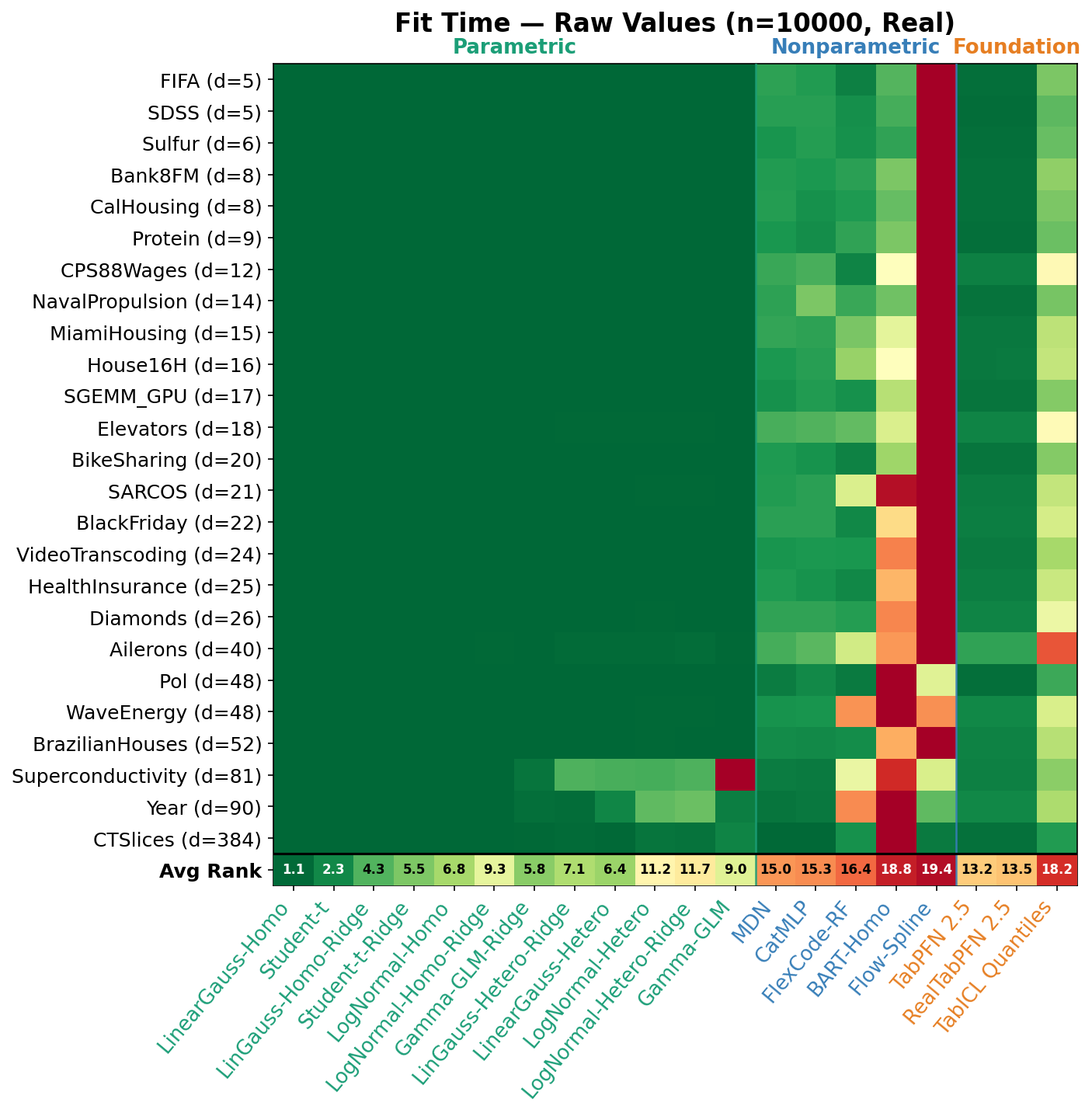}
  \caption{Raw Fit Time -- $n=10000$, real data.}
\end{figure}
\begin{figure}[p]\centering
  \includegraphics[width=\linewidth]{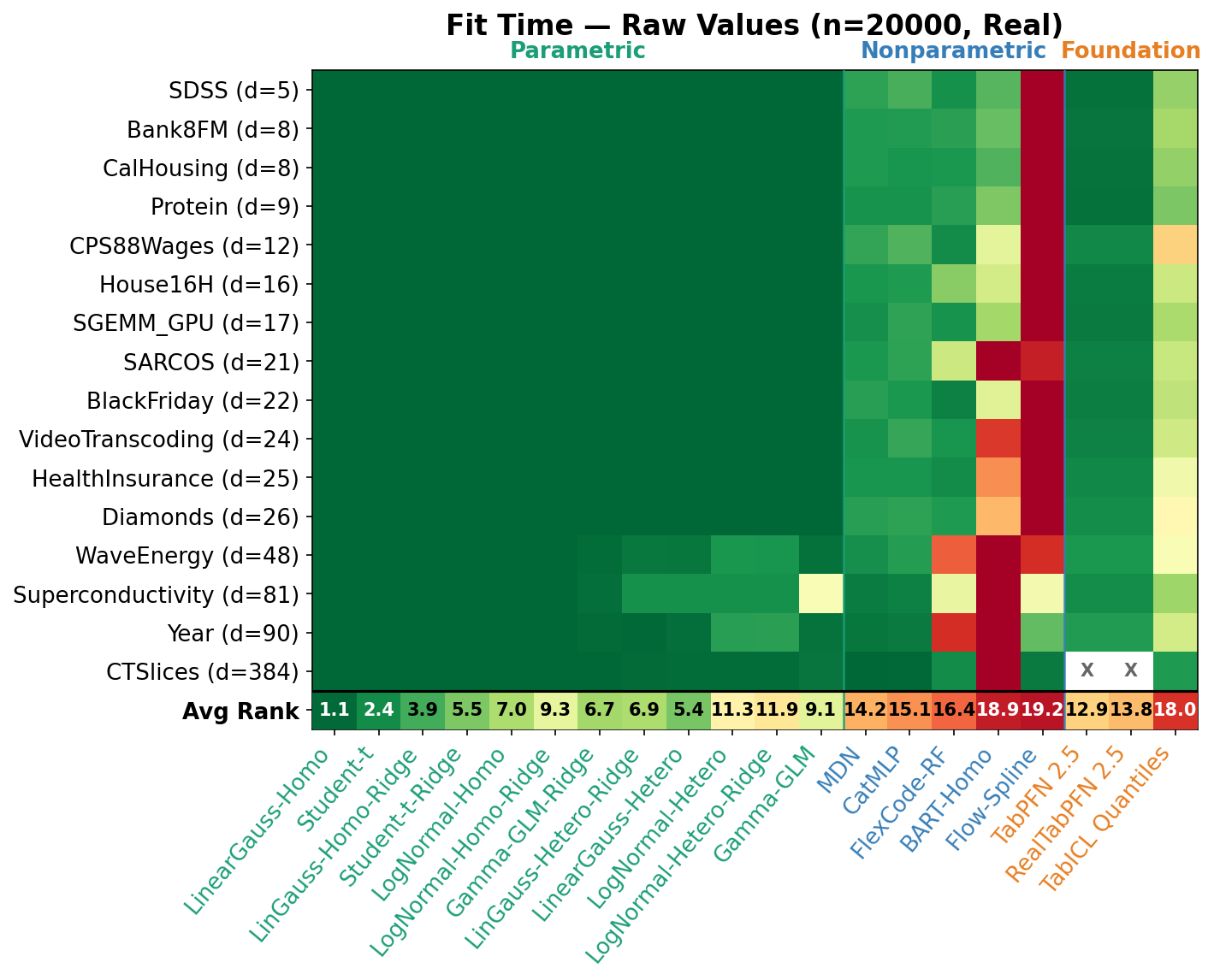}
  \caption{Raw Fit Time -- $n=20000$, real data.}
\end{figure}

\newpage

\subsection{SDSS}

This section shows all results for the SDSS  experiment.

\label{app:sdss}

\begin{figure}[p]\centering
  \includegraphics[width=\linewidth]{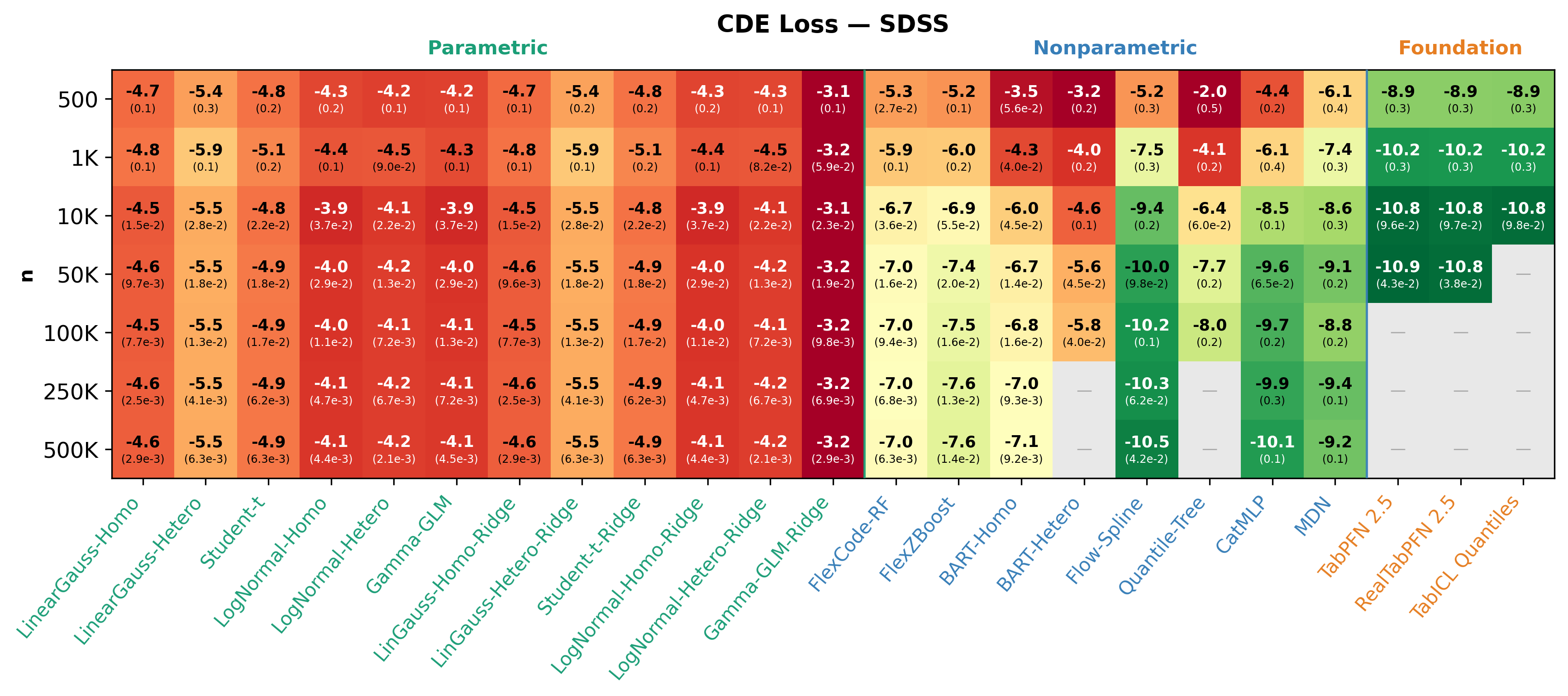}
  \caption{Raw CDE Loss -- SDSS, real data.}
\end{figure}

\begin{figure}[p]\centering
  \includegraphics[width=\linewidth]{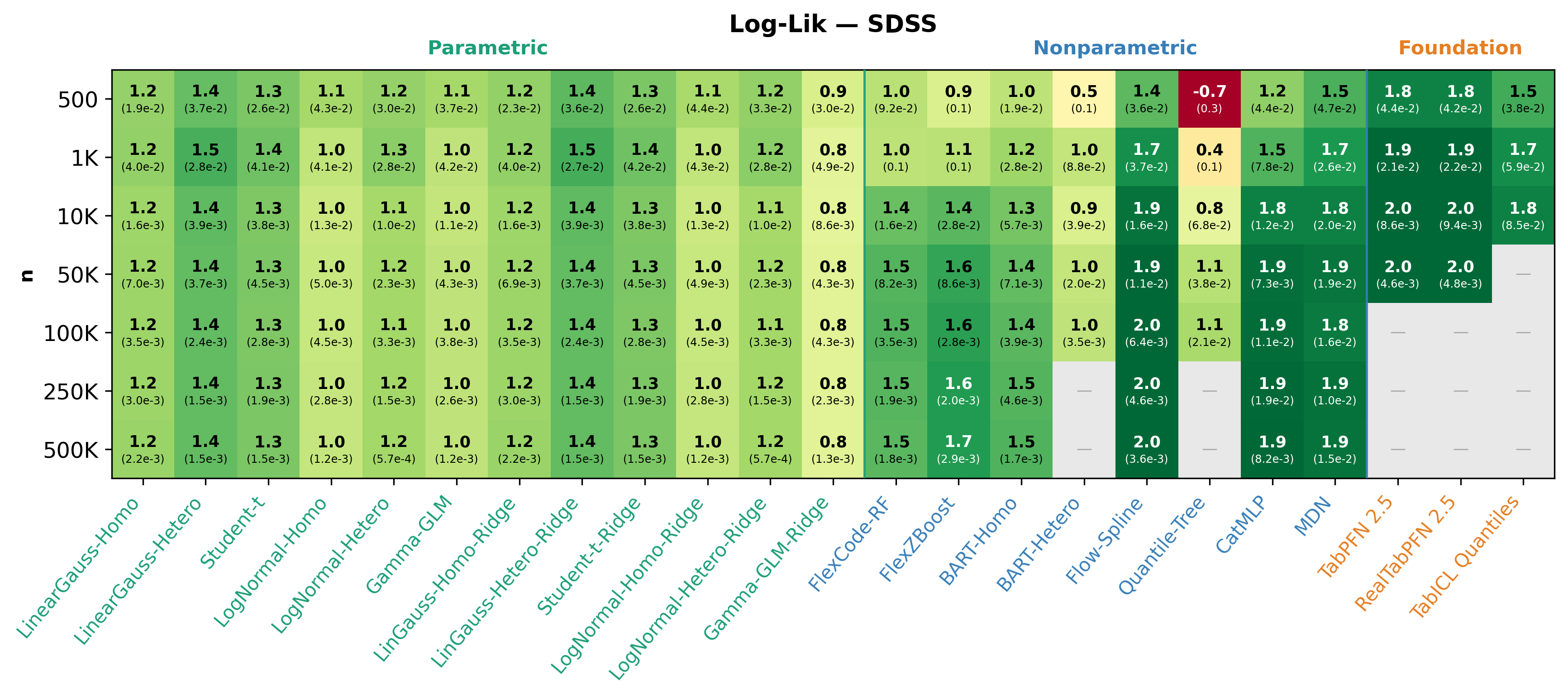}
  \caption{Raw Log-Likelihood -- SDSS, real data.}
\end{figure}

\begin{figure}[p]\centering
  \includegraphics[width=\linewidth]{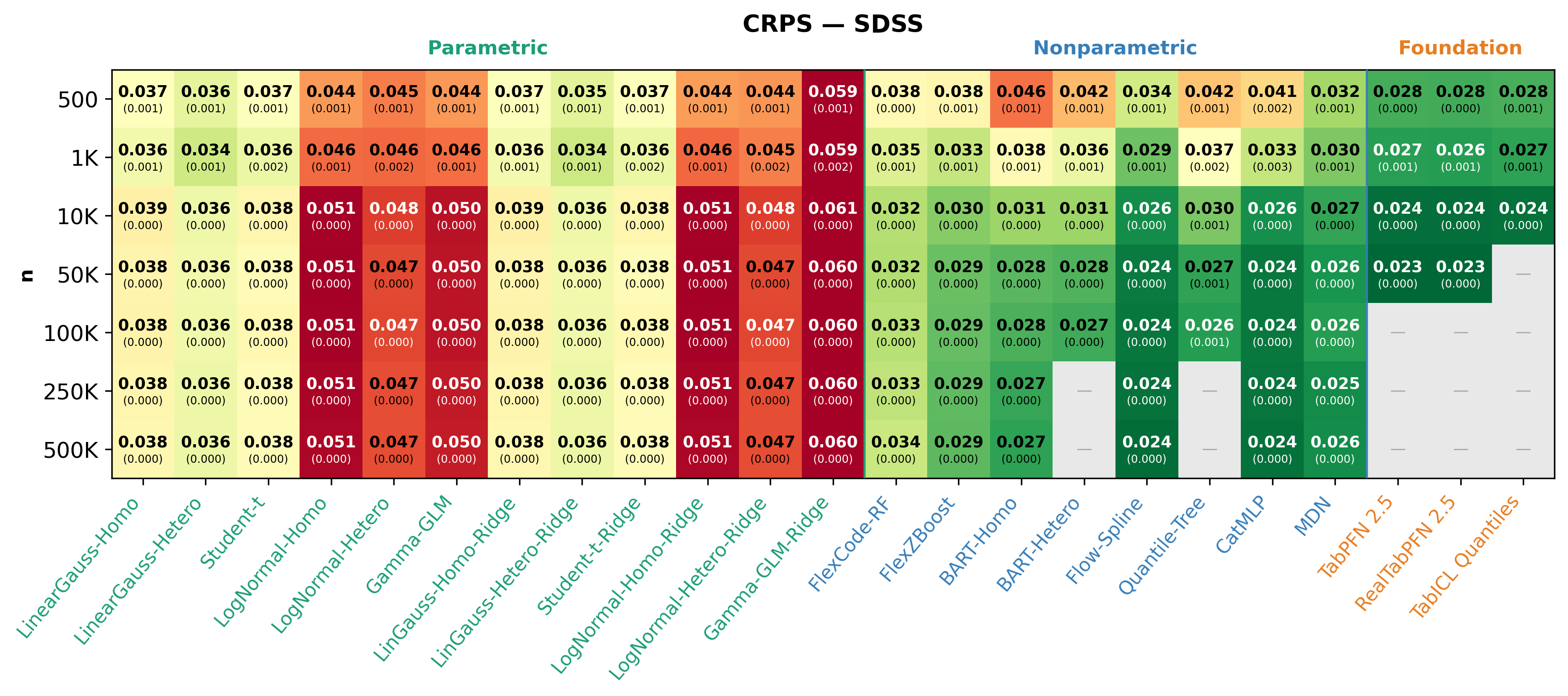}
  \caption{Raw CRPS -- SDSS, real data.}
\end{figure}

\begin{figure}[p]\centering
  \includegraphics[width=\linewidth]{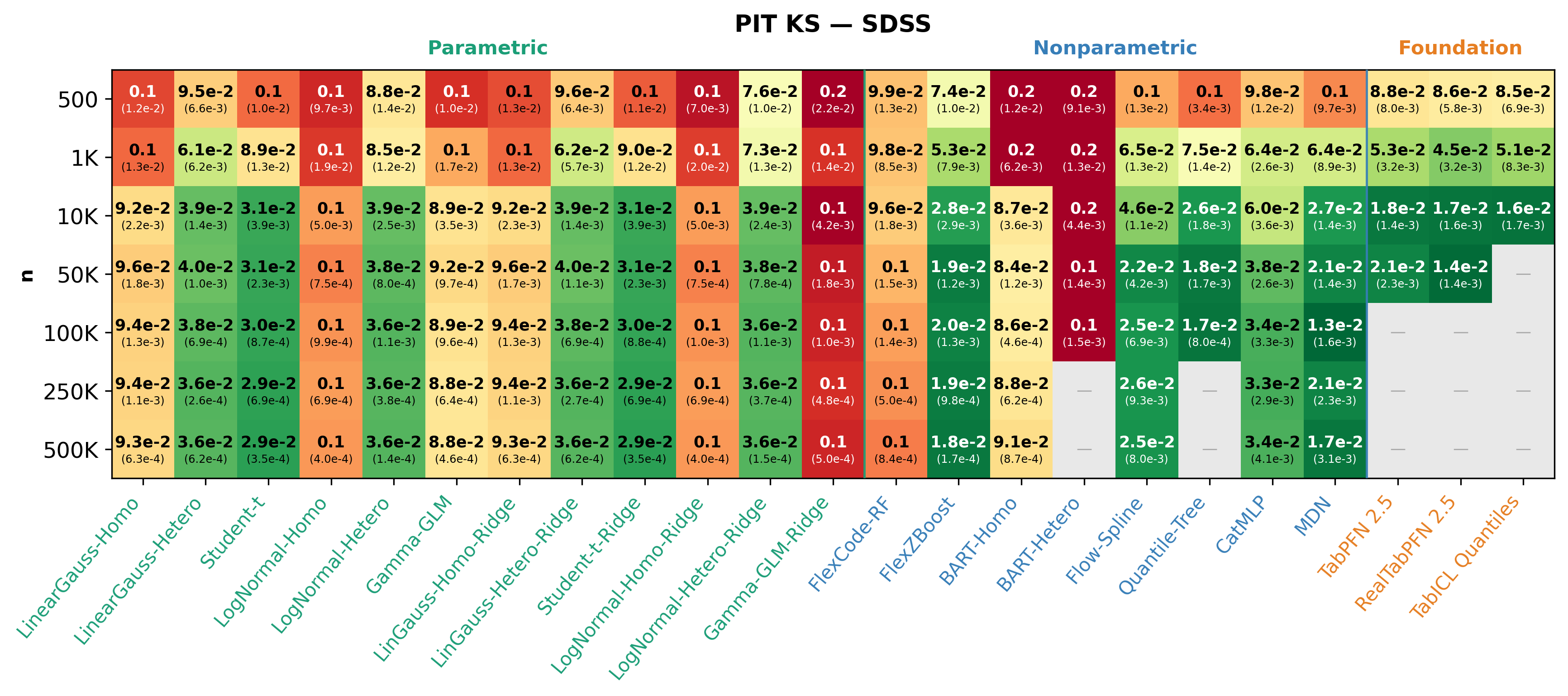}
  \caption{Raw PIT KS -- SDSS, real data.}
\end{figure}

\begin{figure}[p]\centering
  \includegraphics[width=\linewidth]{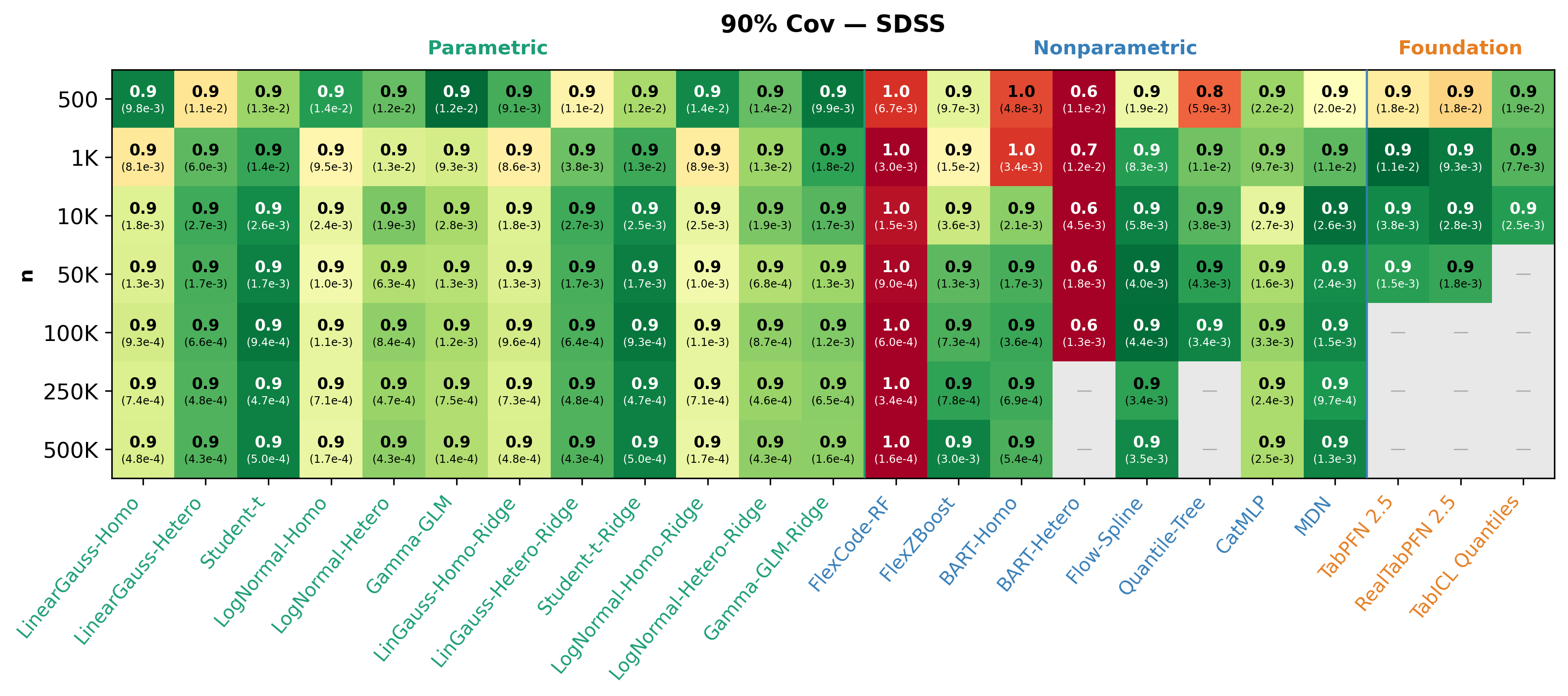}
  \caption{Raw 90\% Coverage -- SDSS, real data.}
\end{figure}

\end{document}